\documentclass[fleqn,twoside]{article}
\usepackage{espcrc1}
\newcommand{\AmS}{{\protect\the\textfont2
  A\kern-.1667em\lower.5ex\hbox{M}\kern-.125emS}}
\title{Discrete Route/Trajectory Decision Making Problems}
\author{Mark Sh. Levin
\address{
 Inst. for Information Transmission Problems, Russian Academy of
 Sciences\\
 19 Bolshoj Karetny Lane, Moscow 127994, Russia\\
 }
\thanks{
  {\it E-mail address}: mslevin@acm.org
  }
 }

\begin{document}

\maketitle

\begin{abstract}
 The paper focuses on composite multistage decision making problems which are targeted
 to design a route/trajectory from an initial decision situation
 (origin) to
 goal (destination) decision situation(s).
 Automobile routing problem is considered as a basic ``physical'' metaphor.
 The problems are based on a discrete (combinatorial) operations/states
 ``design/solving space'' (e.g., digraph).
 The described types of discrete decision making problems can be
 considered as ``intelligent''
 design of a route (trajectory, strategy)
 and can be used in many domains:
 (a) education (planning of student educational trajectory),
 (b) medicine (medical treatment),
 (c) economics (trajectory of start-up development).

 Several types of the route decision making
  problems are described:
  (i) basic route decision making,
  (ii) multi-goal route decision making,
  (iii) multi-route decision making,
  (iv) multi-route decision making with route/trajectory change(s),
  (v) composite multi-route decision making
  (solution is a composition of several routes/trajectories at several corresponding
  domains),
  and
 (vi) composite multi-route decision making
  with coordinated routes/trajectories.
 In addition, problems of
 modeling and building the ``design spaces'' are considered.
%
% Restructuring in the examined route/trajectory decision making
% problems is described as well.
%
 Numerical examples illustrate the suggested approach.
 Three applications are considered:
 educational trajectory (orienteering problem),
 plan of start-up company (modular three-stage design),
 and plan of medical treatment (planning over digraph with two-component vertices).
% (problem formulations, applications).

~~~~~~~~~~~


\vspace{1pc}
\end{abstract}

\newcounter{cms}
\setlength{\unitlength}{1mm}

%%%%%%%%%%%%%%%%%%%%%%%%%%%%%%%%%%%%%%%%%%
\tableofcontents

\newpage
%%%%%%%%%%%%%%%%%%%%%%%%%%%%%%%%%%%%%%%%%%%%%%%%%%%%%%%%
%%%%%%%%%%%%%%%%%%%%%%%%%%%%%%%%%%%%%%%%%%%%%%%%%%%%%%%%
\section{Introduction}

%\subsection{Preliminaries}

% In recent two decades,
% complex composite combinatorial problems
% (complex problem frameworks)
% have been used in many applied domains.
%
 This paper focuses on composite multistage discrete decision making problems
 which are targeted to design
 a route
 (trajectory, strategy)
  from an initial decision situation (source point, origin) to
 goal (destination) decision situation(s).
  The problems are based on  discrete (combinatorial) operations/states
  (i.e., a ``space'', e.g., digraph/network,
% cellular
 automata model).
 A generalized three-part scheme (morphological structure) of the examined domain
 (route/trajectory decision making problems)
  is depicted in Fig. 1.
%
%%%%%%%%%%%%%%%%%%%%%%%%%%%%%%%%%%%%%%%%%%%%%%%%%%%%
%
 Evidently, various versions of the shortest path problem
 correspond to a basis
% are the basic simple problem
 (``reference'' problem) for the
 considered class of combinatorial problems
 (e.g., \cite{ahu93,bellman58,cormen01,dijkstra56,epp99,gar79,hart68,tarata07,yen71}).
 Another basic problem corresponds to control engineering:
 designing a control trajectory (e.g., for controller)
 (e.g., \cite{dorf10,fried05,isid89,nise15,phil07,rho97}).
 The third  basic analogue for the problem corresponds to
 planning of mobile robot trajectory
 (e.g., \cite{bia08,lat91,sprong05,tian04}).
 The fourth basic analogue for the considered problem may be
 found as a search strategy in problem solving
 (e.g., \cite{luger04,nil71,shap87}).
 Two well-known problems
 as basic ``physical'' metaphors are:
% Automobile routing problems are considered as
% (basic ``physical'' metaphors):
%
 (i) automobile routing problems
 (e.g., \cite{boyse88,ding89,fr86,kumar05}),
 (ii) team orienteering problems
 (e.g., \cite{arch07,chao96a,gav14b,golden87,van11a}).

\begin{center}
%\begin{picture}(155,84)
\begin{picture}(155,86)

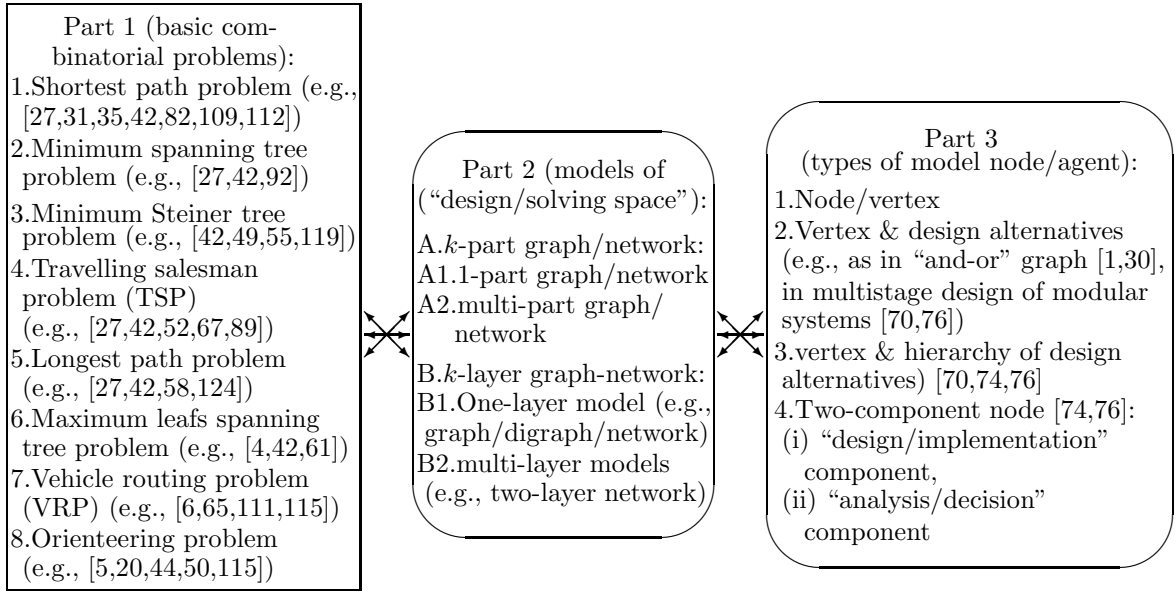
\put(013,00){\makebox(0,0)[bl]{Fig. 1. Generalized three-part
 scheme of route/trajectory decision making problems}}

%============================= Part 1

\put(00,06){\line(1,0){47}} \put(00,84){\line(1,0){47}}
\put(00,06){\line(0,1){78}} \put(47,06){\line(0,1){78}}

\put(07.5,79){\makebox(0,0)[bl]{Part 1 (basic com-}}
\put(06,75){\makebox(0,0)[bl]{binatorial  problems):}}

\put(0.5,71){\makebox(0,0)[bl]{1.Shortest path problem (e.g.,}}
\put(02,67){\makebox(0,0)[bl]{\cite{cormen01,dijkstra56,epp99,gar79,mart84,tarata07,tung92})}}

\put(0.5,63){\makebox(0,0)[bl]{2.Minimum spanning tree}}
\put(02,59){\makebox(0,0)[bl]{problem
  (e.g., \cite{cormen01,gar79,pettie02})}}

\put(0.5,55){\makebox(0,0)[bl]{3.Minimum Steiner tree}}
\put(02,51){\makebox(0,0)[bl]{problem
 (e.g., \cite{gar79,goemans93,hwang92,winter87})}}

\put(0.5,47){\makebox(0,0)[bl]{4.Travelling salesman  }}
\put(02,43){\makebox(0,0)[bl]{problem (TSP) }}
\put(02,39){\makebox(0,0)[bl]{(e.g.,
  \cite{cormen01,gar79,gutin06,lawler85,papa82})}}

\put(0.5,35){\makebox(0,0)[bl]{5.Longest path problem}}
\put(02,31){\makebox(0,0)[bl]{(e.g.,
 \cite{cormen01,gar79,karg97,zhang07})}}

\put(0.5,27){\makebox(0,0)[bl]{6.Maximum leafs spanning}}
\put(02,23){\makebox(0,0)[bl]{tree problem
  (e.g., \cite{alon09,gar79,klei91})}}
%\put(04,23){\makebox(0,0)[bl]{\cite{alon09,gar79,klei91}}}

\put(0.5,19){\makebox(0,0)[bl]{7.Vehicle routing problem}}
\put(02,15){\makebox(0,0)[bl]{(VRP) (e.g.,
 \cite{arch08,lap09,toth01,van11a})}}

\put(0.5,11){\makebox(0,0)[bl]{8.Orienteering problem}}
\put(02,07){\makebox(0,0)[bl]{(e.g.,
 \cite{arch07,chao96a,gav14b,golden87,van11a})}}

%------------

\put(51.5,39){\vector(-1,1){4}} \put(51.5,41){\vector(-1,-1){4}}
\put(49.5,39){\vector(1,1){4}} \put(49.5,41){\vector(1,-1){4}}
\put(49.5,40){\vector(1,0){4}} \put(51.5,40){\vector(-1,0){4}}

%============================ Part 2

\put(74,40){\oval(40,54)}

\put(60.5,60){\makebox(0,0)[bl]{Part 2 (models of}}

\put(54.5,56){\makebox(0,0)[bl]{(``design/solving space''):}}

\put(54.5,50){\makebox(0,0)[bl]{A.\(k\)-part graph/network:}}

\put(54.5,46){\makebox(0,0)[bl]{A1.\(1\)-part graph/network}}

\put(54.5,42){\makebox(0,0)[bl]{A2.multi-part graph/}}
\put(59.5,39){\makebox(0,0)[bl]{network}}

\put(54.5,33){\makebox(0,0)[bl]{B.\(k\)-layer graph-network:}}

\put(54.5,29){\makebox(0,0)[bl]{B1.One-layer model (e.g.,}}

\put(55.5,25){\makebox(0,0)[bl]{graph/digraph/network)}}

\put(54.5,21){\makebox(0,0)[bl]{B2.multi-layer models  }}

\put(55.5,17){\makebox(0,0)[bl]{(e.g., two-layer network)}}

%--------------

\put(98.5,39){\vector(-1,1){4}} \put(98.5,41){\vector(-1,-1){4}}
\put(96.5,39){\vector(1,1){4}} \put(96.5,41){\vector(1,-1){4}}
\put(96.5,40){\vector(1,0){4}} \put(98.5,40){\vector(-1,0){4}}

%================== Part 3

\put(128,40){\oval(54,62)}

\put(122,65){\makebox(0,0)[bl]{Part 3}}

\put(105.5,61){\makebox(0,0)[bl]{(types of model node/agent):}}

\put(102,56){\makebox(0,0)[bl]{1.Node/vertex}}
\put(102,52){\makebox(0,0)[bl]{2.Vertex \& design alternatives}}
\put(103,48){\makebox(0,0)[bl]{(e.g., as in ``and-or'' graph
 \cite{adel02,de90},}}

\put(103,44){\makebox(0,0)[bl]{in multistage design of modular}}

\put(103,40){\makebox(0,0)[bl]{systems
 \cite{lev06,lev15})}}

\put(102,36){\makebox(0,0)[bl]{3.vertex \& hierarchy of design}}
\put(103,32){\makebox(0,0)[bl]{alternatives)
 \cite{lev06,lev13tra,lev15}}}

\put(102,28){\makebox(0,0)[bl]{4.Two-component node
 \cite{lev13tra,lev15}: }}

\put(103,24){\makebox(0,0)[bl]{(i) ``design/implementation''}}

\put(106,20){\makebox(0,0)[bl]{component,}}

\put(103,16){\makebox(0,0)[bl]{(ii) ``analysis/decision''}}

\put(106,12){\makebox(0,0)[bl]{component}}

\end{picture}
\end{center}

 In this paper, the shortest path problem
 (part 1 from Fig. 1)
 is composed
 with one/two-layer model (part 2 from Fig. 1)
 and various versions of node/agent types (part 3 from Fig. 2).
 In addition, the team orienteering problem is used as well
 (in an educational example).
%
%%%%%%%%%%%%%%%%%%%%%%%%%%%%%%%%%
% A generalized list of
  A list of  various route/trajectory-like  decision making problems
%  from the structural  viewpoint
  is pointed out in Table 1.

 The suggested type of composite decision making problems can be
 considered as ``intelligent'' design of a route
 (trajectory, strategy)
% and  can be  used
  in many domains, for example:
 (a) education (e.g., planning of student educational trajectory \cite{lev98}),
 (b) medicine (e.g., medical treatment planning/scheduling
  \cite{lev06,lev15,levsok04,petr11}),
%
% (c) economics (e.g., trajectory of start-up development),
%
 (c) tourism (e.g., tourism route planning/recommendation
% or  Tourist Trip Design Problem
  \cite{gav14b,gav15,souf08,van07,van11}).
%
% (d) manufacturing systems (e.g.,
%  technological routing in manufacturing),
%
% (d) system maintenance (e.g., planning of maintenance operations,
% team orienteering problem \cite{}).
%
 Here, it is  necessary to do the following:
 (1) to build the operations (design/solving) ``space'';
 (2) to specify the goal (possible resultant)
 point (or set of goal points);
 (3) to design the route at the design ``space'',
 (4) online analysis of the route implementation
 and online modification of the route
 (if needed).

 In the paper, several types of the route decision making
  problems are described:
  (i) basic route decision making,
  (ii) multi-goal route decision making,
  (iii) multi-route decision making,
  (iv) multi-route decision making with route change(s),
  (v) composite multi-route decision making
  (solution is a composition of several routes at several corresponding
  domains),
  and
 (vi) composite multi-route decision making
  with coordinated routes.
%%%%%%%%%%%%%%%%%%%%%%%%%%%%%%%%%%%%%%%%%%%%%%%%%%%%%%%%%%%%%
%
% Restructuring in the examined route decision making
% problems is described as well.
%
 The suggested composite approaches/frameworks
 are illustrated by numerical examples including
 three applications:
 (i)design of an individual educational trajectory for a Bachelor student
 (version of multicriteria orienteering problem),
 (ii) planning a development trajectory for a start-up company
  (modular three-stage design based on hierarchical morphological design),
 and
 (iii) planning a medical treatment
 (route/trajectory over digraph with two-component vertices).
% (problem formulations, applications).

%\newpage
\begin{center}
{\bf Table 1.} Route (trajectory/strategy) decision making problems \\
\begin{tabular}{| c | l | l| l |}
\hline
 No.  & Problem  &Illustration & Source \\
\hline

 1. & Basic ``reference''
% analogue
  problems:&&\\

 1.1.& Shortest path problem (including \(k\)-path and
% multi-objective
  & & \cite{boff95,cormen01,dijkstra56,epp99,gar79,hart68,mart84}\\

 &multicriteria formulations)&&\cite{tarata07,tung92,yen71}\\

 1.2.& Design of control trajectory in
%  control
 parameter space&&
 \cite{dorf10,fried05,isid89,nise15,phil07,rho97}\\
% system control&& \cite{}\\
% &parameter space&&\\

% &space&&\\

 1.3.& Design of search strategy in problem solving&& \cite{luger04,nil71,shap87}\\

 1.4.& Design of trajectory for mobile robot movement &
 & \cite{bia08,kavr96,lat91,sprong05,tian04,wood06} \\

 1.5.& Design of a mission for airplane/aerospace &
 &\cite{bell02,botr00,camp00,jun03,kim13,scharf03} \\
 &apparatus (e.g., unmanned aerial vehicles UAVs)&&\cite{scho06,stev03,wzorek06}\\

% Unmanned Aerial/air Vehicle UAVs

 1.6.& Vehicle routing problems &&\cite{arch08,lap09,toth01}\\

 1.7.& X-cast (i.e., anycast broadcast, multicast, unicast,
  &&\cite{cavendish06,chang03,corde03,foro12,grub11,ko03,lee10}\\

 &geocast) routing in communication/sensor&&\cite{low03,mai04,medhi07,pant10,pant11}\\

 &networks&&\cite{perkins01,rap96,saha00,wies00}\\

 1.8.& Design of system development trajectory for &
 &\cite{lev06,lev15}\\

 &modular system (multistage design)&&\\

%\hline

 2.& Basic route DM problems and applied problems: & &\\

 2.1.& Basic route DM problem & Fig. 2&\\

 2.2.& Motion planning, navigation (urban traffic planning,
   &&\cite{boyse88,choset05,cur93,ding89,fr86,kavr96,kumar05} \\

 & automobile routing, robot motion planning,  &&\cite{lav98,lat91,svet98} \\

 & inspection path planning, etc.) &&\\

 2.3.& Team orienteering problem (visiting a subset of && \cite{arch07,chao96a,chao96b,gav15,souf10,van11a} \\

 &graph nodes, combination of knapsack problem &Fig. 7&\\

 &and TSP)&&\\

%% NB: For system testing NB

 2.4.& Tourism route planning/recommendation
  &&\cite{gav14b,gav15,souf08,van07,van11}\\
 &(Tourist Trip Design Problem)&&\\

  2.5.&
% Search trajectory in Web systems
 Trajectory search in information systems
 &    &\cite{bose01,czy03,fuhr01,lev12hier,lev15,wei13}\\

 &(e.g., web, including hotlink assignment problems)&&\\

 2.6.&Planning of system maintenance&&\\

 2.7.&Scenario planning/multistage scenario  planning&& \cite{brad05,cherm05,godet00,ring98,schoem95,varum10}\\
% Multi-goal Multi-stage scenario network/hierarchy/tree NB

 2.8.& Planning of student educational trajectory &Fig. 18&\cite{lev98}\\

 2.9.&  Medical treatment planning/scheduling&Fig. 25&
 \cite{lev06,lev15,levsok04,petr11}\\
 & (two-layer ``design space'' and problem)&&\\

 2.10.& Trajectory of start-up development&  &\cite{lev15}\\

%\hline

 3.& Examined structural problems:&&\\

 3.1.& Multi-route DM problem &Fig. 4&\\

 3.2.& Multi-goal multi-route DM problem &Fig. 5&\\

 3.3.& Multi-route DM problem with route change(s) &Fig. 6&\\

 3.4.& Composite (multi-domain) multi-route DM problem&Fig. 10&\\
  &(solution is a composition of several routes)&&\\
%  &at several corresponding domains)&&\\

 3.5.& Composite (multi-domain)  multi-route  DM problem
  & Fig. 11&\\
   & with coordinated routes&&\\

  3.6.& Composite three-domain  multi-route  DM problem
  & Fig. 12&\\
   &(three basic combinatorial problems)&&\\

   3.6.& Composite two-layer five-domain  multi-route
  & Fig. 13&\\
   &DM problem in communication system&&\\

% 3.7.& Composite (multi-domain)  multi-route  DM problem  & Fig. 16&\\
%   & (based on orienteering problems)&&\\

\hline
\end{tabular}
\end{center}

%%%%%%%%%%%%%%%%%%%%%%%%%%%%%%%%%%%%%%%%%%%%%%%%%%%%%%
\section{Route decision making  problems}

\subsection{Basic problems}

% Evidently, various versions of the shortest path problem
% correspond to a basis (``reference'' problem) for the
% considered class of combinatorial problems
% (e.g., \cite{gar79,tarata07}).
%
% Another basic problem corresponds to control engineering:
% designing a control trajectory (e.g., for controller)
% (e.g., \cite{}).
%
% The third basic analogues for the considered problem may be
% found as a search strategy in problem solving
% (e.g., \cite{luger04}).

% Further, our complicated combinatorial versions of
% the route decision making problems are under examination.

%\subsection{Basic problem}

 The basic routing decision making problem
 (``physical'' car routing)
 is depicted in Fig. 2.
 Here, graph \(G = (H,E)\) is given,
  initial vertex \(h^{0}\in H\)
  and goal (destination) vertex \(h^{g}\in H\)
  are pointed out,
 each edge/arc \(e \in E\) has a length (i.e., positive weight, cost) \(a(e)\).
 The problem is:

~~

 Find the route (path) from vertex \(h^{0} \) to vertex \(h^{g}\)
 \(L = <h^{0},...,h^{g}>\)
 that minimizes the length (cost) of the path
 (i.e., the sum of the path edges/arcs weights).

~~

 Note, various versions of the problems are under examination
 including searching for \(k\)-path problem,
 multi-objective problems,
 online problems
  (e.g.,
  \cite{cormen01,dijkstra56,epp99,gar79,hart68,tarata07,yen71}).
 Several polynomial algorithms have been suggested
  for the problem
  (including polynomial algorithms for multi-objective versions)
 (e.g., \cite{cher96,cormen01,dijkstra56,gar79,mart84,tarata07,tung92,xie12})

  Generally, the following support problems can be pointed out:
 (i) building the route (design),
 (ii) online analysis of the route implementation
  and online modification (correction) of the route.
%
%\subsection{Basic simplification approach}
%
 Basic simplification approach consists in partitioning
 the initial solving ``space'' into series of
 ``subspaces'' (Fig. 3). This approach is close to dynamic programming scheme.

\begin{center}
%\begin{picture}(76,54)
\begin{picture}(81,45)
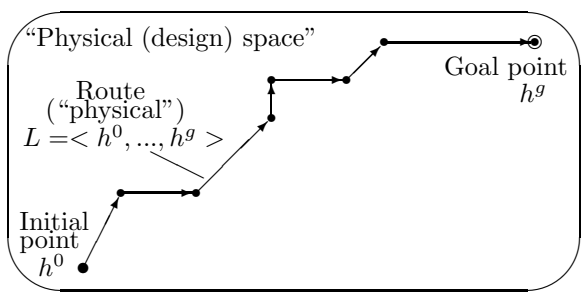
\put(01,00){\makebox(0,0)[bl]{Fig. 2. Basic routing problem
 (e.g., car routing)}}

\put(02,38.7){\makebox(0,0)[bl]{``Physical (design) space''}}

\put(38,25.5){\oval(76,37)}

%------------------ Goal point

\put(70,40){\circle{1.7}} \put(70,40){\circle*{1.2}}

\put(58,35){\makebox(0,0)[bl]{Goal point}}
\put(68,32){\makebox(0,0)[bl]{\(h^{g}\)}}

%------------ Path

\put(09,32.5){\makebox(0,0)[bl]{Route}}
\put(05,29){\makebox(0,0)[bl]{(``physical'')}}
\put(02,25.5){\makebox(0,0)[bl]{\(L=<h^{0},...,h^{g}>\)}}

\put(19,25.5){\line(2,-1){07}}

\put(50,40){\vector(1,0){19.5}}

\put(45,35){\vector(1,1){4.5}} \put(50,40){\circle*{1.0}}

\put(35,35){\vector(1,0){9.5}} \put(45,35){\circle*{1.0}}

\put(35,30){\vector(0,1){04.5}} \put(35,35){\circle*{1.0}}

\put(25,20){\vector(1,1){09.5}} \put(35,30){\circle*{1.0}}

\put(15,20){\vector(1,0){09.5}} \put(25,20){\circle*{1.0}}

\put(10,10){\vector(1,2){4.7}} \put(15,20){\circle*{1.0}}

%------------------ Initial point

\put(10,10){\circle*{1.4}}

\put(01.5,15){\makebox(0,0)[bl]{Initial}}
\put(01.5,12){\makebox(0,0)[bl]{point}}
\put(04,09){\makebox(0,0)[bl]{\(h^{0}\)}}

\end{picture}
%\end{center}
%
%\begin{center}
\begin{picture}(76,45)

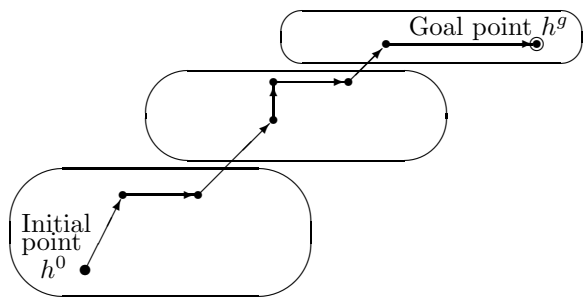
\put(02,00){\makebox(0,0)[bl]{Fig. 3. Simplified/partitioned
 ``solving space''}}

%\put(05,48){\makebox(0,0)[bl]{``Physical ~~ space''}}

\put(56,41){\oval(40,07)}

\put(38,30.5){\oval(40,12)}

\put(20,15){\oval(40,17)}

%------------------ Goal point

\put(70,40){\circle{1.7}} \put(70,40){\circle*{1.2}}

\put(53,40.6){\makebox(0,0)[bl]{Goal point \(h^{g}\)}}

%------------ Path

%\put(14,40.5){\makebox(0,0)[bl]{Route}}
%\put(10,37){\makebox(0,0)[bl]{(``physical'')}}
%\put(23,37){\line(3,-2){10}}

\put(50,40){\vector(1,0){19.5}}

\put(45,35){\vector(1,1){4.5}} \put(50,40){\circle*{1.0}}

\put(35,35){\vector(1,0){9.5}} \put(45,35){\circle*{1.0}}

\put(35,30){\vector(0,1){04.5}} \put(35,35){\circle*{1.0}}

\put(25,20){\vector(1,1){09.5}} \put(35,30){\circle*{1.0}}

\put(15,20){\vector(1,0){09.5}} \put(25,20){\circle*{1.0}}

\put(10,10){\vector(1,2){4.7}} \put(15,20){\circle*{1.0}}

%------------------ Initial point

\put(10,10){\circle*{1.4}}

\put(01.5,15){\makebox(0,0)[bl]{Initial}}
\put(01.5,12){\makebox(0,0)[bl]{point}}
\put(04,09){\makebox(0,0)[bl]{\(h^{0}\)}}

\end{picture}
\end{center}

%%%%%%%%%%%%%%%%%%%%%%%%%%%%%%%%%%%%%%%%==========================================
%\subsection{Structural problem formulations}

%\subsection{Problem statements}

 Now let us consider some
  other versions of basic route/trajectory DM problems.
% our versions of the intelligent route decision making problem.

%
 A multi-route DM problem is depicted in Fig. 4
 (i.e., searching for the best
 \(k\) routes or  concurrent examination of
 several different basic (e.g., shortest path) problems).
 On other hand, an analogical multi-goal problem can be
 considered (Fig. 5)
 (e.g., a set of basic problems with different goal nodes).

\begin{center}
%\begin{picture}(76,54)
\begin{picture}(79,54)
\put(12.5,00){\makebox(0,0)[bl]{Fig. 4. Multi-route  DM
 problem}}

\put(03.5,48){\makebox(0,0)[bl]{``Design/solving space''}}

\put(38,30){\oval(74,46)}

%------------------ Goal point

\put(65,50){\circle{1.7}} \put(65,50){\circle*{1.2}}

\put(65.5,47){\makebox(0,0)[bl]{Goal}}
\put(65.5,44){\makebox(0,0)[bl]{point}}
\put(68,41){\makebox(0,0)[bl]{\(h^{g}\)}}

%------------ Path
%------------------- Route 1 (Top)

\put(13.5,41){\makebox(0,0)[bl]{Route}}
\put(17,37.5){\makebox(0,0)[bl]{\(L_{1}\)}}

\put(45,50){\vector(1,0){19.5}}

\put(40,45){\vector(1,1){04.5}} \put(45,50){\circle*{1.0}}

\put(25,45){\vector(1,0){14.5}} \put(40,45){\circle*{1.0}}

\put(25,40){\vector(0,1){4.5}} \put(25,45){\circle*{1.0}}

\put(20,35){\vector(1,1){04.5}} \put(25,40){\circle*{1.0}}

\put(20,25){\vector(0,1){09.5}} \put(20,35){\circle*{1.0}}

\put(10,10){\vector(2,3){9.7}} \put(20,25){\circle*{1.0}}

%------------------- Route 2 (middle)

\put(25.3,34){\makebox(0,0)[bl]{Route}}
\put(29,30.5){\makebox(0,0)[bl]{\(L_{2}\)}}

\put(55,45){\vector(2,1){09.5}}

\put(45,40){\vector(2,1){9.5}} \put(55,45){\circle*{1.0}}

\put(35,40){\vector(1,0){9.5}} \put(45,40){\circle*{1.0}}

\put(35,30){\vector(0,1){09.5}} \put(35,40){\circle*{1.0}}

\put(25,20){\vector(1,1){09.5}} \put(35,30){\circle*{1.0}}

\put(10,10){\vector(3,2){14.7}} \put(25,20){\circle*{1.0}}

%------------------- Route 3 (bottom)

\put(46.5,25){\makebox(0,0)[bl]{Route}}
\put(50,21.5){\makebox(0,0)[bl]{\(L_{3}\)}}

\put(60,40){\vector(1,2){04.5}}

\put(55,30){\vector(1,2){4.5}} \put(60,40){\circle*{1.0}}

\put(45,30){\vector(1,0){9.5}} \put(55,30){\circle*{1.0}}

\put(45,25){\vector(0,1){04.5}} \put(45,30){\circle*{1.0}}

\put(35,15){\vector(1,1){09.5}} \put(45,25){\circle*{1.0}}

\put(25,15){\vector(1,0){09.5}} \put(35,15){\circle*{1.0}}

\put(10,10){\vector(3,1){14.7}} \put(25,15){\circle*{1.0}}

%------------------ Initial point

\put(10,10){\circle*{1.4}}

\put(02.5,15){\makebox(0,0)[bl]{Initial}}
\put(02.5,12){\makebox(0,0)[bl]{point}}
\put(05,09){\makebox(0,0)[bl]{\(h^{0}\)}}

\end{picture}
%\end{center}
%
%\begin{center}
\begin{picture}(76,54)
%\begin{picture}(79,54)
\put(03.5,00){\makebox(0,0)[bl]{Fig. 5.  Multi-goal, multi-route
 DM  problem}}

\put(03.5,48){\makebox(0,0)[bl]{``Design/solving space''}}

\put(38,30){\oval(74,46)}

%------------------ Goal point
\put(65,45){\oval(04,15)}

\put(65,50){\circle{1.5}} \put(65,50){\circle*{0.8}}

\put(65,45){\circle{1.7}} \put(65,45){\circle*{1.2}}

\put(65,40){\circle{2.1}} \put(65,40){\circle*{1.4}}

\put(63,34){\makebox(0,0)[bl]{Goal}}
\put(63,31){\makebox(0,0)[bl]{points}}

\put(67.5,48.5){\makebox(0,0)[bl]{\(h^{g}_{1}\)}}
\put(67.5,43.5){\makebox(0,0)[bl]{\(h^{g}_{2}\)}}
\put(67.5,38.5){\makebox(0,0)[bl]{\(h^{g}_{3}\)}}

%------------ Path
%------------------- Route 1 (Top)

\put(14,41){\makebox(0,0)[bl]{Route}}
\put(17,37.5){\makebox(0,0)[bl]{\(L_{1}\)}}

\put(45,50){\vector(1,0){19.5}}

\put(40,45){\vector(1,1){04.5}} \put(45,50){\circle*{1.0}}

\put(25,45){\vector(1,0){14.5}} \put(40,45){\circle*{1.0}}

\put(25,40){\vector(0,1){4.5}} \put(25,45){\circle*{1.0}}

\put(20,35){\vector(1,1){04.5}} \put(25,40){\circle*{1.0}}

\put(20,25){\vector(0,1){09.5}} \put(20,35){\circle*{1.0}}

\put(10,10){\vector(2,3){9.7}} \put(20,25){\circle*{1.0}}

%------------------- Route 2 (middle)

\put(25.3,34){\makebox(0,0)[bl]{Route}}
\put(28.5,30.5){\makebox(0,0)[bl]{\(L_{2}\)}}

\put(55,45){\vector(1,0){09.5}}

\put(45,40){\vector(2,1){9.5}} \put(55,45){\circle*{1.0}}

\put(35,40){\vector(1,0){9.5}} \put(45,40){\circle*{1.0}}

\put(35,30){\vector(0,1){09.5}} \put(35,40){\circle*{1.0}}

\put(25,20){\vector(1,1){09.5}} \put(35,30){\circle*{1.0}}

\put(10,10){\vector(3,2){14.7}} \put(25,20){\circle*{1.0}}

%------------------- Route 3 (bottom)

\put(45,21){\makebox(0,0)[bl]{Route}}
\put(48.5,17.5){\makebox(0,0)[bl]{\(L_{3}\)}}

%\put(60,40){\vector(1,2){04.5}}

\put(55,30){\vector(1,1){9.5}}

% \put(60,40){\circle*{1.0}}

\put(55,25){\vector(0,1){4.5}} \put(55,30){\circle*{1.0}}

\put(45,25){\vector(1,0){09.5}} \put(55,25){\circle*{1.0}}

\put(35,15){\vector(1,1){09.5}} \put(45,25){\circle*{1.0}}

\put(25,15){\vector(1,0){09.5}} \put(35,15){\circle*{1.0}}

\put(10,10){\vector(3,1){14.7}} \put(25,15){\circle*{1.0}}

%------------------ Initial point

\put(10,10){\circle*{1.4}}

\put(02.5,15){\makebox(0,0)[bl]{Initial}}
\put(02.5,12){\makebox(0,0)[bl]{point}}
\put(05,09){\makebox(0,0)[bl]{\(h^{0}\)}}

\end{picture}
\end{center}
 Illustration for multi-route DM problem with route changes is depicted
 in Fig. 6.
 This approach can be useful in case of online taking into account external environment
  and changing the solution.
 Thus, the solving framework is the following:

~~

 {\it Stage 1 (preliminary).}
 Analysis of initial environment (i.e., situation) and
 obtaining
 several solutions (paths).

 {\it Stage 2 (start).} Selection of the best path and its execution.

 {\it Stage 3 (intermediate).}
 Analysis of external environment (i.e., change of the situation).
 Change the path  (if it needed):
 (i) selection of the best of the solution (path),
 (ii) movement to the selected path.

 {\it Stage 4.} Stop.

~~

 Note, some basic route DM problems
 (and their variants)
  are well-known, for example
 (Fig. 1):
 (a) minimum spanning tree problems
 (e.g.,
 \cite{cormen01,gar79,goemans93,hwang92,pettie02,winter87});
 (b) traveling salesman problem
 (e.g., \cite{cormen01,gar79,gutin06,papa82});
 (c) longest path problem
 (e.g., \cite{cormen01,gar79,karg97,zhang07});
 (d) maximum leafs spanning tree problem
  (e.g., \cite{alon09,gar79,klei91});
 (e) vehicle routing problem (VRP)
 (e.g., \cite{arch08,boff95,lap09,meng09,toth01,van11a}).

 Further, the  orienteering problem
% (team orienteering problem)
 and its modifications will be used as  basic
 ones (main applied domains: logistics, sport, tourism)
 (e.g., \cite{arch07,chao96a,chao96b,gav14,gav14b,gav15,gav15a,golden87,souf08,souf10,van11,van11a})
 (Fig. 7).
 In fact, the problem integrates knapsack problem and TSP.
 Here, graph \(G = (H,E)\) (\(|H|=n\)) is given,
 each vertex \(h\in H\) has a nonnegative score (profit) \(\theta(h)\),
% (\(\theta(h) \geq 0 ~ \forall h \in H\)),
 each edge/arc \(e \in E\) has a nonnegative length (cost, travel time) \(\lambda(e)\).
 The problem is
 (e.g., \cite{arch07,chao96a,chao96b,gav14,gav14b,gav15,gav15a,golden87,van11a}):

~~

 Find a route
 (a path from the start point \(h^{0} \in H\)  to the end point \(h^{g} \in H\))
 over a subset of the most important graph vertices
 that maximizes the sum of the scores of the selected vertices
 while taking into account a constraint for route length (cost)
 (i.e., combination of knapsack problem and TSP).

~~

 The mathematical model is formulated as follows:
 \(H = \{1,...,i,...,n\}\) is the set of vertex/nodes,
 vertex \(1\) is the start point of the route,
  vertex \(1\) is the end (goal) point of the route,
  binary variable \(x_{ij} =1\) if the built route (path) contains arc
  \((i,j)\)
  and \(x_{ij} = 0\) otherwise
  (vertex \(i\) precedes \(j\)),
 \(\theta_{i}\) is the vertex profit,
 \(\lambda_{ij}\) is the arc  cost (if arc \((i,j) \in E\)),
 (\(d\) is a distance constraint for the built path).
 The model is:
 \[\max \sum_{i=1}^{n} \sum_{j=1}^{n} ~\theta_{i} x_{ij}  \]
 \[s.t. ~\sum_{j=2}^{n} ~x_{1j}  = \sum_{i=1}^{n-1} ~x_{in}  =1; ~~~
 \sum_{i=2}^{n-1} ~x_{ik}  = \sum_{j=2}^{n-1} ~x_{kj} \leq 1, ~
 k=\overline{2,n-1}; ~~~
 \sum_{i=1}^{n} \sum_{j=1}^{n} ~\lambda_{ij} x_{ij} \leq d;\]
 \[x_{ij} \in \{0,1\}, ~ i = \overline{1,n}, ~ j=\overline{1,n}.\]
 The problem is NP-hard \cite{golden87}.
 Multicriteria problem statement can be examined as well
 (e.g., the score of each vertex is a vector estimate
 and the objective function is a vector based on
 the score components summarization).

\begin{center}
%\begin{picture}(76,54)
\begin{picture}(82,54)
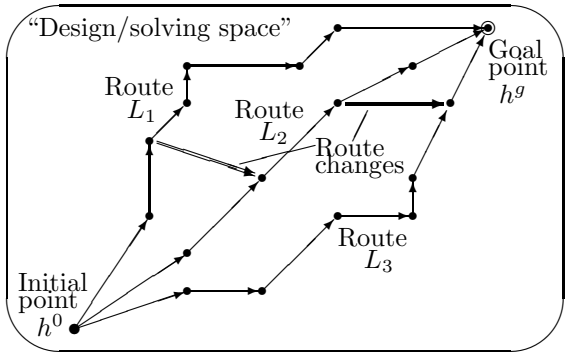
\put(00,00){\makebox(0,0)[bl]{Fig. 6. Multi-route  DM
 problem  (route change)}}

\put(03.5,48){\makebox(0,0)[bl]{``Design/solving space''}}

\put(38,30){\oval(74,46)}

%------------------ Goal point

\put(65,50){\circle{1.7}} \put(65,50){\circle*{1.2}}

\put(65,46){\makebox(0,0)[bl]{Goal}}
\put(65,43){\makebox(0,0)[bl]{point}}
\put(66.5,40){\makebox(0,0)[bl]{\(h^{g}\)}}

%------------ Path
%------------------- Route 1 (Top)

\put(14,41){\makebox(0,0)[bl]{Route}}
\put(17,37.5){\makebox(0,0)[bl]{\(L_{1}\)}}

\put(45,50){\vector(1,0){19.5}}

\put(40,45){\vector(1,1){04.5}} \put(45,50){\circle*{1.0}}

\put(25,45){\vector(1,0){14.5}} \put(40,45){\circle*{1.0}}

\put(25,40){\vector(0,1){4.5}} \put(25,45){\circle*{1.0}}

\put(20,35){\vector(1,1){04.5}} \put(25,40){\circle*{1.0}}

%-- Route change

\put(21,35){\vector(3,-1){013}} \put(21,34.7){\vector(3,-1){013}}

%--

\put(20,25){\vector(0,1){09.5}} \put(20,35){\circle*{1.0}}

\put(10,10){\vector(2,3){9.7}} \put(20,25){\circle*{1.0}}

%------------------- Route 2 (middle)

\put(31.3,38){\makebox(0,0)[bl]{Route}}
\put(34.5,34.5){\makebox(0,0)[bl]{\(L_{2}\)}}

\put(55,45){\vector(2,1){09.5}}

\put(45,40){\vector(2,1){9.5}} \put(55,45){\circle*{1.0}}

%\put(35,40){\vector(1,0){9.5}} \put(45,40){\circle*{1.0}}

%-- Route change

\put(46,40){\vector(1,0){013}} \put(46,39.7){\vector(1,0){013}}

%--

\put(35,30){\vector(1,1){09.5}} \put(45,40){\circle*{1.0}}

\put(25,20){\vector(1,1){09.5}} \put(35,30){\circle*{1.0}}

\put(10,10){\vector(3,2){14.7}} \put(25,20){\circle*{1.0}}

%================ Route changes

\put(47,35){\line(1,2){2}} \put(42,34.4){\line(-4,-1){10}}

\put(42,33){\makebox(0,0)[bl]{Route}}
\put(42,30){\makebox(0,0)[bl]{changes}}

%------------------- Route 3 (bottom)

\put(45,21){\makebox(0,0)[bl]{Route}}
\put(48.5,17.5){\makebox(0,0)[bl]{\(L_{3}\)}}

\put(60,40){\vector(1,2){04.5}}

\put(55,30){\vector(1,2){4.5}} \put(60,40){\circle*{1.0}}

\put(55,25){\vector(0,1){4.5}} \put(55,30){\circle*{1.0}}

\put(45,25){\vector(1,0){09.5}} \put(55,25){\circle*{1.0}}

\put(35,15){\vector(1,1){09.5}} \put(45,25){\circle*{1.0}}

\put(25,15){\vector(1,0){09.5}} \put(35,15){\circle*{1.0}}

\put(10,10){\vector(3,1){14.7}} \put(25,15){\circle*{1.0}}

%------------------ Initial point

\put(10,10){\circle*{1.4}}

\put(02.5,15){\makebox(0,0)[bl]{Initial}}
\put(02.5,12){\makebox(0,0)[bl]{point}}
\put(05,09){\makebox(0,0)[bl]{\(h^{0}\)}}

\end{picture}
%\end{center}
%
%\begin{center}
\begin{picture}(72,54)
%\begin{picture}(79,54)
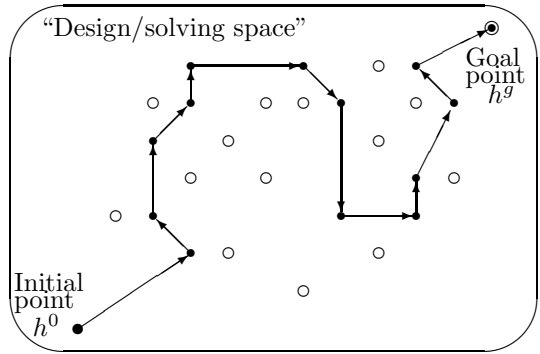
\put(08.5,00){\makebox(0,0)[bl]{Fig. 7. Team orienteering
 problem}}

\put(05,48){\makebox(0,0)[bl]{``Design/solving space''}}

\put(36,30){\oval(70,46)}

%------------------ Goal point

\put(65,50){\circle{1.7}} \put(65,50){\circle*{1.2}}

\put(61.5,45){\makebox(0,0)[bl]{Goal}}
\put(61.5,42){\makebox(0,0)[bl]{point}}
\put(64.5,40){\makebox(0,0)[bl]{\(h^{g}\)}}

%------------ Path

%\put(14,40.5){\makebox(0,0)[bl]{Route}}
%\put(10,37){\makebox(0,0)[bl]{(``physical'')}}
%\put(23,37){\line(3,-2){10}}

%------------------- Route 1 (Top)

%\put(12,41){\makebox(0,0)[bl]{Route 1}}

%\put(45,50){\vector(1,0){19.5}}

\put(40,45){\vector(1,-1){04.5}}

\put(25,45){\vector(1,0){14.5}} \put(40,45){\circle*{1.0}}

\put(25,40){\vector(0,1){4.5}} \put(25,45){\circle*{1.0}}

\put(20,35){\vector(1,1){04.5}} \put(25,40){\circle*{1.0}}

%-- Route change
% \put(21,35){\vector(3,-1){013}} \put(21,34.7){\vector(3,-1){013}}
%--

\put(20,25){\vector(0,1){09.5}} \put(20,35){\circle*{1.0}}

%\put(10,10){\vector(2,3){9.7}}

\put(20,25){\circle*{1.0}}

%------------------- Route 2 (middle)

%\put(30.3,38){\makebox(0,0)[bl]{Route 2}}

%\put(55,45){\vector(2,1){09.5}}

%\put(45,40){\vector(2,1){9.5}} \put(55,45){\circle*{1.0}}

%\put(35,40){\vector(1,0){9.5}} \put(45,40){\circle*{1.0}}

%-- Route change
% \put(46,40){\vector(1,0){013}} \put(46,39.7){\vector(1,0){013}}
%--

\put(45,40){\vector(0,-1){014.5}}

\put(45,40){\circle*{1.0}}

\put(25,20){\vector(-1,1){04.5}}

\put(10,10){\vector(3,2){14.7}} \put(25,20){\circle*{1.0}}

%================ Route changes
%\put(47,35){\line(1,2){2}} \put(42,34.4){\line(-4,-1){10}}
%\put(42,33){\makebox(0,0)[bl]{Route}}
%\put(42,30){\makebox(0,0)[bl]{changes}}

%------------------- Route 3 (bottom)

%\put(45,21){\makebox(0,0)[bl]{Route 3}}

\put(55,45){\vector(2,1){09.5}}

\put(60,40){\vector(-1,1){04.5}}  \put(55,45){\circle*{1.0}}

\put(55,30){\vector(1,2){4.5}} \put(60,40){\circle*{1.0}}

\put(55,25){\vector(0,1){4.5}} \put(55,30){\circle*{1.0}}

\put(45,25){\vector(1,0){09.5}} \put(55,25){\circle*{1.0}}

%\put(35,15){\vector(1,1){09.5}}
  \put(45,25){\circle*{1.0}}

%\put(25,15){\vector(1,0){09.5}} \put(35,15){\circle*{1.0}}

%\put(10,10){\vector(3,1){14.7}} \put(25,15){\circle*{1.0}}

%---------------

\put(50,45){\circle{1.4}}

\put(20,40){\circle{1.4}}

\put(35,40){\circle{1.4}} \put(40,40){\circle{1.4}}
\put(55,40){\circle{1.4}}

\put(30,35){\circle{1.4}} \put(50,35){\circle{1.4}}

\put(25,30){\circle{1.4}} \put(35,30){\circle{1.4}}

\put(60,30){\circle{1.4}}

\put(15,25){\circle{1.4}}

\put(30,20){\circle{1.4}} \put(50,20){\circle{1.4}}

\put(40,15){\circle{1.4}}

%------------------ Initial point

\put(10,10){\circle*{1.4}}

\put(01.5,15){\makebox(0,0)[bl]{Initial}}
\put(01.5,12){\makebox(0,0)[bl]{point}}
\put(04,09){\makebox(0,0)[bl]{\(h^{0}\)}}

\end{picture}
\end{center}
%

%%%%%%%%%%%%%%%%%%%%%%%%%%%%%%%%%%%%%%%%%%%%%%%%%%%%%%%%%%%%
%%%%%%%%%%%%%%%%%%%%%%%%%%%%%%%%%%%%%%%%%%%%%%%%%%%%%%%%%%%%
\subsection{Design and structuring of design space}

 Building and structuring of design  spaces are crucial problems in many
 domains (e.g., engineering design, technology forecasting,
 combinatorial chemistry/drug design, management/decision making)
 (e.g., \cite{ayr69,gad14,goel92,lev98,lev06,lev15,scheubrein06,zwi69}).
 Here,
  ``design/solving  space'' is modeled as a digraph/network.
 The following possible extensions
 of ``design/solving  space''
%  as digraph/network
 are used:

%~~

%
 {\it 1.} multi-layer  structure of ``design/solving  space''
 (illustrations are depicted in Fig. 8, Fig. 9);

 {\it 2.} multi-domain case (or multi-part digraph/'network)
 (illustrations are depicted in Fig. 10, Fig. 11, Fig. 12);

 {\it 3.} combined case.

%~~

%
%
 Composite multi-domain route DM problem
 is illustrated in Fig. 10:
 composite four-domain route (trajectory)
 \(S =  L_{1} \star L_{2} \star L_{3} \star L_{4} \)
 (the design space consists of four subspaces with
% corresponding
 subsolutions).

\begin{center}
%\begin{picture}(76,54)
\begin{picture}(67,44)
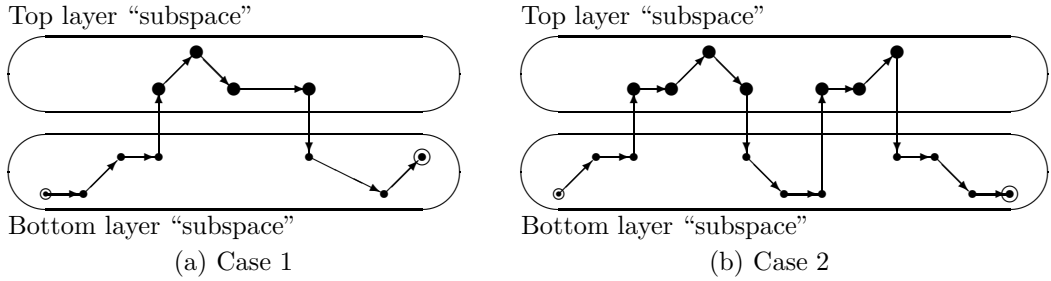
\put(34,00){\makebox(0,0)[bl]{Fig. 8. Two-layer ``design/solving
  spaces''}}

 \put(22,05){\makebox(0,0)[bl]{(a) Case 1}}

\put(0,38){\makebox(0,0)[bl]{Top layer ``subspace''}}
%\put(02,29.5){\makebox(0,0)[bl]{(domain) 3}}

\put(30,32){\oval(60,10)}   %----------------Layer 1
\put(30,19){\oval(60,10)}   %----------------Layer 2

\put(0,10){\makebox(0,0)[bl]{Bottom layer ``subspace''}}

%------------------ Goal point
\put(55,21){\circle{2.0}} \put(55,21){\circle*{1.2}}

%------------ Path

\put(50,16){\vector(1,1){04.5}}

\put(40,21){\vector(2,-1){09.5}} \put(50,16){\circle*{1.0}}
\put(40,30){\vector(0,-1){08.5}} \put(40,21){\circle*{1.0}}

\put(30,30){\vector(1,0){09.5}} \put(40,30){\circle*{1.8}}
\put(25,35){\vector(1,-1){04.5}} \put(30,30){\circle*{1.8}}
\put(20,30){\vector(1,1){04.5}} \put(25,35){\circle*{1.8}}

\put(20,21){\vector(0,1){08.5}} \put(20,30){\circle*{1.8}}
\put(15,21){\vector(1,0){04.5}} \put(20,21){\circle*{1.0}}
\put(10,16){\vector(1,1){04.5}} \put(15,21){\circle*{1.0}}
\put(05,16){\vector(1,0){04.5}} \put(10,16){\circle*{1.0}}

%------------------ Initial point
\put(05,16){\circle*{0.8}} \put(05,16){\circle{1.4}}

\end{picture}
%\end{center}
%
%\begin{center}
\begin{picture}(70,44)
%\put(19,00){\makebox(0,0)[bl]{Fig. 4. Two-layer design/solving
%  spaces}}

 \put(25,05){\makebox(0,0)[bl]{(b) Case 2}}

\put(0,38){\makebox(0,0)[bl]{Top layer ``subspace''}}
%\put(02,29.5){\makebox(0,0)[bl]{(domain) 3}}

\put(35,32){\oval(70,10)}   %----------------Layer 1
\put(35,19){\oval(70,10)}   %----------------Layer 2

\put(0,10){\makebox(0,0)[bl]{Bottom layer ``subspace''}}

%------------------ Goal point
\put(65,16){\circle{2.0}} \put(65,16){\circle*{1.2}}

%------------ Path
\put(60,16){\vector(1,0){04.5}}

\put(55,21){\vector(1,-1){04.5}} \put(60,16){\circle*{1.0}}
\put(50,21){\vector(1,0){04.5}} \put(55,21){\circle*{1.0}}
\put(50,35){\vector(0,-1){013.5}} \put(50,21){\circle*{1.0}}

\put(45,30){\vector(1,1){04.5}} \put(50,35){\circle*{1.8}}
\put(40,30){\vector(1,0){04.5}} \put(45,30){\circle*{1.8}}
\put(40,16){\vector(0,1){13.5}} \put(40,30){\circle*{1.8}}

\put(35,16){\vector(1,0){04.5}} \put(40,16){\circle*{1.0}}
\put(30,21){\vector(1,-1){04.5}} \put(35,16){\circle*{1.0}}
\put(30,30){\vector(0,-1){08.5}} \put(30,21){\circle*{1.0}}

\put(25,35){\vector(1,-1){04.5}} \put(30,30){\circle*{1.8}}
\put(20,30){\vector(1,1){04.5}} \put(25,35){\circle*{1.8}}
\put(15,30){\vector(1,0){04.5}} \put(20,30){\circle*{1.8}}

\put(15,21){\vector(0,1){08.5}} \put(15,30){\circle*{1.8}}
\put(10,21){\vector(1,0){04.5}} \put(15,21){\circle*{1.0}}
\put(05,16){\vector(1,1){04.5}} \put(10,21){\circle*{1.0}}

%------------------ Initial point
\put(05,16){\circle*{0.8}} \put(05,16){\circle{1.4}}

\end{picture}
\end{center}

\begin{center}
\begin{picture}(73,42)
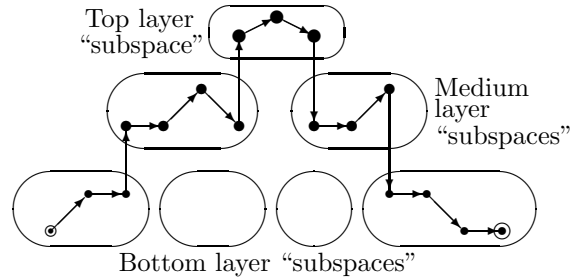
\put(02,00){\makebox(0,0)[bl]{Fig. 9. Three-layer ``design/solving
 space''}}

%----------------%-- Layer 1 top
\put(09.5,37.5){\makebox(0,0)[bl]{Top layer}}
\put(08.5,34){\makebox(0,0)[bl]{``subspace''}}

\put(35,37.5){\oval(18,07)}

%----------------- Layer 2 medium
\put(56,29){\makebox(0,0)[bl]{Medium}}
\put(56,25.5){\makebox(0,0)[bl]{layer}}
\put(56,22){\makebox(0,0)[bl]{``subspaces''}}

\put(22.5,27){\oval(20,10)} \put(46,27){\oval(18,10)}

%---------------%-- Layer 3 bottom

\put(09,14){\oval(18,10)} \put(26.5,14){\oval(14,10)}

\put(40,14){\oval(10,10)} \put(58,14){\oval(23,10)}

\put(14,05){\makebox(0,0)[bl]{Bottom layer ``subspaces''}}

%------------------ Goal point
\put(65,11){\circle{2.0}} \put(65,11){\circle*{1.2}}

%------------ Path
\put(60,11){\vector(1,0){04.5}}

\put(55,16){\vector(1,-1){04.5}} \put(60,11){\circle*{1.0}}
\put(50,16){\vector(1,0){04.5}} \put(55,16){\circle*{1.0}}
\put(50,30){\vector(0,-1){013.5}} \put(50,16){\circle*{1.0}}

\put(45,25){\vector(1,1){04.5}} \put(50,30){\circle*{1.5}}
\put(40,25){\vector(1,0){04.5}} \put(45,25){\circle*{1.5}}
\put(40,37){\vector(0,-1){11.5}} \put(40,25){\circle*{1.5}}

\put(35,39.5){\vector(2,-1){04.5}} \put(40,37){\circle*{1.8}}
\put(30,37){\vector(2,1){04.5}} \put(35,39.5){\circle*{1.8}}
\put(30,25){\vector(0,1){11.5}} \put(30,37){\circle*{1.8}}

\put(25,30){\vector(1,-1){04.5}} \put(30,25){\circle*{1.5}}
\put(20,25){\vector(1,1){04.5}} \put(25,30){\circle*{1.5}}
\put(15,25){\vector(1,0){04.5}} \put(20,25){\circle*{1.5}}

\put(15,16){\vector(0,1){08.5}} \put(15,25){\circle*{1.5}}
\put(10,16){\vector(1,0){04.5}} \put(15,16){\circle*{1.0}}
\put(05,11){\vector(1,1){04.5}} \put(10,16){\circle*{1.0}}

%------------------ Initial point
\put(05,11){\circle*{0.8}} \put(05,11){\circle{1.4}}

\end{picture}
\end{center}

 In addition, it  may be reasonable to consider
 this multi-domain routing DM problem
 with coordination of the subsolutions/routes
 (i.e., \(L_{1}, L_{2},...\))
 at each intermediate time moment
 (\(\{\tau^{1},\tau^{2},...,\tau^{l}\}\))
 between start moment \(\tau^{0}\) and end moment \(\tau^{g}\):~
 \(\tau^{0} < \tau^{1} < \tau^{2} <...< \tau^{l} < \tau^{g}\).
 Fig. 11 illustrates the coordination of four routes
 (\(\{L_{1},L_{2},L_{3},L_{4}  \}\))
  at two time moments:

 ~(a) coordination 1 at time moment  \(\tau'\):~
  points \(h'_{1}\),\(h'_{2}\),\(h'_{3}\),\(h'_{4}\);

 ~(b) coordination 2 at time moment \(\tau'' > \tau' \):~
  points \(h''_{1}\),\(h''_{2}\),\(h''_{3}\),\(h''_{4}\).

 Three-domain route/trajectory DM problem
 based on different basic combinatorial optimization route problems
 (TSP,  orienteering problem, shortest path)
 is depicted in Fig. 12.

\begin{center}
\begin{picture}(50,71)
\put(19,00){\makebox(0,0)[bl]{Fig. 10. Four-domain route
 DM problem}}

\put(02,33){\makebox(0,0)[bl]{``Design space''}}
\put(02,29.5){\makebox(0,0)[bl]{(domain) 3}}

\put(25,22.5){\oval(50,31)}

%------------------ Goal point

\put(40,35){\circle{1.6}} \put(40,35){\circle*{1.0}}

\put(41,32){\makebox(0,0)[bl]{Goal}}
\put(41,28.5){\makebox(0,0)[bl]{point}}
\put(44,25){\makebox(0,0)[bl]{\(h^{g}_{3}\)}}

%------------ Path

\put(22,20){\makebox(0,0)[bl]{Route}}
\put(25,16){\makebox(0,0)[bl]{\(L_{3}\)}}

\put(30,30){\vector(2,1){09.5}}

\put(30,25){\vector(0,1){04.5}} \put(30,30){\circle*{1.0}}
\put(20,25){\vector(1,0){09.5}} \put(30,25){\circle*{1.0}}
\put(20,15){\vector(0,1){09.5}} \put(20,25){\circle*{1.0}}
\put(15,15){\vector(1,0){04.5}} \put(20,15){\circle*{1.0}}
\put(10,10){\vector(1,1){4.6}} \put(15,15){\circle*{1.0}}

%------------------ Initial point

\put(10,10){\circle*{1.4}}

%\put(01.5,14){\makebox(0,0)[bl]{Initial}}
%\put(01.5,10.5){\makebox(0,0)[bl]{point}}

\put(01.5,15){\makebox(0,0)[bl]{Initial}}
\put(01.5,12){\makebox(0,0)[bl]{point}}
\put(04.5,08.5){\makebox(0,0)[bl]{\(h^{0}_{3}\)}}

%%%%%%%%%%%%%%%%%%%%%%%%%%%%%%%%%%%%%%%%%%%%%%%%%%%%%%%%%%%%%%%%%%%%%%%%

\put(02,65){\makebox(0,0)[bl]{``Design space''}}
\put(02,61.5){\makebox(0,0)[bl]{(domain) 1}}

\put(25,54.5){\oval(50,31)}

%------------------ Goal point

\put(45,67){\circle{1.7}} \put(45,67){\circle*{1.2}}

\put(40,63){\makebox(0,0)[bl]{Goal}}
\put(40,59.5){\makebox(0,0)[bl]{point}}
\put(43,56){\makebox(0,0)[bl]{\(h^{g}_{1}\)}}

%------------ Path
\put(26.5,50){\makebox(0,0)[bl]{Route}}
\put(29.5,46){\makebox(0,0)[bl]{\(L_{1}\)}}

\put(35,67){\vector(1,0){9.5}}

\put(35,62){\vector(0,1){04.5}} \put(35,67){\circle*{1.0}}
\put(25,52){\vector(1,1){09.5}} \put(35,62){\circle*{1.0}}
\put(15,52){\vector(1,0){09.5}} \put(25,52){\circle*{1.0}}
\put(10,42){\vector(1,2){4.7}} \put(15,52){\circle*{1.0}}

%------------------ Initial point

\put(10,42){\circle*{1.4}}

\put(01.5,49){\makebox(0,0)[bl]{Initial}}
\put(01.5,45.5){\makebox(0,0)[bl]{point}}
\put(03.5,42){\makebox(0,0)[bl]{\(h^{0}_{1}\)}}

\end{picture}
%\end{center}
%
%\begin{center}
\begin{picture}(50,54)
%\put(20,00){\makebox(0,0)[bl]{Fig. 6. Multi-domain route
% DM problem}}

\put(02,33){\makebox(0,0)[bl]{``Design space''}}
\put(02,29.5){\makebox(0,0)[bl]{(domain) 4}}

\put(25,22.5){\oval(50,31)}

%------------------ Goal point

\put(45,35){\circle{1.7}} \put(45,35){\circle*{1.2}}

\put(40,31){\makebox(0,0)[bl]{Goal}}
\put(40,27){\makebox(0,0)[bl]{point}}
\put(43,23.5){\makebox(0,0)[bl]{\(h^{g}_{4}\)}}

%------------ Path

\put(25,16){\makebox(0,0)[bl]{Route}}
\put(28,12){\makebox(0,0)[bl]{\(L_{4}\)}}

\put(40,35){\vector(1,0){4.5}}

\put(35,30){\vector(1,1){04.5}} \put(40,35){\circle*{1.0}}
\put(35,20){\vector(0,1){09.5}} \put(35,30){\circle*{1.0}}
\put(25,20){\vector(1,0){09.5}} \put(35,20){\circle*{1.0}}
\put(20,15){\vector(1,1){04.5}} \put(25,20){\circle*{1.0}}
\put(20,10){\vector(0,1){04.5}} \put(20,15){\circle*{1.0}}
\put(10,10){\vector(1,0){9.5}} \put(20,10){\circle*{1.0}}

%------------------ Initial point

\put(10,10){\circle*{1.4}}

%\put(01.5,14){\makebox(0,0)[bl]{Initial}}
%\put(01.5,10.5){\makebox(0,0)[bl]{point}}

\put(01.5,15){\makebox(0,0)[bl]{Initial}}
\put(01.5,12){\makebox(0,0)[bl]{point}}
\put(04.5,08.5){\makebox(0,0)[bl]{\(h^{0}_{4}\)}}

%%%%%%%%%%%%%%%%%%%%%%%%%%%%%%%%%%%%%%%%%%%%%%%%%%%%%%%%%%%%%%%%%%%%%%%%

\put(02,65){\makebox(0,0)[bl]{``Design space''}}
\put(02,61.5){\makebox(0,0)[bl]{(domain) 2}}

\put(25,54.5){\oval(50,31)}

%------------------ Goal point

\put(35,67){\circle{1.7}} \put(35,67){\circle*{1.2}}

\put(37,66.5){\makebox(0,0)[bl]{Goal}}
\put(37,63){\makebox(0,0)[bl]{point}}
\put(40,59.5){\makebox(0,0)[bl]{\(h^{g}_{2}\)}}

%------------ Path

\put(26.5,50){\makebox(0,0)[bl]{Route}}
\put(29.5,46){\makebox(0,0)[bl]{\(L_{2}\)}}

\put(35,62){\vector(0,1){04.5}}

\put(25,52){\vector(1,1){09.5}} \put(35,62){\circle*{1.0}}
\put(20,52){\vector(1,0){04.5}} \put(25,52){\circle*{1.0}}
\put(20,47){\vector(0,1){04.5}} \put(20,52){\circle*{1.0}}
\put(10,42){\vector(2,1){9.5}} \put(20,47){\circle*{1.0}}

%------------------ Initial point

\put(10,42){\circle*{1.4}}

\put(01.5,49){\makebox(0,0)[bl]{Initial}}
\put(01.5,45.5){\makebox(0,0)[bl]{point}}
\put(03.5,42){\makebox(0,0)[bl]{\(h^{0}_{2}\)}}

\end{picture}
\end{center}

\begin{center}
%\begin{picture}(76,54)
\begin{picture}(50,71)
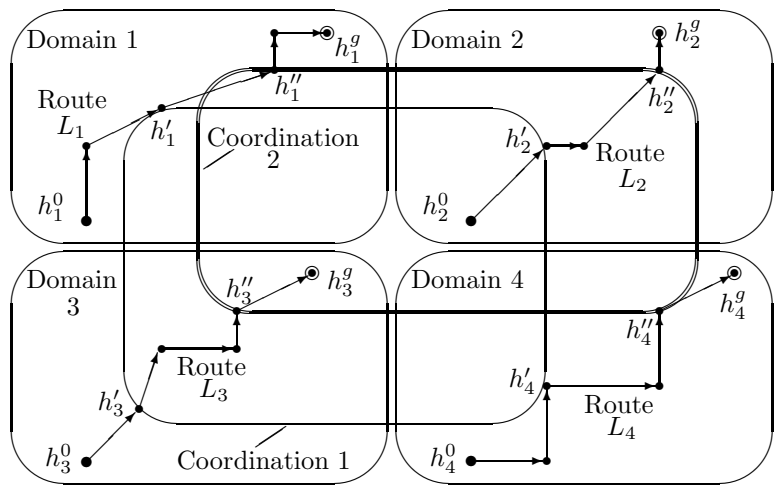
\put(04.5,00){\makebox(0,0)[bl]{Fig. 11. Four-domain route
 DM problem with coordination}}

%============== Coordinations

%--------- 2nd coordination

\put(58,46){\oval(66,32)} \put(58,46){\oval(66.6,32.6)}

%--------- 1st coordination
\put(43,36){\oval(56,42)}

%===========================

\put(02,33){\makebox(0,0)[bl]{Domain}}
\put(07.4,29.5){\makebox(0,0)[bl]{3}}

\put(25,22.5){\oval(50,31)}

%------------------ Goal point

\put(40,35){\circle{1.6}} \put(40,35){\circle*{1.0}}

%\put(41,32){\makebox(0,0)[bl]{Goal}}
%\put(41,28.5){\makebox(0,0)[bl]{point}}
\put(42,32){\makebox(0,0)[bl]{\(h^{g}_{3}\)}}

%------------ Path

\put(32.5,12){\line(3,2){04}}
\put(22,09){\makebox(0,0)[bl]{Coordination 1}}

\put(22,21.5){\makebox(0,0)[bl]{Route}}
\put(25,18){\makebox(0,0)[bl]{\(L_{3}\)}}

\put(28.5,31){\makebox(0,0)[bl]{\(h''_{3}\)}}
\put(11.5,16){\makebox(0,0)[bl]{\(h'_{3}\)}}

\put(30,30){\vector(2,1){09.5}}

\put(30,25){\vector(0,1){04.5}} \put(30,30){\circle*{1.0}}
\put(20,25){\vector(1,0){09.5}} \put(30,25){\circle*{1.0}}

\put(17,17){\vector(1,3){02.5}} \put(20,25){\circle*{1.0}}

%\put(15,15){\vector(1,0){04.5}} \put(20,15){\circle*{1.0}}

\put(10,10){\vector(1,1){6.5}} \put(17,17){\circle*{1.0}}

%------------------ Initial point

\put(10,10){\circle*{1.4}}

%\put(01.5,14){\makebox(0,0)[bl]{Initial}}
%\put(01.5,10.5){\makebox(0,0)[bl]{point}}

%\put(01.5,15){\makebox(0,0)[bl]{Initial}}
%\put(01.5,12){\makebox(0,0)[bl]{point}}
\put(04.5,08.5){\makebox(0,0)[bl]{\(h^{0}_{3}\)}}

%%%%%%%%%%%%%%%%%%%%%%%%%%%%%%%%%%%%%%%%%%%%%%%%%%%%%%%%%%%%%%%%%%%%%%%%

\put(02,65){\makebox(0,0)[bl]{Domain 1}}
%\put(02,61.5){\makebox(0,0)[bl]{(domain) 1}}

\put(25,54.5){\oval(50,31)}

%------------------ Goal point

\put(42,67){\circle{1.7}} \put(42,67){\circle*{1.2}}

%\put(40,63){\makebox(0,0)[bl]{Goal}}
%\put(40,59.5){\makebox(0,0)[bl]{point}}
\put(43,63){\makebox(0,0)[bl]{\(h^{g}_{1}\)}}

%------------ Path
\put(3.5,57){\makebox(0,0)[bl]{Route}}
\put(6,53){\makebox(0,0)[bl]{\(L_{1}\)}}

\put(29.6,51.5){\line(-3,-2){04}}
\put(26,52){\makebox(0,0)[bl]{Coordination}}
\put(34,49){\makebox(0,0)[bl]{2}}

\put(35,67){\vector(1,0){6.5}}

\put(35,62){\vector(0,1){04.5}} \put(35,67){\circle*{1.0}}

\put(20,57){\vector(3,1){14.5}} \put(35,62){\circle*{1.0}}

\put(35,58){\makebox(0,0)[bl]{\(h''_{1}\)}}
\put(18.5,52){\makebox(0,0)[bl]{\(h'_{1}\)}}

\put(10,52){\vector(2,1){09.5}} \put(20,57){\circle*{1.0}}

\put(10,42){\vector(0,1){9.5}} \put(10,52){\circle*{1.0}}

%------------------ Initial point

\put(10,42){\circle*{1.4}}

%\put(01.5,49){\makebox(0,0)[bl]{Initial}}
%\put(01.5,45.5){\makebox(0,0)[bl]{point}}
\put(03.5,42){\makebox(0,0)[bl]{\(h^{0}_{1}\)}}

\end{picture}
%\end{center}
%
%\begin{center}
\begin{picture}(50,54)
%\put(20,00){\makebox(0,0)[bl]{Fig. 7. Multi-domain route
% DM problem}}

\put(02,33){\makebox(0,0)[bl]{Domain 4}}
%\put(02,29.5){\makebox(0,0)[bl]{(domain) 4}}

\put(25,22.5){\oval(50,31)}

%------------------ Goal point

\put(45,35){\circle{1.7}} \put(45,35){\circle*{1.2}}

%\put(40,31){\makebox(0,0)[bl]{Goal}}
%\put(40,27){\makebox(0,0)[bl]{point}}
\put(43,29){\makebox(0,0)[bl]{\(h^{g}_{4}\)}}

%------------ Path

\put(25,16.5){\makebox(0,0)[bl]{Route}}
\put(28,13){\makebox(0,0)[bl]{\(L_{4}\)}}

\put(30.5,25.5){\makebox(0,0)[bl]{\(h''_{4}\)}}
\put(15,19){\makebox(0,0)[bl]{\(h'_{4}\)}}

%\put(40,35){\vector(1,0){4.5}}

\put(35,30){\vector(2,1){09.5}}

\put(35,20){\vector(0,1){09.5}} \put(35,30){\circle*{1.0}}

\put(20,20){\vector(1,0){014.5}} \put(35,20){\circle*{1.0}}

\put(20,10){\vector(0,1){09.5}} \put(20,20){\circle*{1.0}}
\put(10,10){\vector(1,0){9.5}} \put(20,10){\circle*{1.0}}

%------------------ Initial point

\put(10,10){\circle*{1.4}}

%\put(01.5,14){\makebox(0,0)[bl]{Initial}}
%\put(01.5,10.5){\makebox(0,0)[bl]{point}}

%\put(01.5,15){\makebox(0,0)[bl]{Initial}}
%\put(01.5,12){\makebox(0,0)[bl]{point}}
\put(04.5,08.5){\makebox(0,0)[bl]{\(h^{0}_{4}\)}}

%%%%%%%%%%%%%%%%%%%%%%%%%%%%%%%%%%%%%%%%%%%%%%%%%%%%%%%%%%%%%%%%%%%%%%%%

\put(02,65){\makebox(0,0)[bl]{Domain 2}}
%\put(02,61.5){\makebox(0,0)[bl]{(domain) 2}}

\put(25,54.5){\oval(50,31)}

%------------------ Goal point

\put(35,67){\circle{1.7}} \put(35,67){\circle*{1.2}}

%\put(37,66.5){\makebox(0,0)[bl]{Goal}}
%\put(37,63){\makebox(0,0)[bl]{point}}
\put(37,65){\makebox(0,0)[bl]{\(h^{g}_{2}\)}}

\put(33.5,56.4){\makebox(0,0)[bl]{\(h''_{2}\)}}
\put(14.5,51){\makebox(0,0)[bl]{\(h'_{2}\)}}

%------------ Path

\put(26.5,50){\makebox(0,0)[bl]{Route}}
\put(29.5,46){\makebox(0,0)[bl]{\(L_{2}\)}}

\put(35,62){\vector(0,1){04.5}}

\put(25,52){\vector(1,1){09.5}} \put(35,62){\circle*{1.0}}
\put(20,52){\vector(1,0){04.5}} \put(25,52){\circle*{1.0}}

%\put(20,47){\vector(0,1){04.5}}

%\put(20,52){\circle*{1.0}}

\put(10,42){\vector(1,1){9.5}} \put(20,52){\circle*{1.0}}

%------------------ Initial point

\put(10,42){\circle*{1.4}}

%\put(01.5,49){\makebox(0,0)[bl]{Initial}}
%\put(01.5,45.5){\makebox(0,0)[bl]{point}}
\put(03.5,42){\makebox(0,0)[bl]{\(h^{0}_{2}\)}}

\end{picture}
\end{center}

\begin{center}
%\begin{picture}(76,54)
\begin{picture}(30,55)
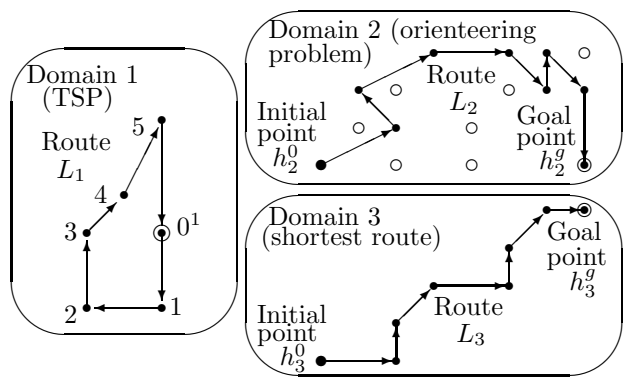
\put(07,00){\makebox(0,0)[bl]{Fig. 12. Three-domain route
 DM problem}}
% (different basic combinatorial problems)}}

\put(02,45){\makebox(0,8)[bl]{Domain \(1\)}}
\put(04,41){\makebox(0,8)[bl]{(TSP)}}

\put(15,30.5){\oval(30,38)}

\put(04,36){\makebox(0,0)[bl]{Route}}
\put(06,31.8){\makebox(0,0)[bl]{\(L_{1}\)}}

%  Center-depot
%\put(20,31.5){\makebox(0,8)[bl]{Depot (center)}}

\put(22,24.7){\makebox(0,8)[bl]{\(0^{1}\)}}

\put(20,25){\circle*{1}} \put(20,25){\circle{2}}

% Points and vectors

\put(21,14){\makebox(0,8)[bl]{\(1\)}}
\put(20,25){\vector(0,-1){09}} \put(20,15){\circle*{1}}

\put(07,13){\makebox(0,8)[bl]{\(2\)}}
\put(20,15){\vector(-1,0){09}} \put(10,15){\circle*{1}}

\put(07,24){\makebox(0,8)[bl]{\(3\)}}
\put(10,15){\vector(0,1){09}} \put(10,25){\circle*{1}}

\put(11,29){\makebox(0,8)[bl]{\(4\)}}
\put(10,25){\vector(1,1){04}} \put(15,30){\circle*{1}}

\put(16,38){\makebox(0,8)[bl]{\(5\)}}
\put(15,30){\vector(1,2){04.5}} \put(20,40){\circle*{1}}
\put(20,40){\vector(0,-1){14}}

\end{picture}
%\end{center}
%
%\begin{center}
\begin{picture}(50,55)
%\put(20,00){\makebox(0,0)[bl]{Fig. 8. Three-domain route
% DM problem}}

\put(03,26){\makebox(0,0)[bl]{Domain 3}}
\put(02,22.5){\makebox(0,0)[bl]{(shortest route)}}

\put(25,18){\oval(50,24)}

%------------------ Goal point

\put(45,28){\circle{1.7}} \put(45,28){\circle*{1.2}}

\put(40,24){\makebox(0,0)[bl]{Goal}}
\put(40,20){\makebox(0,0)[bl]{point}}
\put(43,16.5){\makebox(0,0)[bl]{\(h^{g}_{3}\)}}

%------------ Path

\put(25,14){\makebox(0,0)[bl]{Route}}
\put(28,10){\makebox(0,0)[bl]{\(L_{3}\)}}

\put(40,28){\vector(1,0){4.5}}

\put(35,23){\vector(1,1){04.5}} \put(40,28){\circle*{1.0}}
\put(35,18){\vector(0,1){04.5}} \put(35,23){\circle*{1.0}}
\put(25,18){\vector(1,0){09.5}} \put(35,18){\circle*{1.0}}
\put(20,13){\vector(1,1){04.5}} \put(25,18){\circle*{1.0}}
\put(20,08){\vector(0,1){04.5}} \put(20,13){\circle*{1.0}}
\put(10,08){\vector(1,0){9.5}} \put(20,08){\circle*{1.0}}

%------------------ Initial point

\put(10,08){\circle*{1.4}}

\put(01.5,13){\makebox(0,0)[bl]{Initial}}
\put(01.5,10){\makebox(0,0)[bl]{point}}
\put(04.5,06.5){\makebox(0,0)[bl]{\(h^{0}_{3}\)}}

%%%%%%%%%%%%%%%%%%%%%%%%%%%%%%%%%%%%%%%%%%%%%%%%%%%%%%%%%%%%%%%%%%%%%%%%

\put(03,50){\makebox(0,0)[bl]{Domain 2 (orienteering}}
\put(03,46.8){\makebox(0,0)[bl]{problem)}}

\put(25,43){\oval(50,23)}

%------------------ Goal point

\put(45,34){\circle{1.7}} \put(45,34){\circle*{1.2}}

\put(36,39.5){\makebox(0,0)[bl]{Goal}}
\put(36,36){\makebox(0,0)[bl]{point}}
\put(39,32.5){\makebox(0,0)[bl]{\(h^{g}_{2}\)}}

%------------ Path

\put(24,45){\makebox(0,0)[bl]{Route}}
\put(27,40.8){\makebox(0,0)[bl]{\(L_{2}\)}}

\put(45,44){\vector(0,-1){09.5}}

\put(40,49){\vector(1,-1){04.5}} \put(45,44){\circle*{1.0}}

\put(40,44){\vector(0,1){04.5}} \put(40,49){\circle*{1.0}}
\put(35,49){\vector(1,-1){04.5}} \put(40,44){\circle*{1.0}}

\put(25,49){\vector(1,0){09.5}} \put(35,49){\circle*{1.0}}
\put(15,44){\vector(2,1){09.5}} \put(25,49){\circle*{1.0}}
\put(20,39){\vector(-1,1){04.5}} \put(15,44){\circle*{1.0}}
\put(10,34){\vector(2,1){9.5}} \put(20,39){\circle*{1.0}}

%------------------ Initial point

\put(10,34){\circle*{1.4}}

\put(01.5,40){\makebox(0,0)[bl]{Initial}}
\put(01.5,36.5){\makebox(0,0)[bl]{point}}
\put(03.5,33){\makebox(0,0)[bl]{\(h^{0}_{2}\)}}

%---------------- Other points

\put(45,49){\circle{1.4}}

\put(20,44){\circle{1.4}} \put(35,44){\circle{1.4}}

\put(15,39){\circle{1.4}} \put(30,39){\circle{1.4}}

\put(30,34){\circle{1.4}} \put(20,34){\circle{1.4}}

\end{picture}
\end{center}

 An illustration for two-layer multi-domain routing
 in communication (sensor) systems is depicted in Fig. 13:
 from
 sender node (origin)
% (source)
 \(h^{0}\) to
 goal nodes (destinations)
 \(h^{g}_{31}\), \(h^{g}_{32}\), \(h^{g}_{33}\),
 \(h^{g}_{41}\), \(h^{g}_{42}\),
 \(h^{g}_{51}\), \(h^{g}_{52}\), \(h^{g}_{53}\).
 In general,
 a certain routing problem can be used in each domain
 (e.g., the shortest path, minimum spanning tree).

\begin{center}
\begin{picture}(116,43)
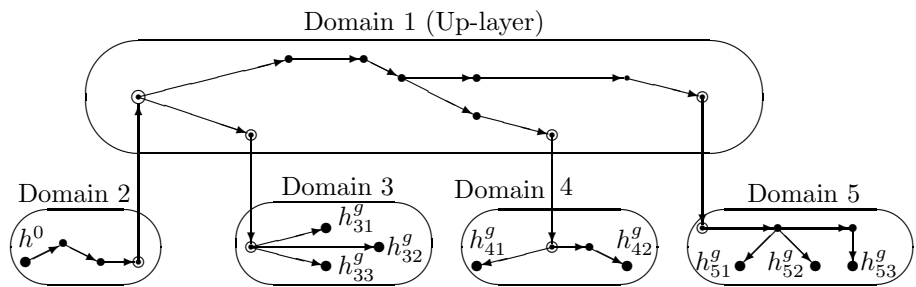
\put(08,00){\makebox(0,0)[bl]{Fig. 13. Illustration of routing
 in two-layer communication system}}

%--------------------- Bottom domain 2 start
\put(01,16){\makebox(0,0)[bl]{Domain 2}}

\put(10,10){\oval(20,10)}

\put(01,10){\makebox(0,0)[bl]{\(h^{0}\)}}

\put(02,08){\circle*{1.4}}

\put(02,08){\vector(2,1){4.5}}    \put(07,10.5){\circle*{0.9}}
\put(07,10.5){\vector(2,-1){4.5}} \put(12,08){\circle*{0.9}}
\put(12,08){\vector(1,0){4.5}}

\put(17,08){\circle*{0.8}} \put(17,08){\circle{1.5}}

\put(17,08){\vector(0,1){21}}

%------------------ Bottom domain 3
\put(36,17){\makebox(0,0)[bl]{Domain 3}}

\put(43,10.5){\oval(26,11)}

\put(32,10){\circle*{0.8}} \put(32,10){\circle{1.5}}

\put(32,10){\vector(4,1){9.5}} \put(42,12.5){\circle*{1.4}}
\put(43.5,12){\makebox(0,0)[bl]{\(h^{g}_{31}\)}}

\put(32,10){\vector(4,-1){9.5}}  \put(42,07.5){\circle*{1.4}}
\put(43.5,05.5){\makebox(0,0)[bl]{\(h^{g}_{33}\)}}

\put(32,10){\vector(1,0){16.5}}  \put(49,10){\circle*{1.4}}
\put(50,08){\makebox(0,0)[bl]{\(h^{g}_{32}\)}}

%-------------------- Bottom 4
\put(59,16.5){\makebox(0,0)[bl]{Domain}}
\put(73,17){\makebox(0,0)[bl]{4}}

\put(73,10){\oval(26,10)}

%\put(61,10){\makebox(0,0)[bl]{\(h^{0}\)}}

\put(72,10){\circle*{0.8}} \put(72,10){\circle{1.5}}
\put(72,10){\vector(-4,-1){9.5}}    \put(62,07.5){\circle*{1.4}}

\put(61,09){\makebox(0,0)[bl]{\(h^{g}_{41}\)}}

\put(72,10){\vector(1,0){4.5}} \put(77,10){\circle*{0.9}}
\put(77,10){\vector(2,-1){4.5}}    \put(82,07.5){\circle*{1.4}}

\put(80.5,09){\makebox(0,0)[bl]{\(h^{g}_{42}\)}}

%----------------- Bottom domain 5
\put(98,16){\makebox(0,0)[bl]{Domain 5}}

\put(105,10){\oval(30,10)}

\put(91,06){\makebox(0,0)[bl]{\(h^{g}_{51}\)}}
\put(100.6,06){\makebox(0,0)[bl]{\(h^{g}_{52}\)}}
\put(113,06){\makebox(0,0)[bl]{\(h^{g}_{53}\)}}

\put(92,12.5){\circle{1.5}} \put(92,12.5){\circle*{0.8}}

\put(92,12.5){\vector(1,0){9.5}}    \put(102,12.5){\circle*{0.9}}

\put(102,12.5){\vector(-1,-1){4.5}} \put(97,07.5){\circle*{1.4}}
\put(102,12.5){\vector(1,-1){4.5}} \put(107,07.5){\circle*{1.4}}

\put(102,12.5){\vector(1,0){9.5}}    \put(112,12.5){\circle*{0.9}}
\put(112,12.5){\vector(0,-1){4.5}} \put(112,07.5){\circle*{1.4}}

%\put(102,08){\vector(1,0){4.5}}    \put(107,08){\circle*{0.9}}

%============================= UP layer - 1
\put(39,38){\makebox(0,0)[bl]{Domain 1 (Up-layer)}}

\put(55,30){\oval(90,15)}

%\put(17,08){\vector(0,1){20}}

\put(17,30){\circle*{0.8}} \put(17,30){\circle{1.9}}

%--------- TO Bottom 2

\put(17,30){\vector(3,-1){14.5}}

\put(32,25){\circle*{0.8}} \put(32,25){\circle{1.5}}

\put(32,25){\vector(0,-1){14.5}}

%--
\put(17,30){\vector(4,1){19.5}} \put(37,35){\circle*{0.9}}

\put(37,35){\vector(1,0){09.5}} \put(47,35){\circle*{0.9}}

\put(47,35){\vector(2,-1){04.5}} \put(52,32.5){\circle*{0.9}}

\put(52,32.5){\vector(1,0){09.5}} \put(62,32.5){\circle*{0.9}}

%----- TO Bottom 3

\put(52,32.5){\vector(2,-1){09.5}} \put(62,27.5){\circle*{0.9}}

\put(62,27.5){\vector(4,-1){09.5}}

%\put(72,25){\circle*{0.9}}

\put(72,25){\circle{1.5}} \put(72,25){\circle*{0.8}}

\put(72,25){\vector(0,-1){14.5}}

%------ TO Bottom 4
\put(62,32.5){\vector(1,0){19.5}} \put(82,32.5){\circle*{0.8}}

\put(82,32.5){\vector(4,-1){09.5}}

%\put(82,32.5){\circle*{0.8}}

\put(92,30){\circle{1.5}} \put(92,30){\circle*{0.8}}

\put(92,30){\vector(0,-1){17}}

%\put(92,12.5){\circle{1.5}} \put(92,12.5){\circle*{0.8}}

\end{picture}
\end{center}
%

%%%%%%%%%%%%%%%%%%%%%%%%%%%%%%%%%%%%%%%%%%%%%%%%%%%%%%%%%%%%
%%%%%%%%%%%%%%%%%%%%%%%%%%%%%%%%%%%%%%%%%%%%%%%%%%%%%%%%%%%%
\subsection{Types of model nodes}

% Building and structuring of design  spaces are crucial problems in many
% domains (e.g., engineering design, technology forecasting,
% combinatorial chemistry/drug design, management/decision making)
% (e.g., \cite{ayr69,gad14,goel92,lev98,lev06,lev15,scheubrein06,zwi69}).

 In this material,
  ``design/solving  space'' is modeled as a digraph/network.
 Here, our types of elements (i.e., model nodes/vertices)
  are considered (Fig. 14):

\begin{center}
%\begin{picture}(76,54)
\begin{picture}(26,38)
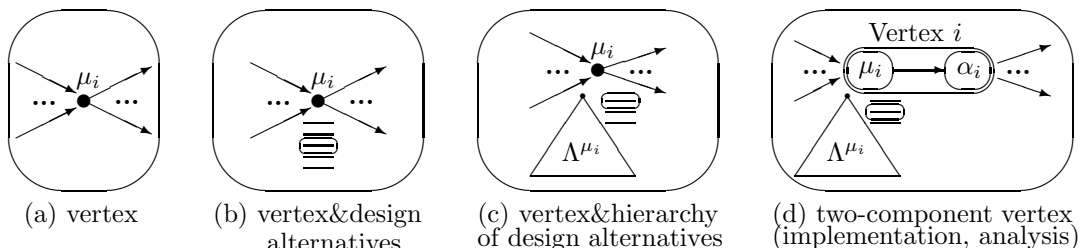
\put(26,00){\makebox(0,0)[bl]{Fig. 14. Four versions of elements
 in ``design/solving  spaces''}}

\put(02,08){\makebox(0,0)[bl]{(a) vertex}}

\put(10,25){\oval(20,24)}

\put(01,30){\vector(2,-1){08}} \put(01,20){\vector(2,1){08}}
\put(03,24.5){\makebox(0,0)[bl]{{\bf ...}}}

\put(09,26.5){\makebox(0,0)[bl]{\(\mu_{i}\)}}
\put(10,25){\circle*{1.7}}

\put(10,25){\vector(2,1){09}} \put(10,25){\vector(2,-1){09}}
\put(14,24.5){\makebox(0,0)[bl]{{\bf ...}}}

%------------------ Initial point  1
%\put(05,15){\circle*{0.8}} \put(05,15){\circle{1.4}}
%\put(01,16){\makebox(0,0)[bl]{\(\mu_{1}\)}}
%\put(05,09.5){\oval(5,2)}
%\put(03,12.5){\line(1,0){4}} \put(03,11){\line(1,0){4}}
%\put(03,09.5){\line(1,0){4}}

\end{picture}
%\end{center}
%
%\begin{center}
%\begin{picture}(76,54)
\begin{picture}(34,38)
%\put(00,00){\makebox(0,0)[bl]{Fig. Versions of elements
% in ``design/solving  spaces''}}

\put(00,08){\makebox(0,0)[bl]{(b) vertex\&design }}
\put(06,05){\makebox(0,0)[bl]{ alternatives}}

\put(14,25){\oval(28,24)}

\put(05,30){\vector(2,-1){08}} \put(05,20){\vector(2,1){08}}
\put(07,24.5){\makebox(0,0)[bl]{{\bf ...}}}

\put(13,26.5){\makebox(0,0)[bl]{\(\mu_{i}\)}}
\put(14,25){\circle*{1.7}}

\put(14,25){\vector(2,1){09}} \put(14,25){\vector(2,-1){09}}
\put(18,24.5){\makebox(0,0)[bl]{{\bf ...}}}

%------------------ Initial point  1
%\put(05,15){\circle*{0.8}} \put(05,15){\circle{1.4}}
%\put(01,16){\makebox(0,0)[bl]{\(\mu_{1}\)}}

\put(14,19){\oval(5,2)}

\put(12,22){\line(1,0){4}} \put(12,20.5){\line(1,0){4}}
\put(12,19){\line(1,0){4}} \put(12,17.5){\line(1,0){4}}
\put(12,16){\line(1,0){4}}
\end{picture}
%\end{center}
%
%\begin{center}
%\begin{picture}(76,54)
\begin{picture}(38,38)
%\put(00,00){\makebox(0,0)[bl]{Fig.  Versions of elements
% in ``design/solving  spaces''}}

\put(00,08){\makebox(0,0)[bl]{(c) vertex\&hierarchy  }}
\put(00,05){\makebox(0,0)[bl]{of design alternatives}}

\put(16,25){\oval(32,24)}

\put(07,34){\vector(2,-1){08}} \put(07,24){\vector(2,1){08}}
\put(09,28.5){\makebox(0,0)[bl]{{\bf ...}}}

\put(15,30.5){\makebox(0,0)[bl]{\(\mu_{i}\)}}
\put(16,29){\circle*{1.7}}

\put(16,29){\vector(3,1){09}} \put(16,29){\vector(3,-1){09}}
\put(20,28.5){\makebox(0,0)[bl]{{\bf ...}}}

%------------------ Initial point  1
%\put(05,15){\circle*{0.8}} \put(05,15){\circle{1.4}}
%\put(01,16){\makebox(0,0)[bl]{\(\mu_{1}\)}}

\put(19,25){\oval(5,2)}

\put(17,25){\line(1,0){4}} \put(17,23.5){\line(1,0){4}}
\put(17,22){\line(1,0){4}}

%-----------------Hierarchy
\put(14,25.6){\circle*{0.8}}

\put(07,15){\line(2,3){7}} \put(21,15){\line(-2,3){7}}
\put(07,15){\line(1,0){14}}

\put(11.2,17){\makebox(0,8)[bl]{\(\Lambda^{\mu_{i}}\)}}

\end{picture}
%\end{center}
%
%\begin{center}
%\begin{picture}(76,54)
\begin{picture}(40,38)
%\put(00,00){\makebox(0,0)[bl]{Fig.  Versions of elements
% in ``design/solving  spaces''}}

\put(00,08){\makebox(0,0)[bl]{(d) two-component vertex}}
\put(00,05){\makebox(0,0)[bl]{(implementation,  analysis)}}

\put(20,25){\oval(40,24)}

%-------------------------
 \put(19.5,29){\oval(20,06)}

\put(12.8,33){\makebox(0,0)[bl]{Vertex \(i\)}}

%-------------------------

\put(03,33){\vector(2,-1){06}} \put(03,25){\vector(2,1){06}}
\put(04,28.5){\makebox(0,0)[bl]{{\bf ...}}}

\put(13,29){\oval(06,05)}
\put(11.5,28){\makebox(0,0)[bl]{\(\mu_{i}\)}}

\put(16,29){\vector(1,0){07}}

\put(26,29){\oval(06,05)}
\put(24.5,28){\makebox(0,0)[bl]{\(\alpha_{i}\)}}

\put(30,30){\vector(3,1){07}} \put(30,28){\vector(3,-1){07}}
\put(31,28.5){\makebox(0,0)[bl]{{\bf ...}}}

%\put(20,28.5){\makebox(0,0)[bl]{{\bf ...}}}

%------------------ Initial point  1
%\put(05,15){\circle*{0.8}} \put(05,15){\circle{1.4}}
%\put(01,16){\makebox(0,0)[bl]{\(\mu_{1}\)}}

\put(15,23.5){\oval(5,2)}

\put(13,25){\line(1,0){4}} \put(13,23.5){\line(1,0){4}}
\put(13,22){\line(1,0){4}}

%-----------------Hierarchy
\put(10,25.6){\circle*{0.8}}

\put(03,15){\line(2,3){7}} \put(17,15){\line(-2,3){7}}
\put(03,15){\line(1,0){14}}

\put(07.2,17){\makebox(0,8)[bl]{\(\Lambda^{\mu_{i}}\)}}

\end{picture}
\end{center}
%

%%%%%%%%%%%%%%%%%%%%%%%%%%%%%%%%%%%%%%%%%%%%%%%%%%%%%%%%%%%%%%%%%
 1. Vertex/node (\(\mu_{i}\)) (Fig. 14a).
 This case corresponds to traditional situation when
 a digraph is used (e.g., in the shortest path problem).

%%%%%%%%%%%%%%%%%%%%%%%%%%%%%%%%%%%%%%%%%%%%%%%%%%%%%%%%%%%%%%%%%%%
 2. Vertex/node (\(\mu_{i}\)) with corresponding design alternatives
 \(\{ A^{\mu_{i}}_{1},..., A^{\mu_{i}}_{q_{\mu_{i}}} \} \)
 (problem: selection of the best design alternative for the vertex)
 (Fig. 14b).
 This case can be used in routing in  ``and-or'' digraphs
 (e.g., \cite{adel02,de90,dinic90}),
 in network routing with selection of the best
 communication protocol at each node
 for the implementation
 (e.g., \cite{chen10,lev12zig,lev13mpeg,gue06}).

%%%%%%%%%%%%%%%%%%%%%%%%%%%%%%%%%%%%%%%%%%%%%%%%%%%%%%%%%%%%%%%%%%%%
 3.  Vertex/node (\(\mu_{i}\)) and corresponding hierarchy of
  design alternatives
 \(\Lambda^{\mu_{i}}\)
 (problem: composition of the best composite design alternative(s)
  on the basis of hierarchy above) (Fig. 14c).
  This case can be used
 in network routing with hierarchical modular design of the implemented
 communication protocol at each node,
 in combinatorial planning of immunoassay technology
 (e.g., \cite{lev06,lev15,levfir05}).

%%%%%%%%%%%%%%%%%%%%%%%%%%%%%%%%%%%%%%%%%%%%%%%%%%%%%%%%%%%%%%%%%%%%
 4. Composite (multi-component) vertex (\(i\)), for example:
  two components as follows:

  (a) ``design/implementation'' part (\(\mu_{i}\))
   (problem:  composition of the best composite design alternative on
 the basis of hierarchy above to implement)
 and

 (b) ``analysis/decision'' part
  (\(\alpha_{i}\))
 (to analyze the result of the implementation above
    and selection of next way/path, based on logical rules) (Fig. 14d).

 This case can be used
 in combinatorial planning of medical treatment
 (i.e., design/implementation and analysis)
 (e.g., \cite{lev13tra,lev15}).

~~

 {\bf Example 1.} An illustrative example for routing based on design
 alternatives in each graph/network vertex is the following.
  For each vertex,
  the resultant design alternative can be  selected in online mode
  or on the basis of off-line solving process
  (e.g., \cite{lev13tra,lev15}).
 Here, each vertex of ``design space'' corresponds to Fig. 14b.
 The example involves the following (Fig. 15):

  (i) \(G = (H,E)\) is a digraph,
 vertex set \(H = \{\mu_{1},\mu_{2},\mu_{3},\mu_{4},\mu_{5},\mu_{6},\mu_{7},\mu_{8}\}\),
arc set \(E = \{ (\mu_{1},\mu_{2}),(\mu_{1},\mu_{3}),\)
 \( (\mu_{2},\mu_{3}),(\mu_{2},\mu_{4}),\) \((\mu_{2},\mu_{5}),\)
 \( (\mu_{3},\mu_{5}),(\mu_{3},\mu_{6}),\)
 \( (\mu_{4},\mu_{6}),(\mu_{3},\mu_{7}),\)
 \( (\mu_{5},\mu_{6}),(\mu_{5},\mu_{7}),\)
 \( (\mu_{6},\mu_{7}),(\mu_{6},\mu_{8}),\)
 \( (\mu_{7},\mu_{8}) \}\);

 (ii) \(\mu_{1}\) is an initial point,
 \(\mu_{8}\) is a goal point;

 (iii) there exist three design alternatives for each
 vertex \(\mu_{i}\) (\(i=\overline{1,8}\)):
 \(A^{\mu_{i}}_{1}\), \(A^{\mu_{i}}_{2}\),\(A^{\mu_{i}}_{3}\);

 (iv) selected alternatives (for each vertex) are:
 \(A^{\mu_{1}}_{3}\), \(A^{\mu_{2}}_{3}\),\(A^{\mu_{3}}_{1}\),
  \(A^{\mu_{4}}_{2}\), \(A^{\mu_{5}}_{2}\),\(A^{\mu_{6}}_{1}\),
   \(A^{\mu_{7}}_{1}\), \(A^{\mu_{8}}_{2}\)
    (In Fig. 14, the alternatives are pointed out by ``oval'');

 (v) the designed global route (by vertices) is:~
 \(L = < \mu_{1},\mu_{2},\mu_{5},\mu_{6},\mu_{8} >\);

 (vi) the resultant route consisting of design alternatives is:~
 \( \widehat{L} = < A^{\mu_{1}}_{3},A^{\mu_{2}}_{3},A^{\mu_{5}}_{2},
  A^{\mu_{6}}_{1},A^{\mu_{8}}_{2}>\).

 Evidently,
 in the problem the selected design alternatives in neighbor
 path vertices have to be ``good'' compatible
 as in combinatorial synthesis approach
 (morphological clique problem)
 (e.g., \cite{lev98,lev06,lev09,lev15}).

\begin{center}
\begin{picture}(77,44)
%\begin{picture}(74,44)
\put(05.5,00){\makebox(0,0)[bl]{Fig. 15.
% Digraph
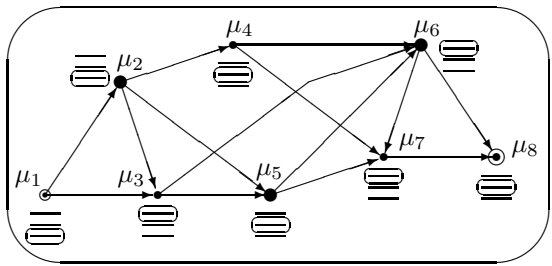
  Design alternatives for vertices}}

\put(36,23){\oval(72,34)}

%------------------ Goal point  8
\put(65,20){\circle{2.0}} \put(65,20){\circle*{1.2}}

\put(67,20){\makebox(0,0)[bl]{\(\mu_{8}\)}}

\put(65,16){\oval(5,2)}

\put(63,17.5){\line(1,0){4}} \put(63,16){\line(1,0){4}}
\put(63,14.5){\line(1,0){4}}

%------------ Path

%\put(50,16){\vector(1,1){04.5}}

%\put(40,21){\vector(2,-1){09.5}} \put(50,16){\circle*{1.0}}
%\put(40,30){\vector(0,-1){08.5}} \put(40,21){\circle*{1.0}}

%\put(30,30){\vector(1,0){09.5}} \put(40,30){\circle*{1.8}}
%\put(25,35){\vector(1,-1){04.5}} \put(30,30){\circle*{1.8}}
%\put(20,30){\vector(1,1){04.5}} \put(25,35){\circle*{1.8}}

%\put(20,21){\vector(0,1){08.5}} \put(20,30){\circle*{1.8}}
%\put(15,21){\vector(1,0){04.5}} \put(20,21){\circle*{1.0}}
%\put(10,16){\vector(1,1){04.5}} \put(15,21){\circle*{1.0}}
%\put(05,16){\vector(1,0){04.5}} \put(10,16){\circle*{1.0}}

%===============================================
%%%%%%%%%%%%%%% Arcs of graph

%-- 1-3
\put(05,15){\vector(1,0){014.3}}

%-- 3-5
\put(20,15){\vector(1,0){014.3}}

%---- 1-2
\put(05,15){\vector(2,3){09.5}}

%---- 2-4
\put(15,30){\vector(3,1){14.5}}

%---- 2-4
\put(15,30){\vector(1,-3){04.7}}

%---- 2-5
\put(15,30){\vector(4,-3){19.4}}

%---- 4-6
\put(30,35){\vector(1,0){24.1}}

%---- 4-5
\put(30,35){\vector(4,-3){19.5}}

%---- 3-6
\put(20,15){\line(4,3){20}} \put(40,30){\vector(3,1){14.2}}

%---- 5-6
\put(35,15){\vector(1,1){19.5}}

%---- 5-7
\put(35,15){\vector(3,1){14.3}}

%---- 7-8
\put(50,20){\vector(1,0){14.4}}

%---- 6-7
\put(55,35){\vector(-1,-3){04.8}}

%---- 6-8
\put(55,35){\vector(2,-3){09.4}}

%---------------------------------------- 7
\put(50,20){\circle*{1.0}}

\put(52,21){\makebox(0,0)[bl]{\(\mu_{7}\)}}

\put(50,17.5){\oval(5,2)}

\put(48,17.5){\line(1,0){4}} \put(48,16){\line(1,0){4}}
\put(48,14.5){\line(1,0){4}}

%---------------------------------------- 6
\put(55,35){\circle*{1.6}}

\put(54,36){\makebox(0,0)[bl]{\(\mu_{6}\)}}

\put(60,34.5){\oval(5,2)}

\put(58,34.5){\line(1,0){4}} \put(58,33){\line(1,0){4}}
\put(58,31.5){\line(1,0){4}}

%---------------------------------------- 5
\put(35,15){\circle*{1.8}}

\put(33,17){\makebox(0,0)[bl]{\(\mu_{5}\)}}

\put(35,11){\oval(5,2)}

\put(33,12.5){\line(1,0){4}} \put(33,11){\line(1,0){4}}
\put(33,09.5){\line(1,0){4}}

%---------------------------------------- 4
\put(30,35){\circle*{1.0}}

\put(29,36){\makebox(0,0)[bl]{\(\mu_{4}\)}}

\put(30,31){\oval(5,2)}

\put(28,32.5){\line(1,0){4}} \put(28,31){\line(1,0){4}}
\put(28,29.5){\line(1,0){4}}

%---------------------------------------- 2
\put(15,30){\circle*{1.8}}

\put(14.5,31.5){\makebox(0,0)[bl]{\(\mu_{2}\)}}

\put(011,30.5){\oval(5,2)}

\put(09,33.5){\line(1,0){4}} \put(09,32){\line(1,0){4}}
\put(09,30.5){\line(1,0){4}}

%---------------------------------------- 3
\put(20,15){\circle*{1.0}}

\put(14.6,16){\makebox(0,0)[bl]{\(\mu_{3}\)}}

\put(20,12.5){\oval(5,2)}

\put(18,12.5){\line(1,0){4}} \put(18,11){\line(1,0){4}}
\put(18,09.5){\line(1,0){4}}

%------------------ Initial point  1
\put(05,15){\circle*{0.8}} \put(05,15){\circle{1.4}}

\put(01,16){\makebox(0,0)[bl]{\(\mu_{1}\)}}

\put(05,09.5){\oval(5,2)}

\put(03,12.5){\line(1,0){4}} \put(03,11){\line(1,0){4}}
\put(03,09.5){\line(1,0){4}}

\end{picture}
%\end{center}
%
%\begin{center}
%\begin{picture}(76,54)
\begin{picture}(78,42)
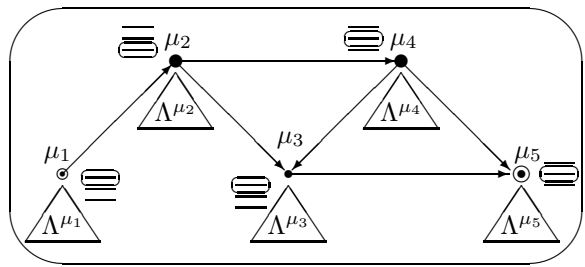
\put(03,00){\makebox(0,0)[bl]{Fig. 16. Hierarchies of alternatives
 for vertices}}

\put(39,22){\oval(76,34)}

%------------------ Intermediate point  2
\put(23,32){\circle*{1.8}}

\put(21.5,33.5){\makebox(0,0)[bl]{\(\mu_{2}\)}}

%--
\put(18,33.5){\oval(5,2)}

\put(016,36.5){\line(1,0){4}} \put(016,35){\line(1,0){4}}
\put(016,33.5){\line(1,0){4}}

%--

\put(18,23){\line(2,3){5}} \put(28,23){\line(-2,3){5}}
\put(18,23){\line(1,0){10}}

\put(20.2,24){\makebox(0,8)[bl]{\(\Lambda^{\mu_{2}}\)}}

%------------------ Intermediate point  4
\put(53,32){\circle*{1.8}}

\put(51.5,33.5){\makebox(0,0)[bl]{\(\mu_{4}\)}}

%----------
\put(48,35){\oval(5,2)}

\put(46,36.5){\line(1,0){4}} \put(46,35){\line(1,0){4}}
\put(46,33.5){\line(1,0){4}}

%---------

\put(48,23){\line(2,3){5}} \put(58,23){\line(-2,3){5}}
\put(48,23){\line(1,0){10}}

\put(50.2,24){\makebox(0,8)[bl]{\(\Lambda^{\mu_{4}}\)}}

%================================================
%------------------ Initial point  1
\put(08,17){\circle{1.4}} \put(08,17){\circle*{0.8}}

\put(05.5,18.5){\makebox(0,0)[bl]{\(\mu_{1}\)}}

\put(13,16.5){\oval(5,2)}

\put(11,16.5){\line(1,0){4}} \put(11,15){\line(1,0){4}}
\put(11,13.5){\line(1,0){4}}

\put(03,08){\line(2,3){5}} \put(13,08){\line(-2,3){5}}
\put(03,08){\line(1,0){10}}

\put(05.2,09){\makebox(0,8)[bl]{\(\Lambda^{\mu_{1}}\)}}

%------------------ Intermediate point  3
\put(38,17){\circle*{1.0}}

\put(36.5,20.5){\makebox(0,0)[bl]{\(\mu_{3}\)}}

%----------
\put(33,15.5){\oval(5,2)}

\put(31,15.5){\line(1,0){4}} \put(31,14){\line(1,0){4}}
\put(31,12.5){\line(1,0){4}}

%---------

\put(33,08){\line(2,3){5}} \put(43,08){\line(-2,3){5}}
\put(33,08){\line(1,0){10}}

\put(35.2,09){\makebox(0,8)[bl]{\(\Lambda^{\mu_{3}}\)}}

%------------------ Goal point  5
\put(69,17){\circle{2.0}} \put(69,17){\circle*{1.2}}

\put(68,18.5){\makebox(0,0)[bl]{\(\mu_{5}\)}}

\put(74,17){\oval(5,2)}

\put(72,18.5){\line(1,0){4}} \put(72,17){\line(1,0){4}}
\put(72,15.5){\line(1,0){4}}

\put(64,08){\line(2,3){5}} \put(74,08){\line(-2,3){5}}
\put(64,08){\line(1,0){10}}

\put(66.2,09){\makebox(0,8)[bl]{\(\Lambda^{\mu_{5}}\)}}

%------------ Path

%===============================================
%%%%%%%%%%%%%%% Arcs of graph

%-- 1-2
\put(08,17){\vector(1,1){014.5}}

%-- 2-3
\put(23,32){\vector(1,-1){014.5}}

%-- 4-3
\put(53,32){\vector(-1,-1){014.5}}

%-- 4-5
\put(53,32){\vector(1,-1){014.5}}

%-- 1-5
%\put(08,17){\vector(1,0){029.3}}

%-- 3-5
\put(38,17){\vector(1,0){029.3}}

%---- 2-4
\put(23,32){\vector(1,0){29.3}}

\end{picture}
\end{center}
%

%~~

 {\bf Example 2.} An illustrative example for routing based
 hierarchy of design
 alternatives in each graph/network vertex is the following.
  In each vertex,
  the resultant set of  design alternative can be  design in online mode
  or on the basis of offline solving process.
 Here, each vertex of ``design space'' corresponds to Fig. 14d.
  In \cite{lev13tra,lev15},
  this problem is examined as multi-stage design of modular
  systems.
 The example involves the following (Fig. 16):

  (i) \(G = (H,E)\) is a digraph,
 vertex set  \(H = \{\mu_{1},\mu_{2},\mu_{3},\mu_{4},\mu_{5}\}\),
 arc set \(E = \{ (\mu_{1},\mu_{2}),(\mu_{2},\mu_{3}),\)
 \( (\mu_{2},\mu_{4}),(\mu_{3},\mu_{5}),\) \((\mu_{4},\mu_{3}),\)
 \( (\mu_{4},\mu_{5}) \}\));

 (ii) \(\mu_{1}\) is an initial point,
 \(\mu_{5}\) is a goal point;

 (iii) there exists a hierarchy of design alternatives for each
 vertex \(\mu_{i}\) (\(i=\overline{1,5}\)):
 \(\Lambda^{\mu_{i}}\);

 (iv) three design alternatives are composed  for each
 vertex \(\mu_{i}\) (\(i=\overline{1,5}\)):
 \(A^{\mu_{i}}_{1}\), \(A^{\mu_{i}}_{2}\),\(A^{\mu_{i}}_{3}\);

 (v) selected alternatives (for each vertex) are:
 \(A^{\mu_{1}}_{1}\), \(A^{\mu_{2}}_{3}\),\(A^{\mu_{3}}_{1}\),
  \(A^{\mu_{4}}_{2}\), \(A^{\mu_{5}}_{2}\)
   (in Fig. 15, the alternatives are pointed out by ``oval'');

 (vi) the designed global route (by vertices) is:~
 \(L = < \mu_{1},\mu_{2},\mu_{4},\mu_{5} >\);

 (vii) the resultant route consisting of design alternatives is:~
 \( \widehat{L} = < A^{\mu_{1}}_{1},A^{\mu_{2}}_{3},A^{\mu_{4}}_{2},
  A^{\mu_{5}}_{2}>\).

 Evidently,
 in the problem the selected design alternatives in neighbor
 path vertices have to be ``good'' compatible
 as in combinatorial synthesis approach
 (morphological clique problem)
 (e.g., \cite{lev98,lev06,lev09,lev15}).

\subsection{Basic solving strategies}

 Let us consider one-layer route DM problem with
 nodes as ``vertex\& alternatives'' and ``vertex\& hierarchy alternatives''
 (Fig. 14c and Fig. 15, Fig. 14d and Fig. 16).
 In general, the following two basic solving strategies can be pointed out
 for these problem types:

~~

 {\it Strategy 1.} ``Global'' route strategy:

 (1.1) designing a ``global'' route over graph vertices,

 (1.2) selection/design of the best design alternative for each graph
 vertex,

 (1.3) designing a resultant route from the best alternatives.

 This strategy 1 is illustrated in Example 1 and in Example 2
 (Fig. 15, Fig. 16).
 The strategy was used in \cite{lev13tra,lev15}.

~~

 {\it Strategy 2.} Extended digraph strategy:

 (2.1) transformation/extension of the initial graph/network model
 as examination of design alternatives for each model node instead
 of the node
% while taking into account design alternatives for each node
 (e.g., Fig. 17 instead of Fig. 16),

 (2.2) designing the best route over the obtained extended graph/network.

 Evidently, this strategy increases the problem dimension.

\begin{center}
\begin{picture}(106,68.5)
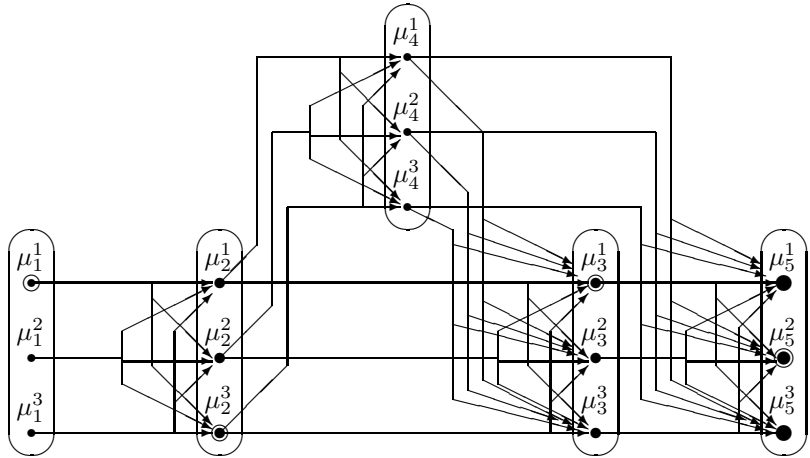
\put(013,00){\makebox(0,0)[bl]{Fig. 17. Illustration for extended
  digraph strategy}}

%---------------- Mu 1

\put(03,22){\oval(06,30)}

\put(03,30){\circle*{1.2}} \put(03,30){\circle{2}}
\put(01,31.5){\makebox(0,0)[bl]{\(\mu_{1}^{1}\)}}

\put(03,20){\circle*{1.2}}
\put(01,21.5){\makebox(0,0)[bl]{\(\mu_{1}^{2}\)}}

\put(03,10){\circle*{1.2}}
\put(01,11.5){\makebox(0,0)[bl]{\(\mu_{1}^{3}\)}}

%--  mu1->mu2

\put(03,30){\line(1,0){16}} \put(19,30){\line(0,-1){11}}
\put(19,30){\vector(1,0){8}} \put(19,28){\vector(1,-1){8}}
\put(19,19){\vector(1,-1){8}}

\put(03,20){\line(1,0){12}} \put(15,16.5){\line(0,1){07}}
\put(15,23.5){\vector(2,1){12}} \put(15,19.5){\vector(1,0){12}}
\put(15,16.5){\vector(2,-1){12}}

\put(03,10){\line(1,0){19}} \put(22,10){\line(0,1){13.5}}
\put(22,10){\vector(1,0){5}} \put(22,14){\vector(1,1){5}}
\put(22,23.5){\vector(1,1){5}}

%---------------- Mu 2

\put(28,22){\oval(06,30)}

\put(28,30){\circle*{1.5}}
\put(26,31.5){\makebox(0,0)[bl]{\(\mu_{2}^{1}\)}}

\put(28,20){\circle*{1.5}}
\put(26,21.5){\makebox(0,0)[bl]{\(\mu_{2}^{2}\)}}

\put(28,10){\circle*{1.5}} \put(28,10){\circle{2.2}}
\put(26,12.5){\makebox(0,0)[bl]{\(\mu_{2}^{3}\)}}

%--  mu2->mu3

\put(28,30){\line(1,0){41}} \put(69,30){\line(0,-1){11}}
\put(69,30){\vector(1,0){8}} \put(69,28){\vector(1,-1){8}}
\put(69,19){\vector(1,-1){8}}

\put(28,20){\line(1,0){37}} \put(65,16.5){\line(0,1){07}}
\put(65,23.5){\vector(2,1){12}} \put(65,19.5){\vector(1,0){12}}
\put(65,16.5){\vector(2,-1){12}}

\put(28,10){\line(1,0){44}} \put(72,10){\line(0,1){13.5}}
\put(72,10){\vector(1,0){5}} \put(72,14){\vector(1,1){5}}
\put(72,23.5){\vector(1,1){5}}

%--  mu2->mu4

\put(28,30){\line(1,1){05}} \put(33,35){\line(0,1){25}}

\put(33,60){\line(1,0){11}} \put(44,60){\line(0,-1){11}}
\put(44,60){\vector(1,0){8}} \put(44,58){\vector(1,-1){8}}
\put(44,49){\vector(1,-1){8}}

\put(28,20){\line(1,1){07}} \put(35,27){\line(0,1){23}}

\put(35,50){\line(1,0){05}} \put(40,46.5){\line(0,1){07}}
\put(40,53.5){\vector(2,1){12}} \put(40,49.5){\vector(1,0){12}}
\put(40,46.5){\vector(2,-1){12}}

\put(28,10){\line(1,1){09}} \put(37,19){\line(0,1){21}}

\put(37,40){\line(1,0){10}} \put(47,40){\line(0,1){13.5}}
\put(47,40){\vector(1,0){5}} \put(47,44){\vector(1,1){5}}
\put(47,53.5){\vector(1,1){5}}

%---------------- Mu 3

\put(78,22){\oval(06,30)}

\put(78,30){\circle*{1.4}} \put(78,30){\circle{2.2}}
\put(76,31.5){\makebox(0,0)[bl]{\(\mu_{3}^{1}\)}}

\put(78,20){\circle*{1.4}}
\put(76,21.5){\makebox(0,0)[bl]{\(\mu_{3}^{2}\)}}

\put(78,10){\circle*{1.4}}
\put(76,12.5){\makebox(0,0)[bl]{\(\mu_{3}^{3}\)}}

%--  mu3->mu5

\put(78,30){\line(1,0){16}} \put(94,30){\line(0,-1){11}}
\put(94,30){\vector(1,0){8}} \put(94,28){\vector(1,-1){8}}
\put(94,19){\vector(1,-1){8}}

\put(78,20){\line(1,0){12}} \put(90,16.5){\line(0,1){07}}
\put(90,23.5){\vector(2,1){12}} \put(90,19.5){\vector(1,0){12}}
\put(90,16.5){\vector(2,-1){12}}

\put(78,10){\line(1,0){19}} \put(97,10){\line(0,1){14}}
\put(97,10){\vector(1,0){5}} \put(97,14){\vector(1,1){5}}
\put(97,24){\vector(1,1){5}}

%---------------- Mu 4

\put(53,52){\oval(06,30)}

\put(53,60){\circle*{1.0}}
\put(51,61.5){\makebox(0,0)[bl]{\(\mu_{4}^{1}\)}}

\put(53,50){\circle*{1.0}}
\put(51,51.5){\makebox(0,0)[bl]{\(\mu_{4}^{2}\)}}

\put(53,40){\circle*{1.0}}
\put(51,42.5){\makebox(0,0)[bl]{\(\mu_{4}^{3}\)}}

%----------------- mu4 -> mu3

\put(53,60){\line(1,-1){10}} \put(63,50){\line(0,-1){33}}
\put(63,38.5){\vector(2,-1){12}} \put(63,27.5){\vector(2,-1){12}}
\put(63,17){\vector(2,-1){12}}

\put(53,50){\line(1,-1){08}} \put(61,42){\line(0,-1){26.3}}
\put(61,36.5){\vector(3,-1){14}} \put(61,25.7){\vector(3,-1){14}}
\put(61,15.7){\vector(3,-1){12}}

\put(53,40){\line(2,-1){06}} \put(59,37){\line(0,-1){22.5}}
\put(59,35){\vector(4,-1){16.6}}
\put(59,24.5){\vector(4,-1){16.6}}
\put(59,14.5){\vector(4,-1){16.6}}

%----------------- mu4 -> mu5

\put(53,60){\line(1,0){35}} \put(88,60){\line(0,-1){43}}
\put(88,38.5){\vector(2,-1){12}} \put(88,27.5){\vector(2,-1){12}}
\put(88,17){\vector(2,-1){12}}

\put(53,50){\line(1,0){33}} \put(86,50){\line(0,-1){34.3}}
\put(86,36.5){\vector(3,-1){14}} \put(86,25.7){\vector(3,-1){14}}
\put(86,15.7){\vector(3,-1){12}}

\put(53,40){\line(1,0){31}} \put(84,40){\line(0,-1){25.5}}
\put(84,35){\vector(4,-1){16.6}}
\put(84,24.5){\vector(4,-1){16.6}}
\put(84,14.5){\vector(4,-1){16.6}}

%---------------- Mu 5

\put(103,22){\oval(06,30)}

\put(103,30){\circle*{2.0}}
\put(101,31.5){\makebox(0,0)[bl]{\(\mu_{5}^{1}\)}}

\put(103,20){\circle*{1.7}} \put(103,20){\circle{2.5}}
\put(101,21.5){\makebox(0,0)[bl]{\(\mu_{5}^{2}\)}}

\put(103,10){\circle*{2.0}}
\put(101,12.5){\makebox(0,0)[bl]{\(\mu_{5}^{3}\)}}

%-----
\end{picture}
\end{center}
%

%~~

 {\bf Example 3.} Here, digraph \(G=(H,E)\) from example 2  above
 (Fig. 16)
 is considered as an initial one.
 For each vertex, three alternatives are examined (Fig. 16),
 and  resultant digraph
 \(\widehat{G}=(\widehat{H},\widehat{E})\)
 (where \( \widehat{H} = \{
  \mu_{1}^{1}, \mu_{1}^{2}, \mu_{1}^{3},
  \mu_{2}^{1}, \mu_{2}^{2}, \mu_{2}^{3},
  \mu_{3}^{1}, \mu_{3}^{2}, \mu_{3}^{3},
  \mu_{4}^{1}, \mu_{4}^{2}, \mu_{4}^{3},
  \mu_{5}^{1}, \mu_{5}^{2}, \mu_{5}^{3}\}\))
 is shown in Fig. 17.
 (Here,
 ``superscript'' index corresponds to the number of alternative,
 i.e., \(\mu_{i}^{j}\) corresponds  to design alternative \(A^{\mu_{i}}_{j}\) ).
 In this problem, it is necessary to select
 the initial point
 from the set
 \(\{\mu_{1}^{1}, \mu_{1}^{2}, \mu_{1}^{3}\}\)
 and
 the goal point from the set
 \(\{\mu_{5}^{1}, \mu_{5}^{2}, \mu_{5}^{3}\}\).
 Further, the route has to be build.

 For example,
 \(\mu_{1}^{1}\) is the initial point (as in Fig. 16),
 \(\mu_{5}^{2}\) is the goal point (as in Fig. 16).
 The global route is:~
 \(L = <\mu_{1},  \mu_{2},\mu_{3}, \mu_{5}> \).
 The resultant  route (by design alternatives) is:~
 \(\widehat{L} = <\mu_{1}^{1},  \mu_{2}^{3},\mu_{3}^{1}, \mu_{5}^{2}> \).

%%%%%%%%%%%%%%%%%%%%%%%%%%%%%%%%%%%%%%%%%%%%%%%%%%%%%%%%%%%%%%%%
%%%%%%%%%%%%%%%%%%%%%%%%%%%%%%%%%%%%%%%%%%%%%%%%%%%%%%%%%%%%%%%%
\section{Some applied
% Intelligent
 route decision making problems}

% Now, it is reasonable to consider ``intelligent'' route decision
% making  problems.

\subsection{Student orienteering problem (educational trajectory)}

%
%%%%%%%%%%%%%%%%%%%%%%%%%%%%%%%%%%%%%%%%% Education
%
 In educational domain,
 the following route decision making problem
 has been examined:~
%  (Fig. ):
 design of educational route
 (e.g., for student/teenager)
 (e.g., \cite{lev98}).
 Here, support problems are the following:
 (i) analysis/diagnosis of initial situation (i.e., point),
 (ii) definition/specification/design of
 goal point(s),
 (iii) design of route ``space''
 (i.e., a set of education/life operations),
 (iv) design of educational route,
 (v) online analysis of the route implementation and online modification (correction)
 of the route.

 Further, a simplified plan (educational trajectory)
  for a BS student of Moscow Inst. of Physics and Technology (State Univ.)
 (Faculty of Radio Engineering and Cybernetics)
 is examined.
 A BS degree in Communication Engineering from Moscow Inst. of
 Physics and Technology
 (Communication Engineering)
 is considered as an initial point  \(a_{1}\).
 Other educational points/nodes
 are presented in Table 2 including their
 characteristics and estimates upon criteria
 (ordinal scale \([1,5]\),
 \(5\) corresponds to the best level):
 (i) quality of education
 (i.e., a set of disciplines,
  basic lectures, seminars)
 \(C_{1}\),
 (ii) possible research results
 (including publication activity)
  \(C_{2}\),
 (iii) integrated index of professional degree prestige
 (World University Rating, quality of professional education/research,
 scientific school(s), etc.)
  \(C_{3}\).

%\newpage
\begin{center}
{\bf Table 2.} Digraph model nodes/vertices for student planning (characteristics, estimates)   \\
\begin{tabular}{|c|c|c |l|l| c |ccc|}
\hline
 No.&Notation  &Degree&University&Profession&Time   &\(\theta^{1}\)&\(\theta^{2}\)&\(\theta^{3}\)\\
    &(node/vertex)&level &  &          &(years)&         &     &      \\
    &        &      &          &          &\(\tau\)&         &     &      \\

\hline

 1.&\(a_{1}\) (initial, origin)&BS& MIPT, Russia & Commun. Eng.&4 &  &   & \\
%   & origin)& &   &  &&&&\\

\hline
 2.&\(b_{1}\) &MS & MIPT, Russia &Commun. Eng.     &\(2\)  &\(5\)&\(2\)&\(4\)\\
 3.&\(b_{2}\) &MS & MIPT, Russia & Appl. Math.     &\(2\)  &\(5\)&\(4\)&\(4\)\\
 4.&\(b_{3}\) &MS & USA univ. & Commun. Eng.       &\(2\)  &\(3\)&\(5\)&\(5\)\\
 5.&\(b_{4}\) &MS & USA univ. & Inform. Syst.      &\(2\)  &\(3\)&\(5\)&\(5\)\\
 6.&\(b_{5}\) &MS & Canadian univ.&Commun. Eng.    &\(2\)  &\(3\)&\(4\)&\(4\)\\
 7.&\(b_{6}\) &MS & UK univ. & OR/Algorithms       &\(2\)  &\(4\)&\(5\)&\(4\)\\
 8.&\(b_{7}\) &MS & German univ.&Inform. Syst.     &\(2\)  &\(3\)&\(2\)&\(4\)\\
\hline
 9.&\(g_{1}\)&MS (2nd)&USA univ.&Commun. Eng.      &\(1\)  &\(3\)&\(5\)&\(5\)\\
 10.&\(g_{2}\)&MS (2nd)&USA univ.&Inform. Syst.     &\(1\)  &\(3\)&\(5\)&\(5\)\\
 11.&\(g_{3}\)&MS (2nd)&Canadian univ.&Commun. Eng.&\(1\)  &\(3\)&\(4\)&\(4\)\\
 12.&\(g_{4}\)&MS (2nd)& UK univ. & OR/Algorithms  &\(1\)  &\(4\)&\(5\)&\(4\)\\
 13.&\(g_{5}\)&MS (2nd)&German univ.&Inform. Syst. &\(1\)  &\(3\)&\(4\)&\(4\)\\

\hline
 14.&\(h_{1}\) &PhD  & MIPT, Russia & Commun. Eng. &\(3\)  &\(5\)&\(3\)&\(4\) \\
 15.&\(h_{2}\) &PhD  & MIPT, Russia & Appl. Math.  &\(3\)  &\(5\)&\(4\)&\(5\) \\
 16.&\(h_{3}\) &PhD & USA univ. & Commun. Eng.     &\(3\)  &\(4\)&\(5\)&\(5\)\\
 17.&\(h_{4}\) &PhD & USA univ. & Inform. Syst.    &\(3\)  &\(4\)&\(5\)&\(5\)\\
 18.&\(h_{5}\) &PhD & Canadian univ. & Commun. Eng.&\(3\)  &\(4\)&\(4\)&\(5\)\\
 19.&\(h_{6}\) &PhD & UK univ. & OR/Algorithms     &\(3\)  &\(5\)&\(5\)&\(5\)\\
 20.&\(h_{7}\) &PhD & German univ. & Inform. Syst. &\(3\)  &\(4\)&\(4\)&\(4\) \\

\hline
 21.&\(f_{1}\)&PhD (2nd)& USA univ. & Commun. Eng.  &\(2\) &\(4\)&\(5\)&\(5\)\\
 22.&\(f_{2}\)&PhD (2nd)& USA univ. & Inform. Syst. &\(2\) &\(4\)&\(5\)&\(5\)\\
 23.&\(f_{3}\)&PhD (2nd)&Canadian univ.&Commun. Eng.&\(2\) &\(4\)&\(5\)&\(5\)\\

\hline
 24.&\(p_{1}\) (goal,&PostDoc& USA univ.&Commun. Eng. &\(3\) &\(3\)&\(5\)&\(5\)\\
    & destination)& &   &  &&&&\\

\hline
\end{tabular}
\end{center}

 Here, four basic ``generalized'' educational routes/trajectiories
 can be examined
% (Fig. 18)
 (\(\iota=\overline{1,7}\),
 \(\kappa=\overline{1,5}\),
 \(\upsilon=\overline{1,7}\),
 \(\xi=\overline{1,3}\)):
%
% Trajectory \(1\):~
  ~\(L^{1} = < a_{1}, b_{\iota}, h_{\upsilon}, p_{1}>\) (Fig. 18a);
%
% Trajectory \(2\):~
  ~\(L^{2} = < a_{1}, b_{\iota}, g_{\kappa}, h_{\upsilon}, p_{1}>\) (Fig. 18b);
%
% Trajectory \(3\):~
  ~\(L^{3} = < a_{1}, b_{\iota}, h_{\upsilon},f_{\xi}, p_{1}>\) (Fig. 18c);
%
% Trajectory \(4\):~
  ~\(L^{4} = < a_{1}, b_{\iota},g_{\kappa}, h_{\upsilon},f_{\xi}, p_{1}>\) (Fig. 18d).
 Table 3, Table 4, Table 5 contain
 ordinal complexity estimates of movement between model nodes
 (scale \([1,5]\),
 \(5\) corresponds to the most complex movement;
 absence of estimate corresponds to impossible movement:
% i.e.,
 the digraph arc is absent).

\begin{center}
\begin{picture}(36,31)
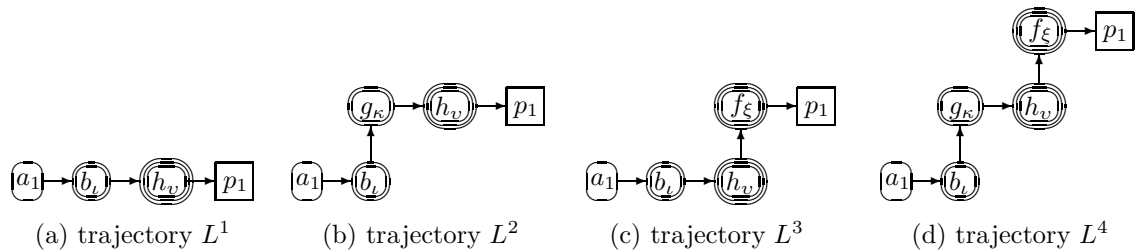
\put(029,00){\makebox(0,0)[bl]{Fig. 18. Four basic
  generalized educational trajectories}}

%---------------- Route 1
\put(3,5){\makebox(0,0)[bl]{(a) trajectory  \(L^{1}\)}}

\put(02,14){\oval(04,05)}
\put(0.5,13){\makebox(0,0)[bl]{\(a_{1}\)}}
\put(04,14){\vector(1,0){4}}

\put(10.5,14){\oval(05,5)} \put(10.5,14){\oval(04,4)}
\put(09,12.3){\makebox(0,0)[bl]{\(b_{\iota}\)}}
\put(13,14){\vector(1,0){4}}

\put(20.5,14){\oval(07,6)} \put(20.5,14){\oval(06,5)}
\put(20.5,14){\oval(05,4)}
\put(18.5,12.3){\makebox(0,0)[bl]{\(h_{\upsilon}\)}}

\put(23,14){\vector(1,0){4}}

\put(27,11.5){\line(1,0){5}} \put(27,16.5){\line(1,0){5}}
\put(27,11.5){\line(0,1){5}} \put(32,11.5){\line(0,1){5}}

\put(28,12.5){\makebox(0,0)[bl]{\(p_{1}\)}}

\end{picture}
%\end{center}
%
%\begin{center}
\begin{picture}(38,31)

%---------------- Route 2 - b
\put(04,05){\makebox(0,0)[bl]{(b) trajectory  \(L^{2}\)}}

\put(02,14){\oval(04,05)}
\put(0.5,13){\makebox(0,0)[bl]{\(a_{1}\)}}
\put(04,14){\vector(1,0){4}}

\put(10.5,14){\oval(05,5)} \put(10.5,14){\oval(04,4)}
\put(09,12.3){\makebox(0,0)[bl]{\(b_{\iota}\)}}
\put(10.5,16.5){\vector(0,1){5}}

\put(10.5,24){\oval(06,5)} \put(10.5,24){\oval(05,4)}
\put(09,22.3){\makebox(0,0)[bl]{\(g_{\kappa}\)}}
\put(13.5,24){\vector(1,0){4}}

\put(21,24){\oval(07,6)} \put(21,24){\oval(06,5)}
\put(21,24){\oval(05,4)}
\put(19,22.3){\makebox(0,0)[bl]{\(h_{\upsilon}\)}}

\put(24.5,24){\vector(1,0){4}}

\put(28.5,21.5){\line(1,0){5}} \put(28.5,26.5){\line(1,0){5}}
\put(28.5,21.5){\line(0,1){5}} \put(33.5,21.5){\line(0,1){5}}

\put(29.5,22.5){\makebox(0,0)[bl]{\(p_{1}\)}}

\end{picture}
%\end{center}
%
%\begin{center}
\begin{picture}(38,31)

%---------------- Route 3 - c
\put(3,5){\makebox(0,0)[bl]{(c) trajectory  \(L^{3}\)}}

\put(02,14){\oval(04,05)}
\put(0.5,13){\makebox(0,0)[bl]{\(a_{1}\)}}
\put(04,14){\vector(1,0){4}}

\put(10.5,14){\oval(05,5)} \put(10.5,14){\oval(04,4)}
\put(09,12.3){\makebox(0,0)[bl]{\(b_{\iota}\)}}
\put(13,14){\vector(1,0){4}}

\put(20.5,14){\oval(07,6)} \put(20.5,14){\oval(06,5)}
\put(20.5,14){\oval(05,4)}
\put(18.5,12.3){\makebox(0,0)[bl]{\(h_{\upsilon}\)}}

\put(20.5,17){\vector(0,1){4}}

\put(20.5,24){\oval(07,6)} \put(20.5,24){\oval(06,5)}
\put(20.5,24){\oval(05,4)}
\put(19,22.3){\makebox(0,0)[bl]{\(f_{\xi}\)}}
\put(24,24){\vector(1,0){4}}

\put(28,21.5){\line(1,0){5}} \put(28,26.5){\line(1,0){5}}
\put(28,21.5){\line(0,1){5}} \put(33,21.5){\line(0,1){5}}
\put(29,22.5){\makebox(0,0)[bl]{\(p_{1}\)}}

\end{picture}
%\end{center}
%
%\begin{center}
\begin{picture}(34,37.5)

%---------------- Route 4 - d
\put(04,05){\makebox(0,0)[bl]{(d) trajectory  \(L^{4}\)}}

\put(02,14){\oval(04,05)}
\put(0.5,13){\makebox(0,0)[bl]{\(a_{1}\)}}
\put(04,14){\vector(1,0){4}}

\put(10.5,14){\oval(05,5)} \put(10.5,14){\oval(04,4)}
\put(09,12.3){\makebox(0,0)[bl]{\(b_{\iota}\)}}
\put(10.5,16.5){\vector(0,1){5}}

\put(10.5,24){\oval(06,5)} \put(10.5,24){\oval(05,4)}
\put(09,22.3){\makebox(0,0)[bl]{\(g_{\kappa}\)}}
\put(13.5,24){\vector(1,0){4}}

\put(21,24){\oval(07,6)} \put(21,24){\oval(06,5)}
\put(21,24){\oval(05,4)}
\put(19,22.3){\makebox(0,0)[bl]{\(h_{\upsilon}\)}}
\put(21,27){\vector(0,1){4}}

\put(21,34){\oval(07,6)} \put(21,34){\oval(06,5)}
\put(21,34){\oval(05,4)}
\put(19.5,32.3){\makebox(0,0)[bl]{\(f_{\xi}\)}}
\put(24.5,34){\vector(1,0){4}}

\put(28.5,31.5){\line(1,0){5}} \put(28.5,36.5){\line(1,0){5}}
\put(28.5,31.5){\line(0,1){5}} \put(33.5,31.5){\line(0,1){5}}

\put(29.5,32.5){\makebox(0,0)[bl]{\(p_{1}\)}}

\end{picture}
\end{center}
%

%\newpage
\begin{center}
{\bf Table 3.} Ordinal estimates of movement complexity  \(\lambda( a_{1} \rightarrow  b_{\iota}) \) \\
\begin{tabular}{| c | c c c c c c c|}
\hline
 \(a_{1}\) \(\backslash \) \(b_{i}\)  &\(b_{1}\)&\(b_{2}\)&\(b_{3}\)&\(b_{4}\)&\(b_{5}\)&\(b_{6}\)&\(b_{7}\)  \\
\hline

 \(a_{1}\) &\(1\)&\(2\)&\(4\)&\(5\)&\(4\)&\(5\)&\(3\) \\

\hline
\end{tabular}
\end{center}

%\newpage
\begin{center}
{\bf Table 4.} Ordinal estimates of movement complexity:
    \(\lambda(b_{\iota}  \rightarrow  g_{\kappa})\),
 \(\lambda(b_{\iota} \rightarrow h_{\upsilon})\)
\\
\begin{tabular}{| c | ccccc  ccccccc|}
\hline
 \(b_{\iota}\)  ~\(\backslash \)~
   \(g_{\kappa}\)/\(h_{\upsilon}\)
 &\(g_{1}\)&\(g_{2}\)&\(g_{3}\)&\(g_{4}\)&\(g_{5}\)&\(h_{1}\)&\(h_{2}\)&\(h_{3}\)&\(h_{4}\) &\(h_{5}\)&\(h_{6}\)&\(h_{7}\)   \\
\hline

 \(b_{1}\) &\(1\)&\(2\)&\(1\)&\(3\)&\(2\) &\(1\)&\(3\)&\(3\)&\(4\)&\(3\)&\(5\)&\(4\)\\
 \(b_{2}\) &\(1\)&\(1\)&\(1\)&\(1\)&\(1\) &\(2\)&\(1\)&\(4\)&\(4\)&\(4\)&\(4\)&\(3\)\\
 \(b_{3}\) &\( \)&\(1\)&\( \)&\(1\)&\( \) &\( \)&\( \)&\(1\)&\(2\)&\(1\)&\(3\)&\(1\)\\
 \(b_{4}\) &\(1\)&\( \)&\(1\)&\( \)&\( \) &\( \)&\( \)&\(2\)&\(1\)&\(1\)&\(3\)&\(1\)\\
 \(b_{5}\) &\(1\)&\(1\)&\( \)&\(1\)&\( \) &\( \)&\( \)&\(2\)&\(2\)&\(1\)&\(3\)&\(1\)\\
 \(b_{6}\) &\(1\)&\(1\)&\(1\)&\( \)&\( \) &\( \)&\( \)&\(1\)&\(1\)&\(1\)&\(1\)&\(1\)\\
 \(b_{7}\) &\(1\)&\(1\)&\(1\)&\(1\)&\( \) &\( \)&\( \)&\(2\)&\(2\)&\(2\)&\(3\)&\(1\)\\

\hline
\end{tabular}
\end{center}

%\newpage
\begin{center}
{\bf Table 5.} Ordinal estimates of movement:
   \( \lambda(g_{\kappa} \rightarrow h_{\upsilon})\),
   \( \lambda(h_{\upsilon} \rightarrow f_{\xi})\),
   \( \lambda(h_{\upsilon} \rightarrow p_{1})\),
   \( \lambda(f_{\xi} \rightarrow p_{1})\)  \\

\begin{tabular}{| c | cccc ccc ccc c|}
\hline

  \(g_{\kappa}\)/\(h_{\upsilon}\) ~\(\backslash \)~
  \(h_{\upsilon}\)/\(f_{\xi}\),\(p_{1}\)
 &\(h_{1}\)&\(h_{2}\)&\(h_{3}\)&\(h_{4}\)&\(h_{5}\)&\(h_{6}\)&\(h_{7}\)&\(f_{1}\)&\(f_{2}\)&\(f_{3}\)  &\(p_{1}\)  \\

 \(\backslash \)~ \(f_{\xi}\)/\(p_{1}\)
 &&&&&&&&&&&  \\

\hline
 \(g_{1}\) &\( \)&\( \)&\(1\)&\(2\)&\(1\)&\(3\)  &\(1\)  &\( \)&\( \)&\( \)  &\( \)\\
 \(g_{2}\) &\( \)&\( \)&\(2\)&\(1\)&\(1\)&\(3\)  &\(1\)  &\( \)&\( \)&\( \)  &\( \)\\
 \(g_{3}\) &\( \)&\( \)&\(3\)&\(3\)&\(1\)&\(3\)  &\(1\)  &\( \)&\( \)&\( \)  &\( \)\\
 \(g_{4}\) &\( \)&\( \)&\(1\)&\(1\)&\(1\)&\(1\)  &\(1\)  &\( \)&\( \)&\( \)  &\( \)\\
 \(g_{5}\) &\( \)&\( \)&\(2\)&\(2\)&\(2\)&\(3\)  &\(1\)  &\( \)&\( \)&\( \)  &\( \)\\

 \(h_{1}\) &\( \)&\( \)&\( \)&\( \)&\( \)&\( \)  &\( \)  &\(2\)&\(3\)&\(2\)  &\(5\)\\
 \(h_{2}\) &\( \)&\( \)&\( \)&\( \)&\( \)&\( \)  &\( \)  &\(1\)&\(1\)&\(1\)  &\(5\)\\
 \(h_{3}\) &\( \)&\( \)&\( \)&\( \)&\( \)&\( \)  &\( \)  &\( \)&\(1\)&\( \)  &\(2\)\\
 \(h_{4}\) &\( \)&\( \)&\( \)&\( \)&\( \)&\( \)  &\( \)  &\(1\)&\( \)&\(1\)  &\(3\)\\
 \(h_{5}\) &\( \)&\( \)&\( \)&\( \)&\( \)&\( \)  &\( \)  &\( \)&\(1\)&\( \)  &\(3\)\\
 \(h_{6}\) &\( \)&\( \)&\( \)&\( \)&\( \)&\( \)  &\( \)  &\(1\)&\(1\)&\(1\)  &\(3\)\\
 \(h_{7}\) &\( \)&\( \)&\( \)&\( \)&\( \)&\( \)  &\( \)  &\(3\)&\(0\)&\(2\)  &\(4\)\\

 \(f_{1}\) &\( \)&\( \)&\( \)&\( \)&\( \)&\( \)  &\( \)  &\( \)&\( \)&\( \)  &\(1\)\\
 \(f_{2}\) &\( \)&\( \)&\( \)&\( \)&\( \)&\( \)  &\( \)  &\( \)&\( \)&\( \)  &\(2\)\\
 \(f_{3}\) &\( \)&\( \)&\( \)&\( \)&\( \)&\( \)  &\( \)  &\( \)&\( \)&\( \)  &\(2\)\\

\hline
\end{tabular}
\end{center}

%~~

 First, a modification of orienteering problem
 (three objective functions, constraint for maximum arc length, constraint for aggregated (summarized) time
 of visited vertices)
 is considered as follows:
 \(H = \{1,...,i,...,n\}\) is the set of vertex/nodes,
 vertex \(1\) is the start point of the route (origin),
  vertex \(n\) is the end (goal) point of the route (destination),
  binary variable \(x_{ij} =1\) if the built route (path) contains arc
  \((i,j)\)
  and \(x_{ij} = 0\) otherwise
  (vertex \(i\) precedes \(j\)),
 \(\theta_{i}\) is the vertex profit,
 \(\lambda_{ij}\) is the arc  cost (if arc \((i,j) \in E\)),
 \(d^{max}\) is a distance constraint for movement between neighbor vertices in the built route/path,
  \(T\) is a time constraint for the built route as summarization of
  times of path vertices.

 The basic model is:
 \[\max \sum_{i=1}^{n} \sum_{j=1}^{n} ~\theta^{1}_{i} x_{ij},
 ~~~\max \sum_{i=1}^{n} \sum_{j=1}^{n} ~\theta^{2}_{i} x_{ij},
 ~~~\max \sum_{i=1}^{n} \sum_{j=1}^{n} ~\theta^{3}_{i} x_{ij},\]
 \[s.t. ~\sum_{j=2}^{n} ~x_{1j}  = \sum_{i=1}^{n-1} ~x_{in}  =1; ~~~
 \sum_{i=2}^{n-1} ~x_{ik}  = \sum_{j=2}^{n-1} ~x_{kj} \leq 1, ~
 k=\overline{2,n-1};\]
 \[\lambda_{ij} x_{ij} \leq d^{max} ~~ \forall i,j;
 ~~~ \sum_{i=1}^{n} \sum_{j=1}^{n} ~\tau_{i} x_{ij} \leq  T ;\]
 \[x_{ij} \in \{0,1\}, ~ i = \overline{1,n}, ~ j=\overline{1,n}.\]
 Here, the Pareto-efficient solutions have to be searched for.
 Clearly, the problem is NP-hard.
 In our case,
 \(a_{1}\) is the start point (i.e., graph vertex),
 \(p_{1}\) is the end (goal) point (i.e., graph vertex).
 The optimization model has to be solved
 for each generalized trajectory above
  (i.e., \(L^{1}\), \(L^{2}\), \(L^{3}\), \(L^{4}\)).

 Let \(L = <l_{1},...,l_{i},...,l_{q}>\)
 be  an admissible route solution
 (i.e., educational trajectory).
 The characteristics of the solution are as follows:
 ~(a) \(\theta^{1} (L) = \sum^{q}_{i=2} \theta^{1}_{i} \)
 is integrated quality of education;
 ~(b) \(\theta^{2} (L) = \sum^{q}_{i=2} \theta^{2}_{i} \)
 is integrated quality of research results
 (including publication results);
  ~(c) \(\theta^{3} (L) = \sum^{q}_{i=2} \theta^{3}_{i} \)
 is integrated parameter of resultant prestige  of the obtained
 academic degrees;
 ~(d)  \(\tau (L) = \sum^{q}_{i=2} \tau_{i} \)
 is integrated required time (years) of the educational
 trajectory;
 ~(e)  \(d(L) = \sum^{q-1}_{i=1}  \lambda (l_{i} \rightarrow l_{i+1})\)
 is integrated  estimate of movement complexity
 (between neighbor educational institutions).
 As a result, the problem can be formulated as the following:

~~

 Find the route \(L\) (solution) such  that

 (1) it fulfils
% satisfies
  two constraints:~
 \(\tau (L) \leq T\) and  \(d(L) \leq d^{max}\);

 (2) it is a Pareto-efficient one by four criteria (objective functions):
 ~\(\max ~\theta^{1} (L)\), ~\(\max ~\theta^{2} (L)\),
 ~\(\max ~\theta^{3} (L)\), ~\(\min ~ d(L)\).

~~

 Note, the usage of educational generalized trajectories
 (\(L^{1}\),  \(L^{2}\), \(L^{3}\), \(L^{4}\))
 leads to simplified/partitioned ``solving space(s)''.
 As a result,
 the optimization problem can be transformed into a version
 of the multicriteria shortest path problem or
 multicriteria multiple choice problem).
 Thus,
  a concurrent'' general solving framework is used (Fig. 19).

\begin{center}
\begin{picture}(108,44)
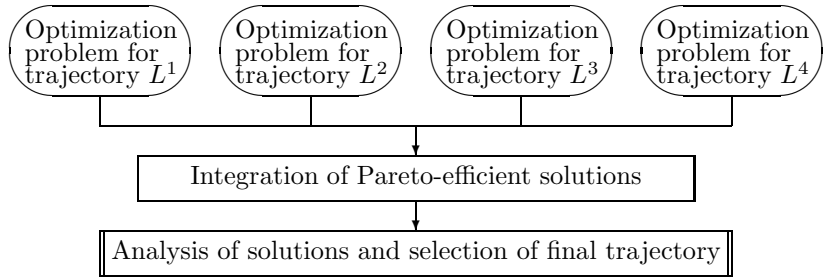
\put(06.2,00){\makebox(0,0)[bl]{Fig. 19. General solving framework
 for educational trajectory}}

%------------- L1
\put(12,36){\oval(24,12)}

\put(02,37){\makebox(0,0)[bl]{Optimization}}
\put(02,34){\makebox(0,0)[bl]{problem for}}
\put(02,31){\makebox(0,0)[bl]{trajectory \(L^{1}\)}}

\put(12,30){\line(0,-1){4}}

%------------- L2
\put(40,36){\oval(24,12)}

\put(30,37){\makebox(0,0)[bl]{Optimization}}
\put(30,34){\makebox(0,0)[bl]{problem for}}
\put(30,31){\makebox(0,0)[bl]{trajectory \(L^{2}\)}}

\put(40,30){\line(0,-1){4}}

%------------- L3
\put(68,36){\oval(24,12)}

\put(58,37){\makebox(0,0)[bl]{Optimization}}
\put(58,34){\makebox(0,0)[bl]{problem for}}
\put(58,31){\makebox(0,0)[bl]{trajectory \(L^{3}\)}}

\put(68,30){\line(0,-1){4}}

%------------- L4
\put(96,36){\oval(24,12)}

\put(86,37){\makebox(0,0)[bl]{Optimization}}
\put(86,34){\makebox(0,0)[bl]{problem for}}
\put(86,31){\makebox(0,0)[bl]{trajectory \(L^{4}\)}}

\put(96,30){\line(0,-1){4}}

%------------------------------

\put(12,26){\line(1,0){84}} \put(54,26){\vector(0,-1){4}}

%--------------------------------

\put(17,16){\line(1,0){74}} \put(17,22){\line(1,0){74}}
\put(17,16){\line(0,1){6}} \put(91,16){\line(0,1){6}}

\put(24,17.6){\makebox(0,0)[bl]{Integration of Pareto-efficient
 solutions}}

\put(54,16){\vector(0,-1){4}}

\put(12,06){\line(1,0){84}} \put(12,12){\line(1,0){84}}
\put(12,06){\line(0,1){6}} \put(96,06){\line(0,1){6}}

\put(12.5,06){\line(0,1){6}} \put(95.5,06){\line(0,1){6}}

\put(13.5,07.6){\makebox(0,0)[bl]{Analysis of solutions and
selection of final trajectory}}

\end{picture}
\end{center}

 In the numerical example,
 the following simplified heuristic solving scheme is considered:

~~

 {\it Stage 1.} Searching for a solution with minimum \(d(L)\)
 for each generalized trajectory
 (\(L^{1}\), \(L^{2}\),\(L^{3}\), \(L^{4}\)).
 The resultant solutions and their estimates are presented in Table 6.

 {\it Stage 2.} Selection of Pareto-efficient solutions:
  ~\(L^{1}_{1} = <a_{1},b_{3},h_{3},p_{1}>\),
  ~\(L^{2}_{1} = <a_{1},b_{1},g_{1},h_{3},p_{1}>\),
  ~\(L^{3}_{1} = <a_{1},b_{2},h_{2},f_{1},p_{1}>\), and
  ~\(L^{4}_{3} = <a_{1},b_{2},g_{2},h_{4},f_{1},p_{1}>\).

 {\it Stage 3.} Selection of the best solution
 (i.e., expert judgment).

~~

 For example, solution
 \(L^{3}_{1} = <a_{1},b_{2},h_{2},f_{1},p_{1}>\)
 can be selected
 while taking into account obtained additional skills
  in  applied mathematics
  (it may be crucial for the future).

\newpage
\begin{center}
{\bf Table 6.} Pareto-efficient solutions and their parameters
 (vector estimate)\\

\begin{tabular}{| c | l |  c c c c c |}
\hline

 No.& Route \(L\)&\(\theta^{1}(L)\)&\(\theta^{2}(L)\)&\(\theta^{3}(L)\)
  &\(\tau(L)\) &\(d(L)\) \\

\hline

 1.& \(L^{1}_{1} = <a_{1},b_{3},h_{3},p_{1}>\)             &\(10\)&\(15\)&\(15\)  &\(8\)&\(7\)\\

%\hline

 2.& \(L^{2}_{1} = <a_{1},b_{1},g_{1},h_{3},p_{1}>\)       &\(15\)&\(17\)&\(19\)  &\(9\)&\(5\)\\

%\hline

 3.& \(L^{3}_{1} = <a_{1},b_{2},h_{2},f_{1},p_{1}>\)       &\(15\)&\(18\)&\(19\)  &\(10\)&\(5\)\\

%\hline

 4.& \(L^{4}_{1} = <a_{1},b_{1},g_{1},h_{7},f_{1},p_{1}>\) &\(19\)&\(21\)&\(23\)  &\(11\)&\(6\)\\
 5.& \(L^{4}_{2} = <a_{1},b_{1},g_{3},h_{3},f_{2},p_{1}>\) &\(19\)&\(21\)&\(23\)  &\(11\)&\(6\)\\
 6.& \(L^{4}_{3} = <a_{1},b_{2},g_{2},h_{4},f_{1},p_{1}>\) &\(19\)&\(22\)&\(24\)  &\(11\)&\(6\)\\

\hline
\end{tabular}
\end{center}

%\subsection{Multi-route DM problems}
\subsection{Scenario planning for start-up company}

 Here, three-stage planning process for a start-up company is
 considered.
%
%%+++++++++++++++++++++++++++++++++++++++++++++++++++
%
 Hierarchical Morphological Multicriteria Design (HMMD)
 based on morphological clique problem is used (combinatorial synthesis)
 (e.g.,
 \cite{lev98,lev06,lev15}).
 A brief description of HMMD (basic version) is the following.
 An examined composite
 (modular) system consists
 of components and their interconnection or compatibility (IC).
 Basic assumptions of HMMD are the following:
 ~(a) a tree-like structure of the system;
 ~(b) a composite estimate for system quality
     that integrates components (subsystems, parts) qualities and
    qualities of IC (compatibility) across subsystems;
 ~(c) monotonic criteria for the system and its components;
 ~(d) quality estimates of system components and IC are evaluated by
% on the basis   of
 coordinated ordinal scales.
 The designations are:
  ~(1) design alternatives (DAs) for
%   leaf
  nodes of the model;
  ~(2) priorities of DAs (\(r=\overline{1,k}\);
      \(1\) corresponds to the best level of quality);
  ~(3) an ordinal compatibility estimate for each pair of DAs
  (\(w=\overline{0,l}\); \(l\) corresponds to the best level of quality).
 The basic phases of HMMD are:
  ~{\it 1.} design of the tree-like system model;
  ~{\it 2.} generation of DAs for leaf nodes of the model;
  ~{\it 3.} hierarchical selection and composing of DAs into composite
    DAs for the corresponding higher level of the system
    hierarchy.
%  ~{\it 4.} analysis and improvement of composite DAs (decisions).

 Let \(S\) be a system consisting of \(m\) parts (components):
 \(P(1),...,P(i),...,P(m)\).
 A set of design alternatives (DAs)
 is generated for each system part above.
 The problem is:

~~

 {\it Find composite design alternative}
 ~ \(S=S(1)\star ...\star S(i)\star ...\star S(m)\)~
% {\it of}~ DAs
 ({\it one representative design alternative} \(S(i)\)
 {\it for each system component/part} ~\(P(i)\), \(i=\overline{1,m}\))
 {\it with non-zero}~ IC
 {\it estimates between the representative design alternatives.}

~~

 A discrete ``space'' of the integrated system excellence is based
 on the following vector:
 ~\(N(S)=(w(S);n(S))\),
 ~where \(w(S)\) is the minimum of pairwise compatibility
 between DAs which correspond to different system components
 (i.e.,
 \(~\forall ~P_{j_{1}}\) and \( P_{j_{2}}\),
 \(1 \leq j_{1} \neq j_{2} \leq m\))
 in \(S\),
 ~\(n(S)=(n_{1},...,n_{r},...n_{k})\),
 ~where \(n_{r}\) is the number of DAs of the \(r\)th quality in \(S\)
 ~(\(\sum^{k}_{r=1} n_{r} = m\)).
 As a result,
 composite decisions which are nondominated by \(N(S)\)
 (i.e., Pareto-efficient solutions)
 are searched for.
 In the numerical example,
 ordinal scale \([1,2,3]\) is used for quality of DAs
 and ordinal scale \([0,1,2,3]\) is used for compatibility.

%%+++++++++++++++++++++++++++++++++++++++++++++++++++

%
 The basic simplified hierarchical structure
 of the considered start-up company is
 (including used DAs):

~~

 {\bf 0.} Hierarchical model ~\(S = P \star T \star M\).

 {\bf 1.} Product ~\(P = A \star B \star E \star W\):

 {\it 1.1.} Models and algorithms ~\(A\):
 prototype(s) \(A_{1}\),
 basic model(s)   \(A_{2}\),
 advanced models \(A_{3}\);

 {\it 1.2.} Algorithms ~\(B\):
 prototype(s) (e.g., simple heuristics) \(B_{1}\),
 library of well-known algorithms   \(B_{2}\),
 extended library of algorithms
 (including advanced algorithms/heuristics) \(B_{3}\);

  {\it 1.3.} Information base of applications ~\(E\):
 none \(E_{1}\),
 simple library of applied examples \(E_{2}\),
 extended library of applied examples
 with educational modes   \(E_{3}\);

  {\it 1.4.} Web-site ~\(W\):
 None \(W_{1}\),
 simplified site-prototype  \(W_{2}\),
 site with an interactive mode(s) for a base of users    \(W_{3}\);

 {\bf 2.} Team  ~\(T = L \star R \star I  \star K\):

 {\it 2.1.} Project leader ~\(L\):
 basic leader \(L_{1}\),
 extended group of leaders   \(L_{2}\);

 {\it 2.2.} Researcher ~\(R\):
 basic researcher (models, algorithms)
  \(R_{1}\),
 extended group of researchers
 (including
 applications in R\&D and engineering,
  educational technology)
  \(R_{2}\);

 {\it 2.3.} Engineer-programmer ~\(I\):
 none \(I_{1}\)
 engineer \(I_{2}\),
 group of engineers   \(I_{3}\),
 extended group of engineers
 (including specialist in Web-design)
  \(I_{4}\);

 {\it 2.4.} Specialist in marketing  ~\(K\):
 none \(K_{1}\),
 basic specialist \(K_{2}\).

 {\bf 3.} Marketing part ~\(M = U \star V\):

 {\it 3.1.} Marketing strategy ~\(U\):
 none \(U_{1}\),
 ``go-to-market''  \(U_{2}\),
% entry into new markets, approach a market, entry into market,
% access to foreign market, niche strategy
% entrance to the market (economics)
 expanding market (e.g., additional market segment(s)) \(U_{3}\).
% market expansion rasshir. graniz rynka
% market development, expanding market

 {\it 3.2.} Market segment ~\(V\):
  none \(V_{1}\),
 education \(V_{2}\),
 R\&D \(V_{3}\).
 engineering \(V_{4}\).
 education and  R\&D \(V_{5} = V_{2} \& V_{3}\).

~~

 Three time stages are examined: ~\(\tau_{1}\), \(\tau_{2}\),
 \(\tau_{3}\).
 The examined trajectory
 ~\(S^{1} \Longrightarrow  S^{2} \Longrightarrow S^{3} \)
 is illustrated in Fig. 20.
 The hierarchical structures above for the time stages
 (including DAs and their ordinal priorities in parentheses for each time stage,
 expert judgment)
 are presented in Fig. 21, Fig. 22, and Fig. 23 (accordantly).
 For stage 1, the initial situation (origin)
 is considered as the following:

 \(S^{1}_{1} = P_{1} \star T_{1} \star M_{1} =
 ( A_{1} \star B_{1} \star E_{1} \star W_{1}) \star
 (L_{1} \star R_{1} \star I_{1} \star K_{1}) \star
 (U_{1} \star V_{1})\).

 Table 7, Table 8, Table 9 contain ordinal estimates of
 compatibility (expert judgment) between DAs.

 The resultant Pareto-efficient composite DAs for system at each time stage
 (on the basis on hierarchical combinatorial synthesis)
% \cite{lev98,lev06,lev15})
 are presented in Fig. 21, Fig. 22, Fig. 23:

 (1) for stage 1 (\(\tau_{1}\)):~
 \(S^{1}_{1} = P^{1}_{1} \star T^{1}_{1} \star M^{1}_{1} =
  ( A_{1} \star B_{1} \star E_{1} \star W_{1}) \star
 (L_{1} \star R_{1} \star I_{1} \star K_{1}) \star
 (U_{1} \star V_{1})\);

 (2) for stage 2 (\(\tau_{2}\)):~
 \(S^{2}_{1} =  P^{2}_{1} \star T^{2}_{1} \star M^{2}_{1} =
  ( A_{2} \star B_{2} \star E_{2} \star W_{2}) \star
 (L_{1} \star R_{2} \star I_{3} \star K_{2}) \star
 (U_{2} \star V_{2})\),

 \(S^{2}_{2} =
   P^{2}_{1} \star T^{2}_{1} \star M^{2}_{1} =
  ( A_{2} \star B_{2} \star E_{2} \star W_{2}) \star
 (L_{1} \star R_{2} \star I_{3} \star K_{2}) \star
 (U_{2} \star V_{3})\);

 (3) for stage 3 (\(\tau_{3}\)):~
 \(S^{3}_{1} =  P^{3}_{1} \star T^{3}_{1} \star M^{3}_{1} =
  ( A_{3} \star B_{3} \star E_{3} \star W_{3}) \star
 (L_{2} \star R_{2} \star I_{3} \star K_{2}) \star
 (U_{2} \star V_{2})\),

 \(S^{3}_{2} = P^{3}_{1} \star T^{3}_{1} \star M^{3}_{1} =
  ( A_{3} \star B_{3} \star E_{3} \star W_{3}) \star
 (L_{2} \star R_{2} \star I_{3} \star K_{2}) \star
 (U_{3} \star V_{3})\).

\begin{center}
\begin{picture}(86,35)
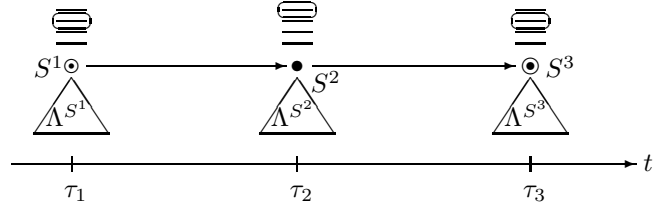
\put(00,00){\makebox(0,0)[bl]{Fig. 20. Three-stage scheme for
 company development}}

%\put(39,22){\oval(76,34)}

\put(84,09){\makebox(0,0)[bl]{\(t\)}}
\put(00,10){\vector(1,0){83}}

\put(08,9){\line(0,1){2}} \put(38,9){\line(0,1){2}}
\put(69,9){\line(0,1){2}}

\put(07,05){\makebox(0,0)[bl]{\(\tau_{1}\)}}
\put(37,05){\makebox(0,0)[bl]{\(\tau_{2}\)}}
\put(68,05){\makebox(0,0)[bl]{\(\tau_{3}\)}}

%------------------  S1
\put(08,23){\circle{1.6}} \put(08,23){\circle*{0.8}}

\put(03,21.5){\makebox(0,0)[bl]{\(S^{1}\)}}

%-- DAs

\put(08,29){\oval(5,2)}

\put(06,30.5){\line(1,0){4}} \put(06,29){\line(1,0){4}}
\put(06,27.5){\line(1,0){4}} \put(06,26){\line(1,0){4}}

%--

\put(03,14){\line(2,3){5}} \put(13,14){\line(-2,3){5}}
\put(03,14){\line(1,0){10}}

\put(04.5,15){\makebox(0,8)[bl]{\(\Lambda^{S^{1}}\)}}

%------------------------------------------- S2
\put(38,23){\circle*{1.5}}

\put(39.8,19.5){\makebox(0,0)[bl]{\(S^{2}\)}}

%---------- DAs
\put(38,30.5){\oval(5,2)}

\put(36,30.5){\line(1,0){4}} \put(36,29){\line(1,0){4}}
\put(36,27.5){\line(1,0){4}} \put(36,26){\line(1,0){4}}

%---------

\put(33,14){\line(2,3){5}} \put(43,14){\line(-2,3){5}}
\put(33,14){\line(1,0){10}}

\put(34.5,15){\makebox(0,8)[bl]{\(\Lambda^{S^{2}}\)}}

%-----------------------------------   S3
\put(69,23){\circle{2.0}} \put(69,23){\circle*{1.2}}

\put(71,21.5){\makebox(0,0)[bl]{\(S^{3}\)}}

%-- DAs

\put(69,29){\oval(5,2)}

\put(67,30.5){\line(1,0){4}} \put(67,29){\line(1,0){4}}
\put(67,27.5){\line(1,0){4}} \put(67,26){\line(1,0){4}}

%-------------

\put(64,14){\line(2,3){5}} \put(74,14){\line(-2,3){5}}
\put(64,14){\line(1,0){10}}

\put(65.5,15){\makebox(0,8)[bl]{\(\Lambda^{S^{3}}\)}}

%===============================================
%%%%%%%%%%%%%%% Arcs of graph

%-- 1-2
\put(10,23){\vector(1,0){026.5}}

%-- 2-3
\put(40,23){\vector(1,0){027}}

\end{picture}
\end{center}

\begin{center}
\begin{picture}(136,40)
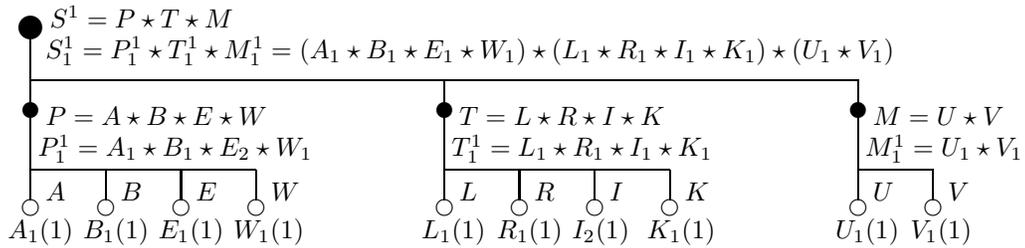
\put(018,00){\makebox(0,0)[bl]{Fig. 21. Hierarchical structure of
 start-up company \(S^{1}\) (stage 1)}}

%>>>>>>>>>>>>>>>>>>>>>>>>>>>>>>>>>>Product-system  PPPP
\put(01,06){\makebox(0,8)[bl]{\(A_{1}(1)\)}}
%\put(01,17){\makebox(0,8)[bl]{\(A_{2}(3)\)}}
%\put(01,13){\makebox(0,8)[bl]{\(A_{3}(1)\)}}

\put(11,06){\makebox(0,8)[bl]{\(B_{1}(1)\)}}
%\put(11,17){\makebox(0,8)[bl]{\(B_{2}(1)\)}}
%\put(11,13){\makebox(0,8)[bl]{\(B_{3}(3)\)}}

\put(21,06){\makebox(0,8)[bl]{\(E_{1}(1)\)}}
%\put(21,17){\makebox(0,8)[bl]{\(E_{2}(1)\)}}
%\put(21,13){\makebox(0,8)[bl]{\(E_{3}(2)\)}}

\put(31,06){\makebox(0,8)[bl]{\(W_{1}(1)\)}}
%\put(31,17){\makebox(0,8)[bl]{\(W_{2}(3)\)}}
%\put(31,13){\makebox(0,8)[bl]{\(W_{3}(1)\)}}

%--

\put(04,16){\line(1,0){30}}

\put(04,16){\line(0,-1){04}} \put(14,16){\line(0,-1){04}}
\put(24,16){\line(0,-1){04}} \put(34,16){\line(0,-1){04}}

%--

\put(04,11){\circle{2}} \put(14,11){\circle{2}}
\put(24,11){\circle{2}} \put(34,11){\circle{2}}

\put(06,12){\makebox(0,8)[bl]{\(A\) }}
\put(16,12){\makebox(0,8)[bl]{\(B\) }}
\put(26,12){\makebox(0,8)[bl]{\(E\) }}
\put(36,12){\makebox(0,8)[bl]{\(W\) }}

%--

\put(04,16){\line(0,1){08}} \put(04,24){\circle*{2}}

\put(06,22){\makebox(0,8)[bl]{\(P = A \star B \star E \star W\)}}

\put(05,17){\makebox(0,8)[bl]{\(P^{1}_{1} = A_{1} \star B_{1}
\star E_{2} \star W_{1}\) }}

%\put(05,32){\makebox(0,8)[bl]{\(P^{1}_{2} = A_{1} \star B_{2}
%\star E_{2} \star W_{4}(3;2,1,1)\) }}

%>>>>>>>>>>>>>>>>>>>>>>>>>>>>>>>>>>>>>>>>>>>>>>>>>>>>>>Team  TTT
\put(56,06){\makebox(0,8)[bl]{\(L_{1}(1)\)}}
%\put(56,17){\makebox(0,8)[bl]{\(L_{2}(3)\)}}

\put(66,06){\makebox(0,8)[bl]{\(R_{1}(1)\)}}
%\put(66,17){\makebox(0,8)[bl]{\(R_{2}(1)\)}}

\put(76,06){\makebox(0,8)[bl]{\(I_{2}(1)\)}}
%\put(76,07){\makebox(0,8)[bl]{\(I_{2}(1)\)}}
%\put(76,13){\makebox(0,8)[bl]{\(I_{3}(2)\)}}
%\put(76,09){\makebox(0,8)[bl]{\(I_{4}(2)\)}}

\put(86,06){\makebox(0,8)[bl]{\(K_{1}(1)\)}}
%\put(86,17){\makebox(0,8)[bl]{\(K_{2}(3)\)}}

%--

\put(59,16){\line(1,0){30}}

\put(59,16){\line(0,-1){04}} \put(69,16){\line(0,-1){04}}
\put(79,16){\line(0,-1){04}} \put(89,16){\line(0,-1){04}}

%--

\put(59,11){\circle{2}} \put(69,11){\circle{2}}
\put(79,11){\circle{2}} \put(89,11){\circle{2}}

\put(61,12){\makebox(0,8)[bl]{\(L\) }}
\put(71,12){\makebox(0,8)[bl]{\(R\) }}
\put(81,12){\makebox(0,8)[bl]{\(I\) }}
\put(91,12){\makebox(0,8)[bl]{\(K\) }}

%--

\put(59,16){\line(0,1){08}} \put(59,24){\circle*{2}}

\put(61,22){\makebox(0,8)[bl]{\(T = L \star R \star I \star K\)}}

\put(60,17){\makebox(0,8)[bl]{\(T^{1}_{1} = L_{1} \star R_{1}
 \star I_{1} \star K_{1}\) }}

%\put(60,17){\makebox(0,8)[bl]{\(T^{1}_{2} = L_{1} \star R_{1}
%\star I_{2} \star K_{1}(3;2,1,1)\) }}

%>>>>>>>>>>>>>>>>>>>>>>>>>>>>>>>>>>>>>>>>>>>>>>>>Marketing Part  MMM
\put(111,06){\makebox(0,8)[bl]{\(U_{1}(1)\)}}
%\put(111,07){\makebox(0,8)[bl]{\(U_{2}(2)\)}}
%\put(111,13){\makebox(0,8)[bl]{\(U_{3}(3)\)}}

\put(121,06){\makebox(0,8)[bl]{\(V_{1}(1)\)}}
%\put(121,07){\makebox(0,8)[bl]{\(V_{2}(2)\)}}
%\put(121,13){\makebox(0,8)[bl]{\(V_{3}(3)\)}}
%\put(121,09){\makebox(0,8)[bl]{\(V_{4}(3)\)}}
%\put(121,05){\makebox(0,8)[bl]{\(V_{5}(3)\)}}

%--

\put(114,16){\line(1,0){10}} \put(114,16){\line(0,-1){04}}
\put(124,16){\line(0,-1){04}}

%--

\put(114,11){\circle{2}} \put(124,11){\circle{2}}

\put(116,12){\makebox(0,8)[bl]{\(U\) }}
\put(126,12){\makebox(0,8)[bl]{\(V\) }}

%--

\put(114,16){\line(0,1){08}} \put(114,24){\circle*{2}}

\put(116,22){\makebox(0,8)[bl]{\(M = U \star V \)}}
\put(115,17){\makebox(0,8)[bl]{\(M^{1}_{1}=U_{1}\star V_{1}\) }}

%\put(115,17){\makebox(0,8)[bl]{\(M^{2}_{2}=U_{2} \star
% V_{2}(3;1,1,0)\) }}

%%%%%%%%%%%%%%%%%%%%%%%%%%%%%%%%%%%%%%%%%%%%%%%%%%%%%%%%%%%%%%%%%%%%

\put(04,24){\line(0,1){04}} \put(59,24){\line(0,1){04}}
\put(114,24){\line(0,1){04}}

\put(04,28){\line(1,0){110}}

\put(04,28){\line(0,1){07}} \put(04,35){\circle*{3}}

\put(06.6,35){\makebox(0,8)[bl]{\(S^{1} = P \star T \star M \)}}

\put(06,30){\makebox(0,8)[bl]{\(S^{1}_{1} = P^{1}_{1} \star
 T^{1}_{1} \star M^{1}_{1} =
  ( A_{1} \star B_{1} \star E_{1} \star W_{1}) \star
 (L_{1} \star R_{1} \star I_{1} \star K_{1}) \star
 (U_{1} \star V_{1})\) }}

%\put(06,30){\makebox(0,8)[bl]{\(S^{1}_{3} = P_{1} \star T_{1}
% \star M_{2}\), ~\(S^{1}_{4} = P_{1} \star T_{2} \star M_{2}\) }}

\end{picture}
\end{center}

\begin{center}
\begin{picture}(136,55)
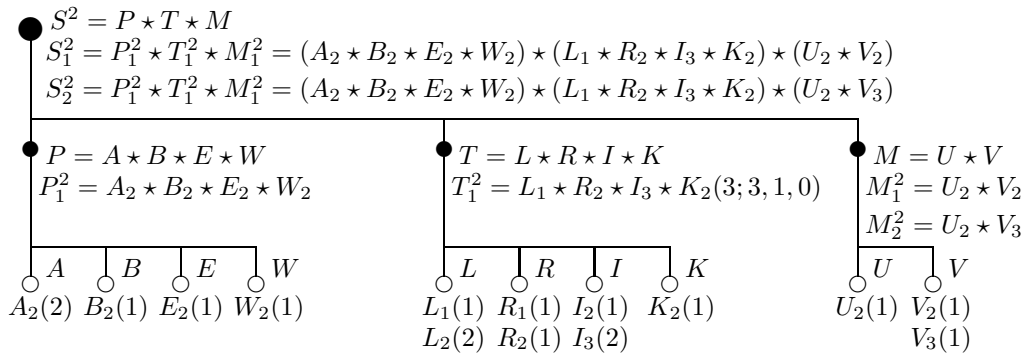
\put(018,00){\makebox(0,0)[bl]{Fig. 22. Hierarchical structure of
 start-up company \(S^{2}\) (stage 2)}}

%>>>>>>>>>>>>>>>>>>>>>>>>>>>>>>>>>>Product-system  PPPP
\put(01,11){\makebox(0,8)[bl]{\(A_{2}(2)\)}}
%\put(01,17){\makebox(0,8)[bl]{\(A_{2}(3)\)}}
%\put(01,13){\makebox(0,8)[bl]{\(A_{3}(1)\)}}

\put(11,11){\makebox(0,8)[bl]{\(B_{2}(1)\)}}
%\put(11,17){\makebox(0,8)[bl]{\(B_{2}(1)\)}}
%\put(11,13){\makebox(0,8)[bl]{\(B_{3}(3)\)}}

\put(21,11){\makebox(0,8)[bl]{\(E_{2}(1)\)}}
%\put(21,07){\makebox(0,8)[bl]{\(E_{3}(2)\)}}
%\put(21,13){\makebox(0,8)[bl]{\(E_{3}(2)\)}}

\put(31,11){\makebox(0,8)[bl]{\(W_{2}(1)\)}}
%\put(31,07){\makebox(0,8)[bl]{\(W_{2}(1)\)}}
%\put(31,13){\makebox(0,8)[bl]{\(W_{3}(1)\)}}

%--

\put(04,21){\line(1,0){30}}

\put(04,21){\line(0,-1){04}} \put(14,21){\line(0,-1){04}}
\put(24,21){\line(0,-1){04}} \put(34,21){\line(0,-1){04}}

%--

\put(04,16){\circle{2}} \put(14,16){\circle{2}}
\put(24,16){\circle{2}} \put(34,16){\circle{2}}

\put(06,17){\makebox(0,8)[bl]{\(A\) }}
\put(16,17){\makebox(0,8)[bl]{\(B\) }}
\put(26,17){\makebox(0,8)[bl]{\(E\) }}
\put(36,17){\makebox(0,8)[bl]{\(W\) }}

%--

\put(04,21){\line(0,1){13}} \put(04,34){\circle*{2}}

\put(06,32){\makebox(0,8)[bl]{\(P = A \star B \star E \star W\)}}

\put(05,27){\makebox(0,8)[bl]{\(P^{2}_{1} = A_{2} \star B_{2}
\star E_{2} \star W_{2}\) }}

%\put(05,22){\makebox(0,8)[bl]{\(P^{2}_{2} = A_{2} \star B_{2}
%\star E_{2} \star W_{2}(3;2,1,1)\) }}

%>>>>>>>>>>>>>>>>>>>>>>>>>>>>>>>>>>>>>>>>>>>>>>>>>>>>>>Team  TTT
\put(56,11){\makebox(0,8)[bl]{\(L_{1}(1)\)}}
\put(56,07){\makebox(0,8)[bl]{\(L_{2}(2)\)}}

\put(66,11){\makebox(0,8)[bl]{\(R_{1}(1)\)}}
\put(66,07){\makebox(0,8)[bl]{\(R_{2}(1)\)}}

\put(76,11){\makebox(0,8)[bl]{\(I_{2}(1)\)}}
\put(76,07){\makebox(0,8)[bl]{\(I_{3}(2)\)}}
%\put(76,03){\makebox(0,8)[bl]{\(I_{3}(2)\)}}
%\put(76,09){\makebox(0,8)[bl]{\(I_{4}(2)\)}}

\put(86,11){\makebox(0,8)[bl]{\(K_{2}(1)\)}}
%\put(86,17){\makebox(0,8)[bl]{\(K_{2}(3)\)}}

%--

\put(59,21){\line(1,0){30}}

\put(59,21){\line(0,-1){04}} \put(69,21){\line(0,-1){04}}
\put(79,21){\line(0,-1){04}} \put(89,21){\line(0,-1){04}}

%--

\put(59,16){\circle{2}} \put(69,16){\circle{2}}
\put(79,16){\circle{2}} \put(89,16){\circle{2}}

\put(61,17){\makebox(0,8)[bl]{\(L\) }}
\put(71,17){\makebox(0,8)[bl]{\(R\) }}
\put(81,17){\makebox(0,8)[bl]{\(I\) }}
\put(91,17){\makebox(0,8)[bl]{\(K\) }}

%--

\put(59,21){\line(0,1){13}} \put(59,34){\circle*{2}}

\put(61,32){\makebox(0,8)[bl]{\(T = L \star R \star I \star K\)}}

\put(60,27){\makebox(0,8)[bl]{\(T^{2}_{1} = L_{1} \star R_{2}
\star I_{3} \star K_{2}(3;3,1,0) \) }}

%\put(60,22){\makebox(0,8)[bl]{\(T^{2}_{2} = L_{4} \star R_{2}
%\star I_{4} \star K_{2}(3;2,1,1)\) }}

%>>>>>>>>>>>>>>>>>>>>>>>>>>>>>>>>>>>>>>>>>>>>>>>>Marketing Part  MMM
\put(111,11){\makebox(0,8)[bl]{\(U_{2}(1)\)}}
%\put(111,17){\makebox(0,8)[bl]{\(U_{2}(3)\)}}
%\put(111,13){\makebox(0,8)[bl]{\(U_{3}(3)\)}}

\put(121,11){\makebox(0,8)[bl]{\(V_{2}(1)\)}}
\put(121,07){\makebox(0,8)[bl]{\(V_{3}(1)\)}}
%\put(121,13){\makebox(0,8)[bl]{\(V_{3}(1)\)}}
%\put(121,09){\makebox(0,8)[bl]{\(V_{4}(1)\)}}
%\put(121,05){\makebox(0,8)[bl]{\(V_{5}(1)\)}}

%--

\put(114,21){\line(1,0){10}} \put(114,21){\line(0,-1){04}}
\put(124,21){\line(0,-1){04}}

%--

\put(114,16){\circle{2}} \put(124,16){\circle{2}}

\put(116,17){\makebox(0,8)[bl]{\(U\) }}
\put(126,17){\makebox(0,8)[bl]{\(V\) }}

%--

\put(114,21){\line(0,1){13}} \put(114,34){\circle*{2}}

\put(116,32){\makebox(0,8)[bl]{\(M = U \star V \)}}
\put(115,27){\makebox(0,8)[bl]{\(M^{2}_{1}=U_{2}\star V_{2}\) }}

\put(115,22){\makebox(0,8)[bl]{\(M^{2}_{2}=U_{2} \star
 V_{3}\) }}

%%%%%%%%%%%%%%%%%%%%%%%%%%%%%%%%%%%%%%%%%%%%%%%%%%%%%%%%%%%%%%%%%%%%

\put(04,34){\line(0,1){04}} \put(59,34){\line(0,1){04}}
\put(114,34){\line(0,1){04}}

\put(04,38){\line(1,0){110}}

\put(04,38){\line(0,1){12}} \put(04,50){\circle*{3}}

\put(06.6,50){\makebox(0,8)[bl]{\(S^{2} = P \star T \star M \)}}

%\put(06,50){\makebox(0,8)[bl]{\(S^{2}_{1}=P^{2}_{2} \star
% T^{2}_{2} \star M^{2}_{1}\) }}

\put(06,45){\makebox(0,8)[bl]{\(S^{2}_{1} =
  P^{2}_{1} \star T^{2}_{1} \star M^{2}_{1} =
  ( A_{2} \star B_{2} \star E_{2} \star W_{2}) \star
 (L_{1} \star R_{2} \star I_{3} \star K_{2}) \star
 (U_{2} \star V_{2})\) }}

 \put(06,40){\makebox(0,8)[bl]{\(S^{2}_{2} =
   P^{2}_{1} \star T^{2}_{1} \star M^{2}_{1} =
  ( A_{2} \star B_{2} \star E_{2} \star W_{2}) \star
 (L_{1} \star R_{2} \star I_{3} \star K_{2}) \star
 (U_{2} \star V_{3})\) }}

\end{picture}
\end{center}

\begin{center}
\begin{picture}(154,65)
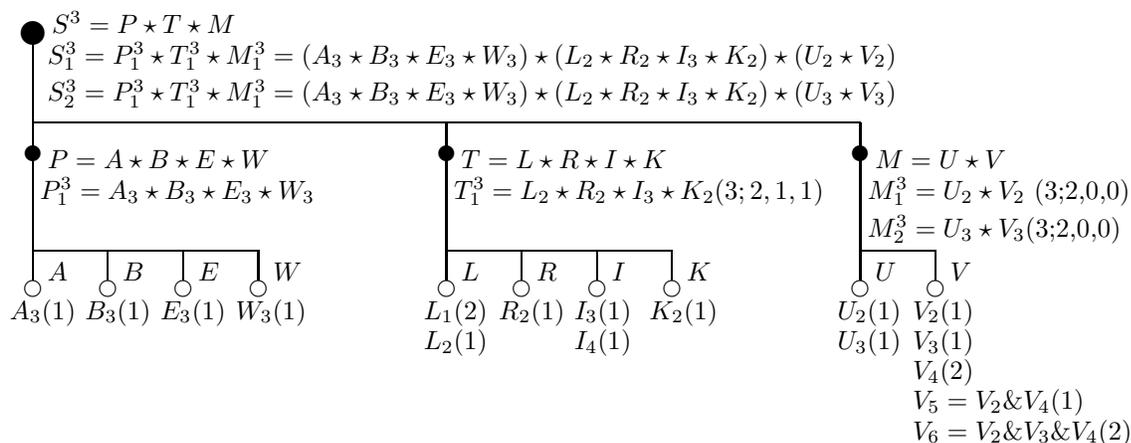
\put(024,00){\makebox(0,0)[bl]{Fig. 23. Hierarchical structure of
 start-up company \(S^{3}\) (stage 3)}}

%>>>>>>>>>>>>>>>>>>>>>>>>>>>>>>>>>>Product-system  PPPP
\put(01,21){\makebox(0,8)[bl]{\(A_{3}(1)\)}}
%\put(01,17){\makebox(0,8)[bl]{\(A_{2}(3)\)}}
%\put(01,13){\makebox(0,8)[bl]{\(A_{3}(1)\)}}

\put(11,21){\makebox(0,8)[bl]{\(B_{3}(1)\)}}
%\put(11,17){\makebox(0,8)[bl]{\(B_{2}(1)\)}}
%\put(11,13){\makebox(0,8)[bl]{\(B_{3}(3)\)}}

\put(21,21){\makebox(0,8)[bl]{\(E_{3}(1)\)}}
%\put(21,17){\makebox(0,8)[bl]{\(E_{2}(1)\)}}
%\put(21,13){\makebox(0,8)[bl]{\(E_{3}(2)\)}}

\put(31,21){\makebox(0,8)[bl]{\(W_{3}(1)\)}}
%\put(31,17){\makebox(0,8)[bl]{\(W_{3}(1)\)}}
%\put(31,13){\makebox(0,8)[bl]{\(W_{3}(1)\)}}

%--

\put(04,31){\line(1,0){30}}

\put(04,31){\line(0,-1){04}} \put(14,31){\line(0,-1){04}}
\put(24,31){\line(0,-1){04}} \put(34,31){\line(0,-1){04}}

%--

\put(04,26){\circle{2}} \put(14,26){\circle{2}}
\put(24,26){\circle{2}} \put(34,26){\circle{2}}

\put(06,27){\makebox(0,8)[bl]{\(A\) }}
\put(16,27){\makebox(0,8)[bl]{\(B\) }}
\put(26,27){\makebox(0,8)[bl]{\(E\) }}
\put(36,27){\makebox(0,8)[bl]{\(W\) }}

%--

\put(04,31){\line(0,1){13}} \put(04,44){\circle*{2}}

\put(06,42){\makebox(0,8)[bl]{\(P = A \star B \star E \star W\)}}

\put(05,37){\makebox(0,8)[bl]{\(P^{3}_{1} = A_{3} \star B_{3}
\star E_{3} \star W_{3}\) }}

%\put(05,32){\makebox(0,8)[bl]{\(P^{3}_{2} = A_{3} \star B_{3}
%\star E_{3} \star W_{3}(3;2,1,1)\) }}

%>>>>>>>>>>>>>>>>>>>>>>>>>>>>>>>>>>>>>>>>>>>>>>>>>>>>>>Team  TTT
\put(56,21){\makebox(0,8)[bl]{\(L_{1}(2)\)}}
\put(56,17){\makebox(0,8)[bl]{\(L_{2}(1)\)}}

\put(66,21){\makebox(0,8)[bl]{\(R_{2}(1)\)}}
%\put(66,17){\makebox(0,8)[bl]{\(R_{2}(1)\)}}

\put(76,21){\makebox(0,8)[bl]{\(I_{3}(1)\)}}
\put(76,17){\makebox(0,8)[bl]{\(I_{4}(1)\)}}
%\put(76,13){\makebox(0,8)[bl]{\(I_{3}(2)\)}}
%\put(76,09){\makebox(0,8)[bl]{\(I_{4}(2)\)}}

\put(86,21){\makebox(0,8)[bl]{\(K_{2}(1)\)}}
%\put(86,17){\makebox(0,8)[bl]{\(K_{2}(3)\)}}

%--

\put(59,31){\line(1,0){30}}

\put(59,31){\line(0,-1){04}} \put(69,31){\line(0,-1){04}}
\put(79,31){\line(0,-1){04}} \put(89,31){\line(0,-1){04}}

%--

\put(59,26){\circle{2}} \put(69,26){\circle{2}}
\put(79,26){\circle{2}} \put(89,26){\circle{2}}

\put(61,27){\makebox(0,8)[bl]{\(L\) }}
\put(71,27){\makebox(0,8)[bl]{\(R\) }}
\put(81,27){\makebox(0,8)[bl]{\(I\) }}
\put(91,27){\makebox(0,8)[bl]{\(K\) }}

%--

\put(59,31){\line(0,1){13}} \put(59,44){\circle*{2}}

\put(61,42){\makebox(0,8)[bl]{\(T = L \star R \star I \star K\)}}

\put(60,37){\makebox(0,8)[bl]{\(T^{3}_{1} = L_{2} \star R_{2}
\star I_{3} \star K_{2}(3;2,1,1) \) }}

%\put(60,32){\makebox(0,8)[bl]{\(T^{3}_{2} = L_{4} \star R_{2}
%\star I_{4} \star K_{2}(3;2,1,1)\) }}

%>>>>>>>>>>>>>>>>>>>>>>>>>>>>>>>>>>>>>>>>>>>>>>>>Marketing Part  MMM
\put(111,21){\makebox(0,8)[bl]{\(U_{2}(1)\)}}
\put(111,17){\makebox(0,8)[bl]{\(U_{3}(1)\)}}
%\put(111,13){\makebox(0,8)[bl]{\(U_{3}(3)\)}}

\put(121,21){\makebox(0,8)[bl]{\(V_{2}(1)\)}}
\put(121,17){\makebox(0,8)[bl]{\(V_{3}(1)\)}}
\put(121,13){\makebox(0,8)[bl]{\(V_{4}(2)\)}}
\put(121,09){\makebox(0,8)[bl]{\(V_{5}=V_{2}\&V_{4}(1)\)}}
%\put(121,05){\makebox(0,8)[bl]{\(V_{5}(1)\)}}
\put(121,05){\makebox(0,8)[bl]{\(V_{6}=V_{2}\&V_{3}\&V_{4}(2)\)}}

%--

\put(114,31){\line(1,0){10}} \put(114,31){\line(0,-1){04}}
\put(124,31){\line(0,-1){04}}

%--

\put(114,26){\circle{2}} \put(124,26){\circle{2}}

\put(116,27){\makebox(0,8)[bl]{\(U\) }}
\put(126,27){\makebox(0,8)[bl]{\(V\) }}

%--

\put(114,31){\line(0,1){13}} \put(114,44){\circle*{2}}

\put(116,42){\makebox(0,8)[bl]{\(M = U \star V \)}}
\put(115,37){\makebox(0,8)[bl]{\(M^{3}_{1}=U_{2}\star V_{2}\)
 (3;2,0,0) }}

\put(115,32){\makebox(0,8)[bl]{\(M^{3}_{2}=U_{3} \star
  V_{3}\)(3;2,0,0)}}

%%%%%%%%%%%%%%%%%%%%%%%%%%%%%%%%%%%%%%%%%%%%%%%%%%%%%%%%%%%%%%%%%%%%

\put(04,44){\line(0,1){04}} \put(59,44){\line(0,1){04}}
\put(114,44){\line(0,1){04}}

\put(04,48){\line(1,0){110}}

\put(04,48){\line(0,1){12}} \put(04,60){\circle*{3}}

\put(06.6,60){\makebox(0,8)[bl]{\(S^{3} = P \star T \star M \)}}

%\put(06,60){\makebox(0,8)[bl]{\(S^{3}_{1}=P^{3}_{2} \star
% T^{3}_{2} \star M^{3}_{1}\) }}

\put(06,55){\makebox(0,8)[bl]{\(S^{3}_{1} =
  P^{3}_{1} \star T^{3}_{1} \star M^{3}_{1} =
  ( A_{3} \star B_{3} \star E_{3} \star W_{3}) \star
 (L_{2} \star R_{2} \star I_{3} \star K_{2}) \star
 (U_{2} \star V_{2})\) }}

 \put(06,50){\makebox(0,8)[bl]{\(S^{3}_{2} =
   P^{3}_{1} \star T^{3}_{1} \star M^{3}_{1} =
  ( A_{3} \star B_{3} \star E_{3} \star W_{3}) \star
 (L_{2} \star R_{2} \star I_{3} \star K_{2}) \star
 (U_{3} \star V_{3})\) }}

\end{picture}
\end{center}

\begin{center}
{\bf Table 7.} Compatibility  \\
\begin{tabular}{| l |  c c c c c c c c c|}
\hline
 &
  \(B_{1}\)&\(B_{2}\)&\(B_{3}\)&
   \(E_{1}\)&\(E_{2}\)&\(E_{3}\)&\(W_{1}\)&\(W_{2}\) &\(W_{3}\) \\

%-------------------------------
\hline

 \(A_{1}\) &\(3\)&\(0\)&\(0\) &\(3\)&\(3\)&\(1\)&\(3\)&\(2\)&\(0\) \\
 \(A_{2}\) &\(0\)&\(3\)&\(0\) &\(0\)&\(3\)&\(3\)&\(0\)&\(3\)&\(2\)\\
 \(A_{3}\) &\(0\)&\(0\)&\(3\) &\(0\)&\(2\)&\(3\)&\(0\)&\(3\)&\(1\) \\

 \(B_{1}\)  & &&& \(3\)&\(0\)&\(0\)  &\(3\)&\(3\)&\(2\) \\
 \(B_{2}\)  & &&& \(0\)&\(3\)&\(3\)  &\(3\)&\(3\)&\(3\) \\
 \(B_{3}\)  & &&& \(0\)&\(3\)&\(3\)  &\(3\)&\(0\)&\(3\) \\

 \(E_{1}\)  & &&& &&& \(3\)&\(0\)&\(0\) \\
 \(E_{2}\)  & &&& &&& \(3\)&\(3\)&\(0\) \\
 \(E_{3}\)  & &&& &&& \(0\)&\(0\)&\(3\) \\

\hline
\end{tabular}
\end{center}

\begin{center}
{\bf Table 8.} Compatibility  \\
\begin{tabular}{| l | c c c c c c|}
\hline
 &  \(V_{1}\)&\(V_{2}\)&\(V_{3}\)&\(V_{4}\)&\(V_{5}\)&\(V_{6}\)\\
%-------------------------------
\hline
 \(U_{1}\) &\(3\)&\(0\)&\(0\)&\(0\)&\(0\)&\(0\)\\
 \(U_{2}\) &\(0\)&\(3\)&\(2\)&\(2\)&\(1\)&\(1\)\\
 \(U_{3}\) &\(0\)&\(2\)&\(3\)&\(2\)&\(1\)&\(1\)\\
\hline
\end{tabular}
\end{center}

%\newpage
\begin{center}
{\bf Table 9.} Compatibility  \\
\begin{tabular}{| l |  c c c c c c c c|}
\hline
 &
 \(R_{1}\)&\(R_{2}\)&\(I_{1}\)& \(I_{2}\)&\(I_{3}\)&\(I_{4}\)& \(K_{1}\)&\(K_{2}\)\\

%-------------------------------
\hline

 \(L_{1}\) &\(3\)&\(3\)  &\(2\)&\(3\)&\(3\)&\(2\)  &\(3\)&\(3\) \\
 \(L_{2}\) &\(2\)&\(3\)  &\(0\)&\(0\)&\(3\)&\(3\)  &\(0\)&\(3\) \\

 \(R_{1}\) & &&  \(2\)&\(3\)&\(2\)&\(1\)  &\(3\)&\(3\)\\
 \(R_{2}\) & &&  \(0\)&\(2\)&\(3\)&\(3\)  &\(0\)&\(3\)\\

 \(I_{1}\) & &&& &&& \(3\)&\(0\) \\
 \(I_{2}\) & &&& &&& \(0\)&\(3\) \\
 \(I_{3}\) & &&& &&& \(0\)&\(3\) \\
 \(I_{4}\) & &&& &&& \(0\)&\(3\) \\

\hline
\end{tabular}
\end{center}

 Table 10 contains ordinal estimates of compatibility
 (expert judgment) between
 DAs for the composite system at time stages.
 The final Pareto-efficient system trajectory is
 (hierarchical combinatorial synthesis) (Fig. 24):
 ~\(\alpha = < S^{1}_{1},  S^{2}_{1}, S^{3}_{1}  > \).

\begin{center}
{\bf Table 10.} Compatibility  \\
\begin{tabular}{| l | c c c c |}
\hline
 &
  \(S^{2}_{1}\)&\(S^{2}_{2}\)&\(S^{3}_{1}\)&\(S^{3}_{2}\)\\

%-------------------------------
\hline

 \(S^{1}_{1}\)  &\(3\)&\(0\) &\(3\)&\(0\) \\

%-----------

  \(S^{2}_{1}\)  &\( \)&\( \) &\(3\)&\(2\)\\
  \(S^{2}_{2}\)  &\( \)&\( \) &\(3\)&\(3\) \\

\hline
\end{tabular}
\end{center}

\begin{center}
\begin{picture}(72,22)
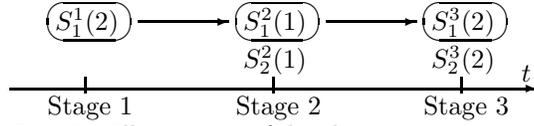
\put(00,0){\makebox(0,0)[bl] {Fig. 24. Illustration of development
 trajectory}}

\put(0,8){\vector(1,0){70}} \put(68,9){\makebox(0,8)[bl]{\(t\)}}

\put(5,4){\makebox(0,8)[bl]{Stage 1}}
\put(30,4){\makebox(0,8)[bl]{Stage 2}}
\put(55,4){\makebox(0,8)[bl]{Stage 3}}

\put(10,7.5){\line(0,1){2}} \put(35,7.5){\line(0,1){2}}
\put(60,7.5){\line(0,1){2}}

%\put(01,21){\makebox(0,8)[bl]{\(\alpha :\)}}
%\put(0,25){\makebox(0,8)[bl]{\(\alpha^{2}:\)}}
%\put(0,20){\makebox(0,8)[bl]{\(\alpha^{''}:\)}}

%\put(13.5,22){\vector(1,0){17.6}}

%\put(38,22){\vector(1,0){18.5}}

%\put(6,20){\makebox(0,8)[bl]{\(M^{1}_{1}(1)\)}}
\put(6,15){\makebox(0,8)[bl]{\(S^{1}_{1}(2)\)}}
%\put(6,15){\makebox(0,8)[bl]{\(S^{1}_{5}(3)\)}}
%\put(6,10){\makebox(0,8)[bl]{\(S^{1}_{6}(3)\)}}

\put(10.5,17){\oval(11,5)}

\put(17,17){\vector(1,0){12}}

%\put(17,16){\vector(4,-1){12}}

%======================================

\put(31,15){\makebox(0,8)[bl]{\(S^{2}_{1}(1)\)}}
\put(31,10){\makebox(0,8)[bl]{\(S^{2}_{2}(1)\)}}
%\put(31,10){\makebox(0,8)[bl]{\(M^{2}_{3}(2)\)}}

\put(35.5,17){\oval(11,5)}

\put(42,17){\vector(1,0){12}}

%\put(42,12){\vector(4,1){12}}

%==============================

%\put(56,20){\makebox(0,8)[bl]{\(M^{3}_{1}(1)\)}}
\put(56,15){\makebox(0,8)[bl]{\(S^{3}_{1}(2)\)}}
\put(56,10){\makebox(0,8)[bl]{\(S^{3}_{2}(2)\)}}

\put(60.5,17){\oval(11,5)}

\end{picture}
\end{center}

% Firm (Start-Up company) development planning - several goals
% Multi-goal and multi-part scenario planning
% for a start-up company development process.

\subsection{Simplified example for planning of medical treatment}
%
%%%%%%%%%%%%%%%%%%%%%%%%%%%%%%%%%%%
%
%%%%%%%%%%%%%%%%%%%%%%%%%%%%%%%%%%% Medicine
%
 Generally,
 there exists a significant  problem in medicine (Fig. 25):
 planning of medical treatment (as a treatment route)
 for a certain patient
 (e.g., \cite{lev06,levsok04})
 or joint planning/designing a route of dignosis/treatment operations
 (e.g., \cite{lev12hier,lev15}).
%
% Here, support problems are the following:
% (i) analysis/diagnosis of initial situation (i.e., point),
% (ii) design of route ``space''
% (i.e., a set of treatment operations
% or a joint set of diagnosis/treament operations),
% (iii) design of route,
% (iv) online analysis/diagnosis of the route implementation and
% online modification (correction) of the route.
%

 %
 A two-phase scheme of medical treatment planning and
 implementation
% (including support layer)
 is depicted in Fig. 26
 (basic flow-chart that can involve online modes, support layer).

%%%%%%%%%%%%%%%%%%%%%%%%%%%%%%%%%%%%%%%%%%%%%%%%%%%%%%%%%%
%\section{Example of Tree-like Structure for Medical Treatment}
%\subsection{Tree-like Trajectory for Medical Treatment}

\begin{center}
\begin{picture}(76,54)
%\begin{picture}(85,54)

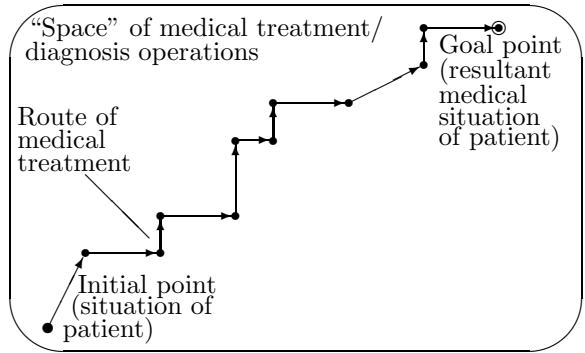
\put(01,00){\makebox(0,0)[bl]{Fig. 25.  Routing
 as medical treatment planning}}

\put(02.4,48){\makebox(0,0)[bl]{``Space'' of medical treatment/}}
\put(02,45){\makebox(0,0)[bl]{diagnosis operations}}

\put(38,30){\oval(76,46)}

%------------------ Goal point

\put(65,50){\circle{1.7}} \put(65,50){\circle*{1.2}}

\put(57,46){\makebox(0,0)[bl]{Goal point}}
\put(57,42.5){\makebox(0,0)[bl]{(resultant}}
\put(57,40){\makebox(0,0)[bl]{medical}}
\put(57,37){\makebox(0,0)[bl]{situation}}
\put(57,33.5){\makebox(0,0)[bl]{of patient)}}

%------------ Path

\put(01,37){\makebox(0,0)[bl]{Route of}}
\put(01,34){\makebox(0,0)[bl]{medical}}
\put(01,31){\makebox(0,0)[bl]{treatment}}
\put(10,30.5){\line(1,-1){08.5}}

\put(55,50){\vector(1,0){9.5}}

\put(55,45){\vector(0,1){4.5}} \put(55,50){\circle*{1.0}}

\put(45,40){\vector(2,1){9.5}} \put(55,45){\circle*{1.0}}

\put(35,40){\vector(1,0){9.5}} \put(45,40){\circle*{1.0}}

\put(35,35){\vector(0,1){4.5}} \put(35,40){\circle*{1.0}}

\put(30,35){\vector(1,0){4.5}} \put(35,35){\circle*{1.0}}

\put(30,25){\vector(0,1){9.5}} \put(30,35){\circle*{1.0}}

\put(20,25){\vector(1,0){9.5}} \put(30,25){\circle*{1.0}}

\put(20,20){\vector(0,1){04.5}} \put(20,25){\circle*{1.0}}

\put(10,20){\vector(1,0){9.5}} \put(20,20){\circle*{1.0}}

\put(05,10){\vector(1,2){4.7}} \put(10,20){\circle*{1.0}}

%------------------ Initial point

\put(05,10){\circle*{1.4}}

\put(09,14){\makebox(0,0)[bl]{Initial point}}
\put(08,11){\makebox(0,0)[bl]{(situation of}}
\put(07,08){\makebox(0,0)[bl]{patient)}}

\end{picture}
\end{center}

 In Fig. 26,
 a general networked framework of medical treatment
 (diagnosis, desing/planning,
 implementation)
% (including support layer)
 is presented.

\begin{center}
\begin{picture}(147,74)
\put(21,00){\makebox(0,0)[bl]{Fig. 26. General networked framework
 of medical treatment}}

%++++++++++++++++++++++++++++++++++++

\put(02,47){\oval(04,04)}

\put(02,45){\line(0,-1){06}}

\put(00,43){\line(1,0){04}}

\put(02,39){\line(-1,-2){02}} \put(02,39){\line(1,-2){02}}

%--------------------------

\put(08.5,47){\makebox(0,8)[bl]{Analysis/}}
\put(08.5,43){\makebox(0,8)[bl]{diagnosis}}
\put(09.5,39){\makebox(0,8)[bl]{~~~~of}}
\put(10,35){\makebox(0,8)[bl]{patient}}

\put(08,34){\line(1,0){15}}\put(08,51){\line(1,0){15}}
\put(08,34){\line(0,1){17}}\put(23,34){\line(0,1){17}}

\put(23,44.5){\vector(1,2){4}}

\put(23,42.5){\vector(1,0){4}}

\put(23,40.5){\vector(1,-3){4}}

%========================= Hierarchies, DAs, compatibility

\put(24,70){\makebox(0,8)[bl]{PRELIMINARY}}
\put(27,67){\makebox(0,8)[bl]{DIAGNOSIS }}
\put(31,64){\makebox(0,8)[bl]{PHASE: }}

\put(60,70){\makebox(0,8)[bl]{BASIC}}
\put(55,67){\makebox(0,8)[bl]{TREATMENT}}
\put(60,64){\makebox(0,8)[bl]{PHASE: }}

%========================= Hierarchies, DAs, compatibility

\put(95,70){\makebox(0,8)[bl]{ADDITIONAL}}
\put(95,67){\makebox(0,8)[bl]{TREATMENT}}
\put(100,64){\makebox(0,8)[bl]{PHASE:}}

%----------------------

\put(32,58.5){\makebox(0,8)[bl]{Non-}}
\put(29,55.5){\makebox(0,8)[bl]{standard}}
\put(29,52.5){\makebox(0,8)[bl]{situation}}

\put(27,51){\line(1,0){19}} \put(27,62){\line(1,0){19}}
\put(27,51){\line(0,1){11}} \put(46,51){\line(0,1){11}}

\put(46,57){\vector(1,0){04}}

%--

\put(85,57){\vector(1,0){05}}

\put(67.5,57){\oval(35,08)}

\put(57,57.5){\makebox(0,8)[bl]{Non-standard}}
\put(53.5,54){\makebox(0,8)[bl]{medical treatment}}

%--

%\put(124,57){\vector(1,0){04}}

\put(107,57){\oval(34,10)}

\put(99,59){\makebox(0,8)[bl]{Additional}}
\put(97,56){\makebox(0,8)[bl]{non-standard}}
\put(93,53){\makebox(0,8)[bl]{medical treatment}}

%>>>>>>>>>>>>>>>>>>>>>>>>>>>>>>>>>>>>>>>>>>>>>>>>
%----------------------

\put(29,42.5){\makebox(0,8)[bl]{Standard}}
\put(28,39.5){\makebox(0,8)[bl]{situation 1}}

\put(27,38){\line(1,0){19}} \put(27,46){\line(1,0){19}}
\put(27,38){\line(0,1){08}} \put(46,38){\line(0,1){08}}

\put(46,42){\vector(1,0){04}}

%--

\put(32.6,35){\makebox(0,8)[bl]{{\bf . . .}}}
%--

\put(85,42){\vector(1,0){05}}

\put(67.5,42){\oval(35,08)}

\put(54,42.5){\makebox(0,8)[bl]{Standard medical}}
\put(53,39){\makebox(0,8)[bl]{treatment (type \(1\))}}

%--

%\put(124,42){\vector(1,0){04}}

\put(107,42){\oval(34,08)}

\put(91,42.5){\makebox(0,8)[bl]{Additional standard}}
\put(92,39){\makebox(0,8)[bl]{treatment (type \(1\))}}

%----

\put(64,35){\makebox(0,8)[bl]{{\bf . . .}}}
\put(103.5,35){\makebox(0,8)[bl]{{\bf . . .}}}

%----------------------

\put(29,29.5){\makebox(0,8)[bl]{Standard}}
\put(28,26.5){\makebox(0,8)[bl]{situation N}}

\put(27,25){\line(1,0){19}} \put(27,33){\line(1,0){19}}
\put(27,25){\line(0,1){08}} \put(46,25){\line(0,1){08}}

\put(46,29){\vector(1,0){04}}

%--

\put(85,29){\vector(1,0){05}}

\put(67.5,29){\oval(35,08)}

\put(54,29.5){\makebox(0,8)[bl]{Standard medical}}
\put(53,26){\makebox(0,8)[bl]{treatment (type \(N\))}}

%--

%\put(124,29){\vector(1,0){04}}

\put(107,29){\oval(34,08)}

\put(91,29.5){\makebox(0,8)[bl]{Additional standard}}
\put(92,26){\makebox(0,8)[bl]{treatment (type \(N\))}}

%>>>>>>>>>>>>>>>>>>>>>>>>>>>>>>>>>>>>>>>>>>>>>>>>
%%%%%%%%%%%%%%%%%%%%%%%%%%%%%%%%%%%%

\put(38,21){\line(1,0){08}}\put(48,21){\line(1,0){08}}
\put(58,21){\line(1,0){08}}\put(69,21){\line(1,0){07}}
\put(78,21){\line(1,0){08}}\put(88,21){\line(1,0){08}}
\put(98,21){\line(1,0){08}}\put(108,21){\line(1,0){08}}
\put(118,21){\line(1,0){08}}

%\put(128,21){\line(1,0){08}}
%\put(110,36){\line(1,0){08}}\put(120,36){\line(1,0){08}}
%-----------------------------------------

\put(30,15){\makebox(0,8)[bl]{SUPPORT}}
\put(30,11){\makebox(0,8)[bl]{LAYER}}

%==========================

\put(67.5,19){\vector(0,1){06}}

\put(80.8,19){\vector(3,1){16.6}}

\put(106,18){\vector(-3,1){22}}

%========================= Hierarchies, DAs, compatibility

\put(50.5,15){\makebox(0,8)[bl]{Hierarchies of medical }}
\put(55,12){\makebox(0,8)[bl]{treatment plans }}
\put(55,09){\makebox(0,8)[bl]{(including DAs,}}
\put(56,06){\makebox(0,8)[bl]{compatibility)}}

\put(50,05){\line(1,0){35}}\put(50,19){\line(1,0){35}}
\put(50,05){\line(0,1){14}}\put(85,05){\line(0,1){14}}

%============================ Logical rules

\put(107,18){\vector(0,1){07}}

\put(94,06){\line(1,0){26}} \put(94,18){\line(1,0){26}}

\put(91,12){\line(1,2){03}}\put(91,12){\line(1,-2){03}}
\put(123,12){\line(-1,2){03}}\put(123,12){\line(-1,-2){03}}

\put(94,13){\makebox(0,8)[bl]{Logical rules for}}
\put(92.5,10){\makebox(0,8)[bl]{``analysis/decision''}}
\put(99,07){\makebox(0,8)[bl]{points}}

%+++++++++++++++++++++++++++++++++++++End point

%\put(8,24){\oval(16,10)}

\put(130,47){\makebox(0,8)[bl]{End point}}
\put(130.2,43){\makebox(0,8)[bl]{(resultant}}
\put(131.4,39.7){\makebox(0,8)[bl]{medical}}
\put(130.5,35){\makebox(0,8)[bl]{situation)}}

\put(129,34){\line(1,0){18}}\put(129,51){\line(1,0){18}}
\put(129,34){\line(0,1){17}}\put(147,34){\line(0,1){17}}

%=================

\put(124,29){\vector(2,3){5}}

\put(124,42){\vector(1,0){5}}

\put(124,57){\vector(2,-3){5}}

\end{picture}
\end{center}

% Here, multi-stage design for medical treatment is examined.
%
 Here, a simplified illustrative example for designing
 a two-phase  trajectory
 for medical treatment of children asthma is briefly described
 (for standard medical treatment, for non-standard medical treatment).
 The scheme is based on materials from
 \cite{lev12hier,lev15,levsok04}.
 A two-phase scheme of medical treatment planning and
 implementation
% (including support layer)
 is depicted in Fig. 27
 (basic flow-chart that can involve online modes, support layer).

%%%%%%%%%%%%%%%%%%%%%%%%%%%%%%%%%%%%%%%%%%%%%%%%%%%%%%%%%%
%\section{Example of Tree-like Structure for Medical Treatment}
%\subsection{Tree-like Trajectory for Medical Treatment}

\begin{center}
\begin{picture}(117,74.5)
\put(19,00){\makebox(0,0)[bl]{Fig. 27. Two-phase scheme
 of medical treatment}}

%+++++++++++++++++++++++++++++++++++++++++++++ Initial point

%\put(8,24){\oval(16,10)}

\put(02.7,57){\makebox(0,8)[bl]{Initial}}
\put(03.7,53){\makebox(0,8)[bl]{point}}
\put(01,49){\makebox(0,8)[bl]{(medical)}}
\put(0.7,45){\makebox(0,8)[bl]{situation)}}

\put(00,44){\line(1,0){16}}\put(00,61){\line(1,0){16}}
\put(00,44){\line(0,1){17}}\put(16,44){\line(0,1){17}}

\put(16,52.5){\vector(1,0){4}}

%========================= Hierarchies, DAs, compatibility

\put(24,62){\makebox(0,8)[bl]{BASIC PHASE: }}

\put(21,57){\makebox(0,8)[bl]{1.Design/selection \&}}
\put(24,54){\makebox(0,8)[bl]{implementation of}}
\put(24,51.5){\makebox(0,8)[bl]{medical treatment}}
\put(21,48){\makebox(0,8)[bl]{2.Analysis\&selection}}
\put(24,45){\makebox(0,8)[bl]{of next direction}}

\put(20,44){\line(1,0){35}}\put(20,61){\line(1,0){35}}
\put(20,44){\line(0,1){17}}\put(55,44){\line(0,1){17}}

\put(55,57.5){\vector(1,0){5}} \put(55,52.5){\vector(1,0){5}}
\put(55,47.5){\vector(1,0){5}}

%========================= Hierarchies, DAs, compatibility

\put(58.5,70){\makebox(0,8)[bl]{ADDITIONAL PHASE:}}

\put(61,65){\makebox(0,8)[bl]{1.Design/selection \&}}
\put(64,62){\makebox(0,8)[bl]{implementation of}}
\put(64,59.5){\makebox(0,8)[bl]{medical treatment}}
\put(61,56){\makebox(0,8)[bl]{2.Analysis\&selection}}
\put(64,53.5){\makebox(0,8)[bl]{of next direction}}
\put(61,50){\makebox(0,8)[bl]{3.Aditional}}
\put(64,46.5){\makebox(0,8)[bl]{design/selection \&}}
\put(64,43){\makebox(0,8)[bl]{implementation of}}
\put(64,40.5){\makebox(0,8)[bl]{medical treatment}}

\put(60,39){\line(1,0){34}}\put(60,69){\line(1,0){34}}
\put(60,39){\line(0,1){30}}\put(94,39){\line(0,1){30}}

%%%%%%%%%%%%%%%%%%%%%%%%%%%%%%%%%%%%

\put(08,36){\line(1,0){08}}\put(18,36){\line(1,0){08}}
\put(28,36){\line(1,0){08}}\put(39,36){\line(1,0){07}}
\put(48,36){\line(1,0){08}}\put(58,36){\line(1,0){08}}
\put(68,36){\line(1,0){08}}\put(78,36){\line(1,0){08}}
\put(88,36){\line(1,0){08}}\put(98,36){\line(1,0){08}}

%\put(110,36){\line(1,0){08}}\put(120,36){\line(1,0){08}}

\put(00,31){\makebox(0,8)[bl]{SUPPORT}}
\put(00,27){\makebox(0,8)[bl]{LAYER}}

%========================= Hierarchies, DAs, compatibility

\put(37.5,34){\vector(0,1){10}}

\put(50.5,34){\vector(3,1){14.7}}

\put(21,30){\makebox(0,8)[bl]{Preliminary designed}}
\put(22.7,27){\makebox(0,8)[bl]{composite DAs for}}
\put(21.6,24){\makebox(0,8)[bl]{treatment plan parts}}

\put(20,23){\line(1,0){35}}\put(20,34){\line(1,0){35}}
\put(20,23){\line(0,1){11}}\put(55,23){\line(0,1){11}}

%==========================

\put(37.5,19){\vector(0,1){4}}

%========================= Hierarchies, DAs, compatibility

\put(20.5,15){\makebox(0,8)[bl]{Hierarchies of medical }}
\put(25,12){\makebox(0,8)[bl]{treatment plans }}
\put(25,09){\makebox(0,8)[bl]{(including DAs,}}
\put(26,06){\makebox(0,8)[bl]{compatibility)}}

\put(20,05){\line(1,0){35}}\put(20,19){\line(1,0){35}}
\put(20,05){\line(0,1){14}}\put(55,05){\line(0,1){14}}

%\put(37.5,24){\vector(1,0){4}}

%============================ Logical rules

\put(73,34){\vector(-3,1){29}}

\put(77,34){\vector(0,1){05}}

%\put(77,28){\oval(32,12)}

\put(64,22){\line(1,0){26}}\put(64,34){\line(1,0){26}}
%\put(61,06){\line(0,1){12}}\put(93,06){\line(0,1){12}}

\put(61,28){\line(1,2){03}}\put(61,28){\line(1,-2){03}}
\put(93,28){\line(-1,2){03}}\put(93,28){\line(-1,-2){03}}

\put(64,29){\makebox(0,8)[bl]{Logical rules for}}
\put(62.5,26){\makebox(0,8)[bl]{``analysis/decision''}}
\put(69,23){\makebox(0,8)[bl]{points}}

%\put(21,06){\makebox(0,8)[bl]{compatibility)}}

%\put(60,08){\line(1,0){30}}\put(60,19){\line(1,0){30}}
%\put(60,08){\line(0,1){11}}\put(90,08){\line(0,1){11}}
%\put(15.5,24){\vector(1,0){4}}

%+++++++++++++++++++++++++++++++++++++End point

%\put(8,24){\oval(16,10)}

\put(100,57){\makebox(0,8)[bl]{End point}}
\put(100.2,53){\makebox(0,8)[bl]{(resultant}}
\put(101.4,49.7){\makebox(0,8)[bl]{medical}}
\put(100.5,45){\makebox(0,8)[bl]{situation)}}

\put(99,44){\line(1,0){18}}\put(99,61){\line(1,0){18}}
\put(99,44){\line(0,1){17}}\put(117,44){\line(0,1){17}}

\put(94,57.5){\vector(3,-1){5}} \put(94,52.5){\vector(1,0){5}}
\put(94,47.5){\vector(3,1){5}}

\end{picture}
\end{center}

% Here, multi-stage design for medical treatment is examined.
%
% Here, a simplified illustrative example for designing
% a two-phase  trajectory
% for medical treatment of children asthma is briefly described
% (for standard medical treatment, for non-standard medical teratment).
%
% The scheme is based on materials from
% \cite{lev12hier,lev15,levsok04}.
%
 The considered general trajectory scheme for medical treatment
 is depicted in Fig. 28 with two kinds of node elements (points):
 ~(i) ``design/implementation'' elements
 (based on hierarchies of design alternatives and their
 estimates),
 ~(ii) ``analysis/decision''  elements (based on logical rules).

 Table 11 contains descriptions of design/implementation points.
 Table 12 contains descriptions (logical rules) of
 ``analysis/decision'' points.

\begin{center}
\begin{picture}(106,47)
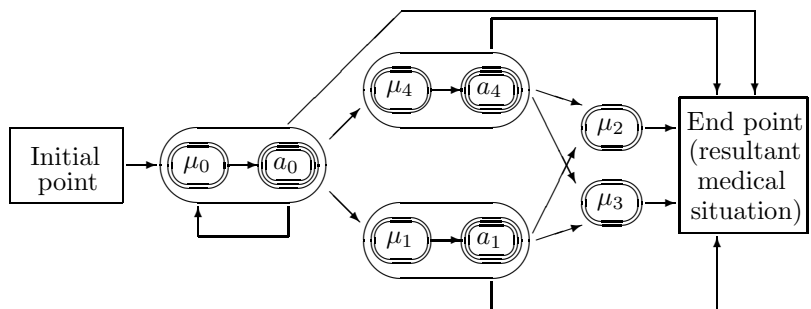
\put(02,00){\makebox(0,0)[bl]{Fig. 28. Medical treatment scheme
 (basis of medical trajectories)
% \cite{lev12hier,lev15}
 }}

%+++++++++++++++++++++++++++++++++++++++++++++ Initial point

%\put(8,24){\oval(16,10)}

\put(02.7,24){\makebox(0,8)[bl]{Initial}}
\put(03.7,20){\makebox(0,8)[bl]{point}}
%\put(01,20){\makebox(0,8)[bl]{(situation)}}

\put(00,19){\line(1,0){15}}\put(00,29){\line(1,0){15}}
\put(00,19){\line(0,1){10}}\put(15,19){\line(0,1){10}}

\put(15.5,24){\vector(1,0){4}}

%+++++++++++++++++++++++++++++++++++++++++++++ tau0-a0

\put(31,24){\oval(22,10)}

\put(25,24){\oval(8,6)} \put(25,24){\oval(7,5)}
\put(23,23){\makebox(0,8)[bl]{\(\mu_{0}\)}}

\put(29,24){\vector(1,0){4}}

%--A0

\put(37,24){\oval(8,6)} \put(37,24){\oval(7,5)}
\put(37,24){\oval(6,4)}

\put(35,23){\makebox(0,8)[bl]{\(a_{0}\)}}

\put(42.5,27.5){\vector(1,1){4}} \put(42.5,20.5){\vector(1,-1){4}}

\put(37,18.5){\line(0,-1){4}} \put(37,14.5){\line(-1,0){12}}
\put(25,14.5){\vector(0,1){4}}

\put(37,29.5){\line(1,1){15}} \put(52,44.5){\line(1,0){47}}
\put(99,44.5){\vector(0,-1){11}}

%--------------+++++++++++++++++++++++++++++ tau1-a1

\put(58,14){\oval(22,10)}

\put(52,14){\oval(8,6)} \put(52,14){\oval(7,5)}
\put(50,13){\makebox(0,8)[bl]{\(\mu_{1}\)}}

\put(56,14){\vector(1,0){4}}

%--A1

\put(64,14){\oval(8,6)} \put(64,14){\oval(7,5)}
\put(64,14){\oval(6,4)}

\put(62,13){\makebox(0,8)[bl]{\(a_{1}\)}}

\put(69.5,15){\vector(1,2){5.7}} \put(70,14){\vector(3,1){6}}

%\put(69,17.5){\vector(3,1){6}} \put(69,10.5){\vector(3,-1){6}}

\put(64,8.5){\line(0,-1){4}} \put(64,4.5){\line(1,0){30}}
\put(94,4.5){\vector(0,1){10}}

%---------------------------------------------

%\put(80,09){\oval(8,6)} \put(80,09){\oval(7,5)}
%\put(78,08){\makebox(0,8)[bl]{\(\mu_{2}\)}}

%\put(84.5,09){\vector(1,2){4}}

%--

\put(80,19){\oval(8,6)} \put(80,19){\oval(7,5)}
\put(78,18){\makebox(0,8)[bl]{\(\mu_{3}\)}}

\put(84.5,19){\vector(1,0){4}}

%%%%%%%%%%%%%%%%%%%%%%%%%%%%%%%%%%%%%%%%%%%%%%

%--+++++++++++++++++++++++++++++ tau4-a4

\put(58,34){\oval(22,10)}

\put(52,34){\oval(8,6)} \put(52,34){\oval(7,5)}
\put(50,33){\makebox(0,8)[bl]{\(\mu_{4}\)}}

\put(56,34){\vector(1,0){4}}

%--A4

\put(64,34){\oval(8,6)} \put(64,34){\oval(7,5)}
\put(64,34){\oval(6,4)}

\put(62,33){\makebox(0,8)[bl]{\(a_{4}\)}}

\put(69.5,33){\vector(1,-2){5.7}} \put(70,34){\vector(3,-1){6}}

%\put(47,39.5){\line(1,3){2}} \put(49,45.5){\vector(2,1){6}}

\put(64,39.5){\line(0,1){4}} \put(64,43.5){\line(1,0){30}}
\put(94,43.5){\vector(0,-1){10}}

%------------------------------------------

\put(80,29){\oval(8,6)} \put(80,29){\oval(7,5)}
\put(78,28){\makebox(0,8)[bl]{\(\mu_{2}\)}}

\put(84.5,29){\vector(1,0){4}}

%--

%\put(80,39){\oval(8,6)} \put(80,39){\oval(7,5)}
%\put(78,38){\makebox(0,8)[bl]{\(\mu_{6}\)}}

%\put(84.5,39){\vector(1,-2){4}}

%--

%\put(60,49){\oval(8,6)} \put(60,49){\oval(7,5)}
%\put(58,48){\makebox(0,8)[bl]{\(\tau_{7}\)}}

%%%%%%%%%%%%%%%%%%%%%%%%%%%%%%%%%%%%%%%%%%%%
%+++++++++++++++++++++++++++++++++++++++++++++ End point

%\put(8,24){\oval(16,10)}

\put(90,28){\makebox(0,8)[bl]{End point}}
\put(90.2,24){\makebox(0,8)[bl]{(resultant}}
\put(91.4,20.8){\makebox(0,8)[bl]{medical}}
\put(90.5,16){\makebox(0,8)[bl]{situation)}}

\put(89,15){\line(1,0){17}}\put(89,33){\line(1,0){17}}
\put(89,15){\line(0,1){18}}\put(106,15){\line(0,1){18}}

%\put(16,24){\vector(1,0){4}}

\end{picture}
\end{center}
%

%\newpage
\begin{center}
{\bf Table 11.} Design/implementation points  (design/selection \& implementation of plan parts)\\
\begin{tabular}{| c|c | l | }
\hline
 No.&Design/imple-&Description\\
    &mentation &            \\
     &point &             \\

%-------------------------------
\hline
 1.&\(\mu_{0}\)&Design\&implementation of initial (basic) treatment plan \\
 2.&\(\mu_{1}\)&Design\&implementation of additional environmental  treatment\\
 3.&\(\mu_{2}\)&Design\&implementation of additional treatment by relaxation \\
 4.&\(\mu_{3}\)&Design\&implementation of additional physical therapy\\
 5.&\(\mu_{4}\)&Design\&implementation of additional joint physical therapy\\
  & &and drug based treatment  \\

% 6.&\(\mu_{5}\) (as \(\mu_{1}\))&Design\&implementation of additional environmental treatment&Fig. 28\\
% 7.&\(\mu_{6}\) (as \(\mu_{4}\))&Design\&implementation of joint physical therapy &Fig. 31\\
% &&and drug based treatment &\\

\hline
\end{tabular}
\end{center}

\newpage
\begin{center}
{\bf Table 12.} ``Analysis/decision'' points \\
\begin{tabular}{|c| c | l | }
\hline
 No.&``Analysis/''&Description (logical rules)\\
    &``decision'' &           \\
    &point        &           \\

%-------------------------------
\hline
 1.&\(a_{0}\)&(i) repetition of basic treatment is required, ``Go To'' \(\mu_{0}\) \\
   &         &(ii) excellent medical results, ``Go To''  ``End point'' \\
   &         &(iii) good medical results, ``Go To'' \(\mu_{1}\) (additional environmental treatment)\\
   &         &(iv) about sufficient medical results, ``Go To'' \(\mu_{4}\) \\

 2.&\(a_{1}\)&(i) good medical results, ``Go To'' \(\mu_{2}\) (relaxation) \\
   &         &(ii) excellent medical results, ``Go To''  ``End point'' \\
   &         &(iii) about sufficient medical results, ``Go To'' \(\mu_{3}\) (additional physical therapy) \\

 3.&\(a_{4}\)&(i) good medical results, ``Go To'' \(\mu_{2}\) (additional environmental treatment)\\
   &         &(ii) excellent medical results, ``Go To''  ``End point'' \\
   &         &(iii) about sufficient medical results, ``Go To'' \(\mu_{3}\)  \\

\hline
\end{tabular}
\end{center}

  Each ``design/implementation'' node element (point) of the treatment trajectory
 (i.e., node \(\mu_{i}, i=\overline{0,4}\))
   is based on a simplified hierarchical
 structure of medical treatment
%  for children asthma
 that has been suggested in \cite{levsok04}.
 Combinatorial  synthesis (HMMD) is used for composition
 of composite DAs (brief description is presented in previous section).
 The basic hierarchical structure of medical treatment
 is presented in Fig. 29
 (priorities of DAs are shown in  parentheses, expert judgment):

\begin{center}
\begin{picture}(129,84)
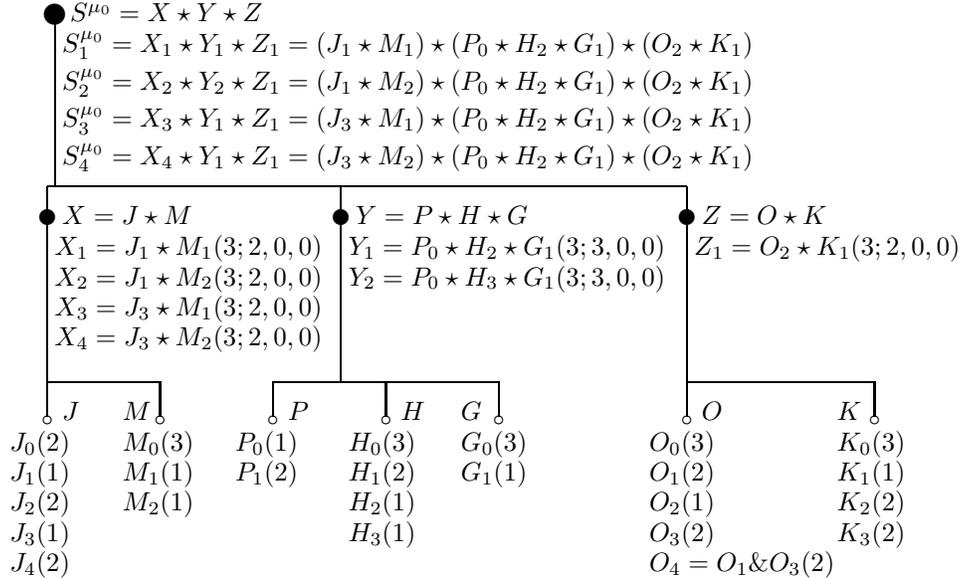
\put(21,00){\makebox(0,0)[bl] {Fig. 29. Hierarchical model of
 medical treatment plan
%  \cite{lev06,levsok04}
   }}
%----------------------------------------------------------------------
\put(06,57){\line(0,1){23}} \put(06,80){\circle*{2.8}}

\put(08,79){\makebox(0,0)[bl]{\(S^{\mu_{0}}=X\star Y\star Z\)}}

\put(07,74){\makebox(0,0)[bl]{\(S_{1}^{\mu_{0}}=X_{1}\star
 Y_{1}\star Z_{1}=
 (J_{1}\star M_{1})\star (P_{0}\star H_{2}\star G_{1})
  \star (O_{2}\star K_{1})\)  }}

\put(07,69){\makebox(0,0)[bl]{\(S_{2}^{\mu_{0}}=X_{2}\star
 Y_{2}\star Z_{1}=
  (J_{1}\star M_{2})\star (P_{0}\star H_{2}\star G_{1})
  \star (O_{2}\star K_{1})\)  }}

\put(07,64){\makebox(0,0)[bl]{\(S_{3}^{\mu_{0}}=X_{3}\star
 Y_{1}\star Z_{1}=
  (J_{3}\star M_{1})\star (P_{0}\star H_{2}\star G_{1})
  \star (O_{2}\star K_{1})\)  }}

\put(07,59){\makebox(0,0)[bl]{\(S_{4}^{\mu_{0}}=X_{4}\star
 Y_{1}\star Z_{1}=
  (J_{3}\star M_{2})\star (P_{0}\star H_{2}\star G_{1})
  \star (O_{2}\star K_{1})\)  }}

%--------------------------------------------------------------------
\put(05,57){\line(1,0){85}}

\put(05,31){\line(0,1){26}} \put(05,53){\circle*{2}}

\put(07,52){\makebox(0,0)[bl]{\(X=J\star M\)}}

\put(06,47){\makebox(0,0)[bl]{\(X_{1}=J_{1}\star M_{1}
  (3;2,0,0)\)}}
\put(06,43){\makebox(0,0)[bl]{\(X_{2}=J_{1}\star M_{2}
 (3;2,0,0)\)}}

\put(06,39){\makebox(0,0)[bl]{\(X_{3}=J_{3}\star
 M_{1}(3;2,0,0)\)}}

\put(06,35){\makebox(0,0)[bl]{\(X_{4}=J_{3}\star
 M_{2}(3;2,0,0)\)}}

%------------------------------

\put(44,31){\line(0,1){26}} \put(44,53){\circle*{2}}

\put(46,52){\makebox(0,0)[bl]{\(Y=P\star H\star G\)}}

\put(45,47){\makebox(0,0)[bl]{\(Y_{1}=P_{0}\star H_{2}\star
 G_{1}(3;3,0,0)\)}}

\put(45,43){\makebox(0,0)[bl]{\(Y_{2}=P_{0}\star H_{3}\star
 G_{1}(3;3,0,0)\)}}

%----

\put(90,31){\line(0,1){26}} \put(90,53){\circle*{2}}

\put(92,52){\makebox(0,0)[bl]{\(Z=O\star K\)}}

\put(91,47){\makebox(0,0)[bl]{\(Z_{1}=O_{2} \star K_{1}
 (3;2,0,0)\)}}

%%%%%%%%%%%%%%%%%%%%%%%%%%%%%%%%%%%%%%%%%%%%%%%%%%%%%%%%%%

\put(05,31){\line(1,0){15}} \put(05,26.5){\line(0,1){4.5}}
\put(05,26){\circle{1}}

\put(7,26){\makebox(0,0)[bl]{\(J\)}}

\put(0,21){\makebox(0,0)[bl]{\(J_{0}(2)\)}}
\put(0,17){\makebox(0,0)[bl]{\(J_{1}(1)\)}}
\put(0,13){\makebox(0,0)[bl]{\(J_{2}(2)\)}}
\put(0,09){\makebox(0,0)[bl]{\(J_{3}(1)\)}}
\put(0,05){\makebox(0,0)[bl]{\(J_{4}(2)\)}}

\put(20,26.5){\line(0,1){4.5}} \put(20,26){\circle{1}}

\put(15,26){\makebox(0,0)[bl]{\(M\)}}

\put(15,21){\makebox(0,0)[bl]{\(M_{0}(3)\)}}
\put(15,17){\makebox(0,0)[bl]{\(M_{1}(1)\)}}
\put(15,13){\makebox(0,0)[bl]{\(M_{2}(1)\)}}

%-----------------------------------------------------

\put(35,31){\line(1,0){30}} \put(35,26.5){\line(0,1){4.5}}
\put(35,26){\circle{1}}

\put(37,26){\makebox(0,0)[bl]{\(P\)}}

\put(30,21){\makebox(0,0)[bl]{\(P_{0}(1)\)}}
\put(30,17){\makebox(0,0)[bl]{\(P_{1}(2)\)}}

\put(50,26.5){\line(0,1){4.5}} \put(50,26){\circle{1}}

\put(52,26){\makebox(0,0)[bl]{\(H\)}}

\put(45,21){\makebox(0,0)[bl]{\(H_{0}(3)\)}}
\put(45,17){\makebox(0,0)[bl]{\(H_{1}(2)\)}}
\put(45,13){\makebox(0,0)[bl]{\(H_{2}(1)\)}}
\put(45,09){\makebox(0,0)[bl]{\(H_{3}(1)\)}}

\put(65,26.5){\line(0,1){4.5}} \put(65,26){\circle{1}}

\put(60,26){\makebox(0,0)[bl]{\(G\)}}

\put(60,21){\makebox(0,0)[bl]{\(G_{0}(3)\)}}
\put(60,17){\makebox(0,0)[bl]{\(G_{1}(1)\)}}
%--------------------------------------------------------------
\put(90,31){\line(1,0){25}}

\put(90,26.4){\line(0,1){4.5}} \put(90,26){\circle{1}}

\put(92,26){\makebox(0,0)[bl]{\(O\)}}

\put(85,21){\makebox(0,0)[bl]{\(O_{0}(3)\)}}
\put(85,17){\makebox(0,0)[bl]{\(O_{1}(2)\)}}
\put(85,13){\makebox(0,0)[bl]{\(O_{2}(1)\)}}
\put(85,09){\makebox(0,0)[bl]{\(O_{3}(2)\)}}
\put(85,05){\makebox(0,0)[bl]{\(O_{4}=O_{1}\&O_{3}(2)\)}}

\put(115,26.5){\line(0,1){4.5}} \put(115,26){\circle{1}}

\put(110,26){\makebox(0,0)[bl]{\(K\)}}

\put(110,21){\makebox(0,0)[bl]{\(K_{0}(3)\)}}
\put(110,17){\makebox(0,0)[bl]{\(K_{1}(1)\)}}
\put(110,13){\makebox(0,0)[bl]{\(K_{2}(2)\)}}
\put(110,09){\makebox(0,0)[bl]{\(K_{3}(2)\)}}

\end{picture}
\end{center}

%~~

 {\bf 0.} Plan of medical treatment ~\(S = X \star Y \star  Z \).

 {\bf 1.} Basic treatment  ~\(X = J \star M\):

 {\it 1.1.} Physical therapy  \(J \):~
  none \(J_{0}(2)\),
  massage \(J_{1}(1)\),
%  inhalation \(J_{2}(2)\),
%  sauna \(J_{3}(3)\),
%  reflexological therapy \(J_{4}(3)\),
  laser-therapy \(J_{2}(2)\),
  massage for special centers/points \(J_{3}(1)\),
%  reflexological therapy for special centers \(J_{7}(4)\),
% and
  halo-cameras or salt mines \(J_{4}(2)\).

 {\it 1.2.} Drug based treatment  \(M\):~
  none \(M_{0}(3)\),
  vitamins \(M_{1}(1)\),
  sodium chromoglycate
%  (one month and two times in a year)
  \(M_{2}(1)\).
%  sodium chromoglycate (two months) \(M_{3}\)(3),
% and
%  sodium chromoglycate (three months) \(M_{4}(3)\).

 {\bf 2.} Improvement of psychological and ecological environment
 ~\(Y = P \star H \star G \):

 {\it 2.1.} Psychological climate \(P\):~
  none \(P_{0}(1)\),
% and
  consulting of a psychologist \(P_{1}(2)\).

 {\it 2.2.} Home ecological environment \(H\):~
  none \(H_{0}(3)\),
%  water cleaning \(H_{1}(1)\),
  to clean a book dust \(H_{1}(2)\),
%  to take away cotton wool things (blanket, pillow, mattress) \(H_{3}(1)\),
%  to take away carpets \(H_{4}(1)\),
  to exclude contacts with home animals \(H_{2}(1)\),
%  to destroy cockroach environment \(H_{6}(2)\),
  to take away flowers \(H_{3}(1)\).
%  and
%  aggregated alternative \(H_{8}=H_{1}\&H_{4}\&H_{5}\&H_{7}(1)\).

 {\it 2.2.} General ecological environment \(G\):~
 none \(G_{0}(3)\),
%  and
 improving the area of the residence \(G_{1}(1)\).

 {\bf 3.} Improvement of mode, rest and relaxation ~\(Z = O \star K \):

 {\it 3.1.} Mode \(O \):~
  none \(O_{0}(3)\),
%  relaxation at the noon \(O_{1}(1)\),
  special physical actions (drainage, expectoration) \(O_{1}(2)\),
  sport (running, skiing, swimming) \(O_{2}(1)\),
  comfort shower-bath \(O_{3}(2)\).
%  cold shower-bath \(O_{5}(2)\),
%  the exclude electronic games \(O_{6}(2)\),
  aggregated alternative \(O_{4}=O_{1}\&O_{3}(2)\).
% and
%  aggregated alternative \(O_{8}=O_{3}\&O_{5}(2)\).

 {\it 3.2.} Relaxation/rest \(K \):~
 none \(K_{0}(3)\),
 rest at forest-like environment \(K_{1}(1)\),
 rest near see \(K_{2}(2)\),
% a rest at mountains \(K_{3}(4)\),
% special camps \(K_{4}(3)\),
% and
 special treatment in salt mines \(K_{3}(2)\).

~~

 In Fig. 29, the hierarchy (i.e., morphological structure)
 corresponds to design/implementaiton point
 \(\mu_{0}\) (Fig. 28).
 Estimates of compatibility for DAs are presented in
 Table 13, Table 14, and Table 15
  \cite{lev06,levsok04}
 (as simplified version, the estimates are the same ones for all points
 \(\{\mu_{0},\mu_{1},\mu_{2},\mu_{3},\mu_{4},\mu_{5},\mu_{6}\}\)).
 Estimates of compatibility for DAs at the higher hierarchical level
 are presented in Table 16 (for point \(\mu_{0}\)).

%\newpage
\begin{center}
 {\bf Table 13.} Compatibility  \\
\begin{tabular}{| l | c c c |}
\hline
 &  \(M_{0}\)&\(M_{1}\)&\(M_{2}\)\\
%-------------------------------
\hline
 \(J_{0}\) &\(0\)&\(3\)&\(3\)\\
 \(J_{1}\) &\(2\)&\(3\)&\(3\)\\
 \(J_{2}\) &\(2\)&\(3\)&\(3\)\\
 \(J_{3}\) &\(2\)&\(3\)&\(3\)\\
 \(J_{4}\) &\(1\)&\(3\)&\(2\)\\

\hline
\end{tabular}
\end{center}

%\newpage
\begin{center}
 {\bf Table 14.} Compatibility  \\
\begin{tabular}{| l | c c c c|}
\hline
 &  \(K_{0}\)&\(K_{1}\)&\(K_{2}\)&\(K_{3}\)\\
%-------------------------------
\hline
 \(O_{0}\) &\(0\)&\(3\)&\(3\)&\(3\)\\
 \(O_{1}\) &\(0\)&\(3\)&\(3\)&\(2\)\\
 \(O_{2}\) &\(0\)&\(3\)&\(3\)&\(2\)\\
 \(O_{3}\) &\(0\)&\(3\)&\(3\)&\(1\)\\
 \(O_{4}\) &\(0\)&\(3\)&\(3\)&\(3\)\\

\hline
\end{tabular}
\end{center}

\begin{center}
 {\bf Table 15.} Compatibility  \\
\begin{tabular}{| l | c c c c c c|}
\hline
 &  \(G_{0}\)&\(G_{1}\)&\(H_{0}\)&\(H_{1}\)&\(H_{2}\)&\(H_{3}\)\\
%-------------------------------
\hline
 \(P_{0}\) &\(1\)&\(3\)&\(0\)&\(3\)&\(3\)&\(3\)\\
 \(P_{1}\) &\(3\)&\(2\)&\(2\)&\(3\)&\(3\)&\(3\)\\
 \(G_{0}\) &\( \)&\( \)&\(0\)&\(0\)&\(0\)&\(0\)\\
 \(G_{1}\) &\( \)&\( \)&\(0\)&\(3\)&\(3\)&\(3\)\\

\hline
\end{tabular}
\end{center}

%\newpage
\begin{center}
 {\bf Table 16.} Compatibility  \\
\begin{tabular}{| l | c c c |}
\hline
 &  \(Y_{1}\)&\(Y_{2}\)&\(Z_{1}\)\\
%-------------------------------
\hline
 \(X_{1}\) &\(3\)&\(3\)&\(3\)\\
 \(X_{2}\) &\(3\)&\(3\)&\(3\)\\
 \(X_{3}\) &\(3\)&\(3\)&\(3\)\\
 \(X_{4}\) &\(3\)&\(3\)&\(3\)\\
 \(Y_{1}\) &     &     &\(3\)\\
 \(Y_{2}\) &     &     &\(1\)\\

\hline
\end{tabular}
\end{center}

 For basic (initial) point \(\mu_{0}\) (Fig. 29),
 the resultant composite Pareto-efficient DAs are:

 (1) local Pareto-efficient solutions for subsystem \(X\):~
 \(X_{1} = J_{1}\star M_{1}\),  \(N(X_{1})= (3;2,0,0)\);
 \(X_{2} = J_{1}\star M_{2}\),  \(N(X_{2})= (3;2,0,0)\);
 \(X_{3} = J_{3}\star M_{1}\),  \(N(X_{3})= (3;2,0,0)\);
 \(X_{4} = J_{3}\star M_{2}\),  \(N(X_{4})= (3;2,0,0)\);

 (2) local Pareto-efficient solutions for subsystem \(Y\):~
 \(Y_{1} = P_{0}\star H_{2} \star G_{1}\),  \(N(Y_{1})= (3;3,0,0)\);
 \(Y_{2} = P_{0}\star H_{3} \star G_{1}\),  \(N(Y_{2})= (3;3,0,0)\);

  (3) local Pareto-efficient solutions for subsystem \(Z\):~
 \(Z_{1} = O_{2}\star K_{1}\),  \(N(Z_{1})= (3;2,0,0)\).

 (4) final composite Pareto-efficient DAs for system  \(S\)
 (\(N(S^{\mu_{0}}_{\iota})= (3;3,0,0)\), \(\iota =\overline{1,4}\)):~
 (a)
 \(S^{\mu_{0}}_{1} = X_{1} \star Y_{1} \star Z_{1}  \),
% \(N(S^{\mu_{0}}_{1})= (2;4,0,0)\);
% and
%
 (b)
 \(S^{\mu_{0}}_{2} = X_{2} \star Y_{1} \star Z_{2} \),
% \(N(S^{\mu_{0}}_{2})= (3;3,1,0)\).
%
 (c)
 \(S^{\mu_{0}}_{3} = X_{3} \star Y_{1} \star Z_{1} \),
% \(N(S^{\mu_{0}}_{1})= (2;4,0,0)\);
% and
%
 (d)
 \(S^{\mu_{0}}_{4} = X_{4} \star Y_{1} \star Z_{1} \).
% \(N(S^{\mu_{0}}_{2})= (3;3,1,0)\).

 Further, composite DAs for
 points \(\mu_{1}\), \(\mu_{2}\),
 \(\mu_{3}\), \(\mu_{4}\) are designed.
 Here, compatibility estimates correspond to
 Table 13, Table 14, Table 15;
 priorities of DAs are based on new expert judgment
 (Fig. 30, Fig. 31, Fig. 32, Fig. 33; in parentheses).

%%%%%%%%%%%%%%%%%%%%%%%%%%%%%%%%%%%%%%%%%%%%%%%%%%%%%%%%%%%%%%%%%%%%

 For point \(\mu_{1}\) (Fig. 30),
 the resultant composite Pareto-efficient DA are:~
 (a) \(S^{\mu_{1}}_{1}=P_{0}\star H_{2}\star G_{1}\),
 \(N(S^{\mu_{1}}_{1})= (3;3,0,0)\);
 (b) \(S^{\mu_{1}}_{2}=P_{0}\star H_{3}\star G_{1}\),
 \(N(S^{\mu_{1}}_{1})= (3;3,0,0)\).

%=========================================================

 For point \(\mu_{2}\) (Fig. 31, and for \(\mu_{5}\)),
 the resultant composite Pareto-efficient DAs are:~
 (a) \(S^{\mu_{2}}_{1}=O_{4}\star K_{1}\),
 \(N(S^{\mu_{2}}_{1})= (3;2,0,0)\);
 (b) \(S^{\mu_{2}}_{2}=O_{4}\star K_{3}\),
   \(N(S^{\mu_{2}}_{2})= (3;2,0,0)\).

%%%%%%%%%%%%%%%%%%%%%%%%%%%%%%%%%%%%%%%%%%%%%%%%%%%%%%%%%%%%

 For point \(\mu_{3}\) (Fig. 32),
 the resultant composite Pareto-efficient DAs are:~
 (a) \(S^{\mu_{3}}_{1}=J_{3}\),
% \(N(S^{\mu_{3}}_{1})= (3;2,0)\);
%
 (b) \(S^{\mu_{3}}_{2}=J_{4}\).
%   \(N(S^{\mu_{3}}_{2})= (3;2,0)\).

%%%%%%%%%%%%%%%%%%%%%%%%%%%%%%%%%%%%%%%%%%%%%%%%%%

 For point \(\mu_{4}\) (Fig. 33, and for \(\mu_{6}\)),
 the resultant composite Pareto-efficient DAs are:~
 (a) \(S^{\mu_{4}}_{1}=J_{3}\star M_{1}\),
 \(N(S^{\mu_{2}}_{1})= (3;2,0)\);
 (b) \(S^{\mu_{4}}_{2}=J_{4}\star M_{1}\),
   \(N(S^{\mu_{2}}_{2})= (3;2,0)\).

\begin{center}
%\begin{picture}(50,61)
\begin{picture}(60,43)
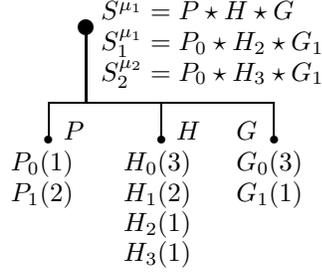
\put(00,00){\makebox(0,0)[bl] {Fig. 30. Treatment
 for point \(\mu_{1}\) }}

%------------------------------

\put(10,27){\line(0,1){10}} \put(10,37){\circle*{2}}

\put(12,38){\makebox(0,0)[bl]{\(S^{\mu_{1}}=P\star H\star G\)}}

\put(12,33){\makebox(0,0)[bl]{\(S^{\mu_{1}}_{1}=P_{0}\star
 H_{2}\star G_{1}\)}}

\put(12,29){\makebox(0,0)[bl]{\(S^{\mu_{2}}_{2}=P_{0}\star
  H_{3}\star G_{1}\)}}

%-----------------------------------------------------

\put(05,27){\line(1,0){30}}

\put(05,22){\line(0,1){5}} \put(05,22){\circle*{1}}

\put(07,22){\makebox(0,0)[bl]{\(P\)}}

\put(00,17){\makebox(0,0)[bl]{\(P_{0}(1)\)}}
\put(00,13){\makebox(0,0)[bl]{\(P_{1}(2)\)}}

%--

\put(20,22){\line(0,1){5}} \put(20,22){\circle*{1}}

\put(22,22){\makebox(0,0)[bl]{\(H\)}}

\put(15,17){\makebox(0,0)[bl]{\(H_{0}(3)\)}}
\put(15,13){\makebox(0,0)[bl]{\(H_{1}(2)\)}}
\put(15,09){\makebox(0,0)[bl]{\(H_{2}(1)\)}}
\put(15,05){\makebox(0,0)[bl]{\(H_{3}(1)\)}}

%--

\put(35,22){\line(0,1){5}} \put(35,22){\circle*{1}}

\put(30,22){\makebox(0,0)[bl]{\(G\)}}

\put(30,17){\makebox(0,0)[bl]{\(G_{0}(3)\)}}
\put(30,13){\makebox(0,0)[bl]{\(G_{1}(1)\)}}

\end{picture}
%\end{center}
%\begin{center}
\begin{picture}(50,41)
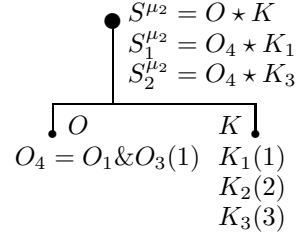
\put(00,00){\makebox(0,0)[bl] {Fig. 31. Treatment
  for point \(\mu_{2}\)}}

%----------------------------------------------

\put(13,26){\line(0,1){11}} \put(13,37){\circle*{2}}

\put(15,37){\makebox(0,0)[bl]{\(S^{\mu_{2}} =O\star K\)}}

\put(15,32){\makebox(0,0)[bl]{\(S^{\mu_{2}}_{1}=O_{4}\star
 K_{1}\)}}

\put(15,28){\makebox(0,0)[bl]{\(S^{\mu_{2}}_{2}=O_{4}\star
 K_{3}\)}}

%------------------------------

\put(05,26){\line(1,0){27}}

\put(05,22){\line(0,1){4}} \put(05,22){\circle*{1}}

\put(7,22){\makebox(0,0)[bl]{\(O\)}}

\put(0,17){\makebox(0,0)[bl]{\(O_{4}=O_{1}\&O_{3}(1)\)}}

\put(32,22){\line(0,1){4}} \put(32,22){\circle*{1}}

\put(27,22){\makebox(0,0)[bl]{\(K\)}}

\put(27,17){\makebox(0,0)[bl]{\(K_{1}(1)\)}}
\put(27,13){\makebox(0,0)[bl]{\(K_{2}(2)\)}}
\put(27,09){\makebox(0,0)[bl]{\(K_{3}(3)\)}}

\end{picture}
\end{center}

\begin{center}
%\begin{picture}(50,67)
\begin{picture}(60,35)
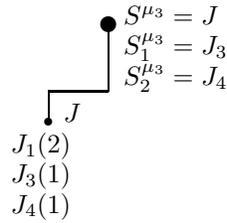
\put(00,00){\makebox(0,0)[bl] {Fig. 32. Treatment
  for point \(\mu_{3}\)}}

%----------------------------------------------

\put(13,22){\line(0,1){09}} \put(13,31){\circle*{2}}

\put(15,31){\makebox(0,0)[bl]{\(S^{\mu_{3}} =J\)}}

\put(15,26){\makebox(0,0)[bl]{\(S^{\mu_{3}}_{1}=J_{3}\)}}

\put(15,22){\makebox(0,0)[bl]{\(S^{\mu_{3}}_{2}=J_{4}\)}}

%------------------------------

\put(05,22){\line(1,0){08}}

\put(05,18){\line(0,1){4}} \put(05,18){\circle*{1}}

\put(7,18){\makebox(0,0)[bl]{\(J\)}}

\put(0,13){\makebox(0,0)[bl]{\(J_{1}(2)\)}}
\put(0,09){\makebox(0,0)[bl]{\(J_{3}(1)\)}}
\put(0,05){\makebox(0,0)[bl]{\(J_{4}(1)\)}}

\end{picture}
%\end{center}
%
%\begin{center}
\begin{picture}(50,37)
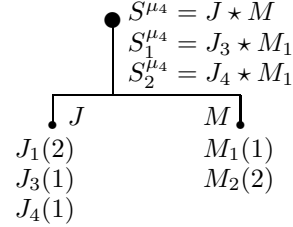
\put(00,00){\makebox(0,0)[bl] {Fig. 33. Treatment
  for point \(\mu_{4}\)}}

%----------------------------------------------

\put(13,22){\line(0,1){10}} \put(13,32){\circle*{2}}

\put(15,32){\makebox(0,0)[bl]{\(S^{\mu_{4}} =J\star M\)}}

\put(15,27){\makebox(0,0)[bl]{\(S^{\mu_{4}}_{1}=J_{3}\star
 M_{1}\)}}

\put(15,23){\makebox(0,0)[bl]{\(S^{\mu_{4}}_{2}=J_{4}\star
 M_{1}\)}}

%------------------------------

\put(05,22){\line(1,0){25}}

\put(05,18){\line(0,1){4}} \put(05,18){\circle*{1}}

\put(7,18){\makebox(0,0)[bl]{\(J\)}}

\put(0,13){\makebox(0,0)[bl]{\(J_{1}(2)\)}}
\put(0,09){\makebox(0,0)[bl]{\(J_{3}(1)\)}}
\put(0,05){\makebox(0,0)[bl]{\(J_{4}(1)\)}}

%---

\put(30,18){\line(0,1){4}} \put(30,18){\circle*{1}}

\put(25,18){\makebox(0,0)[bl]{\(M\)}}

\put(25,13){\makebox(0,0)[bl]{\(M_{1}(1)\)}}
\put(25,09){\makebox(0,0)[bl]{\(M_{2}(2)\)}}

\end{picture}
\end{center}

 Fig. 34 depicts
 an individual version of  medical treatment scheme
 with designed Pareto-efficient composite DAs.
 Evidently, the scheme version has to be designed for
 the certain patient at a preliminary phase
 (i.e., the individual scheme).
 An example of the final individual medical treatment trajectory
 is:

 \(L^{ind} =  < S^{\mu_{0}}_{3} \rightarrow
  S^{\mu_{0}}_{1}   \rightarrow
 S^{\mu_{4}}_{2} \rightarrow S^{\mu_{2}}_{1} >\).

\begin{center}
\begin{picture}(89,53)
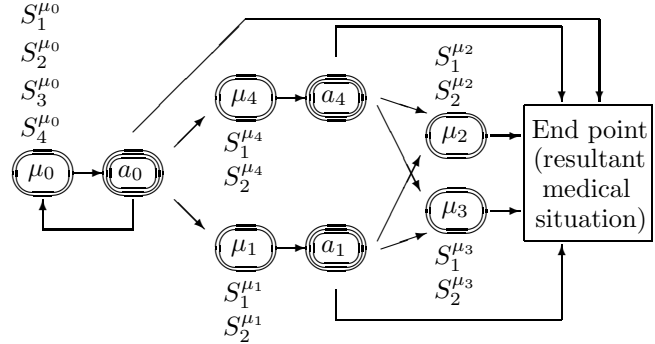
\put(00,00){\makebox(0,0)[bl]{Fig. 34. Individual version of
% basic
 treatment scheme with DAs}}

%+++++++++++++++++++++++++++++++++++++++++++++ Initial point
%+++++++++++++++++++++++++++++++++++++++++++++ tau0-a0

\put(02,48){\makebox(0,8)[bl]{\(S^{\mu_{0}}_{1}\)}}
\put(02,43){\makebox(0,8)[bl]{\(S^{\mu_{0}}_{2}\)}}
\put(02,38){\makebox(0,8)[bl]{\(S^{\mu_{0}}_{3}\)}}
\put(02,33){\makebox(0,8)[bl]{\(S^{\mu_{0}}_{4}\)}}

\put(05,29){\oval(8,6)} \put(05,29){\oval(7,5)}
\put(03,28){\makebox(0,8)[bl]{\(\mu_{0}\)}}

\put(09,29){\vector(1,0){4}}

%--A0

\put(17,29){\oval(8,6)} \put(17,29){\oval(7,5)}
\put(17,29){\oval(6,4)}

\put(15,28){\makebox(0,8)[bl]{\(a_{0}\)}}

\put(22.5,32.5){\vector(1,1){4}} \put(22.5,25.5){\vector(1,-1){4}}

\put(17,25.5){\line(0,-1){4}} \put(17,21.5){\line(-1,0){12}}
\put(05,21.5){\vector(0,1){4}}

\put(17,34.5){\line(1,1){15}} \put(32,49.5){\line(1,0){47}}
\put(79,49.5){\vector(0,-1){11}}

%--------------+++++++++++++++++++++++++++++ tau1-a1

\put(29,11){\makebox(0,8)[bl]{\(S^{\mu_{1}}_{1}\)}}
\put(29,06.5){\makebox(0,8)[bl]{\(S^{\mu_{1}}_{2}\)}}

\put(32,19){\oval(8,6)} \put(32,19){\oval(7,5)}
\put(30,18){\makebox(0,8)[bl]{\(\mu_{1}\)}}

\put(36,19){\vector(1,0){4}}

%--A1

\put(44,19){\oval(8,6)} \put(44,19){\oval(7,5)}
\put(44,19){\oval(6,4)}

\put(42,18){\makebox(0,8)[bl]{\(a_{1}\)}}

\put(49.5,20){\vector(1,2){5.7}} \put(50,19){\vector(3,1){6}}

\put(44,13.5){\line(0,-1){4}} \put(44,09.5){\line(1,0){30}}
\put(74,09.5){\vector(0,1){10}}

%---------------------------------------------

\put(60,24){\oval(8,6)} \put(60,24){\oval(7,5)}
\put(58,23){\makebox(0,8)[bl]{\(\mu_{3}\)}}

\put(64.5,24){\vector(1,0){4}}

\put(57,16){\makebox(0,8)[bl]{\(S^{\mu_{3}}_{1}\)}}
\put(57,11.5){\makebox(0,8)[bl]{\(S^{\mu_{3}}_{2}\)}}

%%%%%%%%%%%%%%%%%%%%%%%%%%%%%%%%%%%%%%%%%%%%%%
%--+++++++++++++++++++++++++++++ tau4-a4

\put(32,39){\oval(8,6)} \put(32,39){\oval(7,5)}
\put(30,38){\makebox(0,8)[bl]{\(\mu_{4}\)}}

\put(36,39){\vector(1,0){4}}

\put(29,31){\makebox(0,8)[bl]{\(S^{\mu_{4}}_{1}\)}}
\put(29,26.5){\makebox(0,8)[bl]{\(S^{\mu_{4}}_{2}\)}}

%--A4

\put(44,39){\oval(8,6)} \put(44,39){\oval(7,5)}
\put(44,39){\oval(6,4)}

\put(42,38){\makebox(0,8)[bl]{\(a_{4}\)}}

\put(49.5,38){\vector(1,-2){5.7}} \put(50,39){\vector(3,-1){6}}

\put(44,44.5){\line(0,1){4}} \put(44,48.5){\line(1,0){30}}
\put(74,48.5){\vector(0,-1){10}}

%------------------------------------------

\put(60,34){\oval(8,6)} \put(60,34){\oval(7,5)}
\put(58,33){\makebox(0,8)[bl]{\(\mu_{2}\)}}

\put(64.5,34){\vector(1,0){4}}

\put(57,42.5){\makebox(0,8)[bl]{\(S^{\mu_{2}}_{1}\)}}
\put(57,38){\makebox(0,8)[bl]{\(S^{\mu_{2}}_{2}\)}}

%+++++++++++++++++++++++++++++++++++++++++++++ End point

\put(70,33){\makebox(0,8)[bl]{End point}}
\put(70.2,29){\makebox(0,8)[bl]{(resultant}}
\put(71.4,25.8){\makebox(0,8)[bl]{medical}}
\put(70.5,21){\makebox(0,8)[bl]{situation)}}

\put(69,20){\line(1,0){17}}\put(69,38){\line(1,0){17}}
\put(69,20){\line(0,1){18}}\put(86,20){\line(0,1){18}}

\end{picture}
\end{center}
%

%%%%%%%%%%%%%%%%%%%%%%%%%%%%%%%%%%%%%%%%%%%%%%%%%%
\section{Conclusion}

 In the paper,
 a new class of composite multistage decision making problems
 has been suggested and described:
 route/trajectory DM problems.
 This problem class is an extension (by several ways)
 of the well-known routing problem
 as the shortest path problem.
 The suggested problems can be considered as
 ``intelligent'' routing at special ``design/solving space(s)''
 based on a digraph
 over a set of connected composite objects/agents.
 The composite objects can contain the following:
 several alternatives, hierarchy of alternatives,
  subobject of implementation and subobject of analysis).
 In general, the ``design/solving space(s)'' can have multi-layer
 and/or multi-domain structure.
 The solving frameworks are two level ones:
 (i) bottom level as decision making operations/problems over the
 composite objects (e.g., selection/composition of alternatives)
 and
 (ii) top-level as routing problem(s) over the ``design/solving space''
 (over the set of objects/agents).
 Mainly, problem descriptions are based on structural approach.
 New problems, models, solving frameworks, and applications
 are discussed.
 In addition,
 restructuring  approach for considered route
 decision making problems is described as well.

 Some future research directions can include the following:
%
%%% MODELS METHODS
%
  {\it 1.} analysis, modeling and usage of various kinds of
 ``design/solving spaces'' including dynamical ``design/solving spaces'';
 {\it 2.} study and usage of various basic combinatorial routing
 problems (e.g., spanning trees problems, versions of TSP)
 for construction of the corresponding route/trajectory DM problems;
 {\it 3.} study of multi-layer (hierarchical)
 ``design/solving spaces'' and route/trajectory DM problems over them;
 {\it 4.} special investigation of multiple vehicle routing problems
 (i.e., multi-domain problems)
 including coordination solving modes
 (e.g., as in multi-robot motion planning problems,
 in cooperative path planning for multiple UAVs
  \cite{bell02,choset05,fer98,huang15,lav98,lat91,svet98});
%
% Unmanned Aerial Vehicle UAV
%
%%%%%%%%%%% APPLICATION DOMAINS
%
 {\it 5.} usage of route/trajectory DM problems
 for testing/inspection/maintenance of networked systems;
 {\it 6.} applications of the examined route/trajectory DM problems
 in economics/management (e.g., modeling of firm/project development,
 forecasting, scenario planning);
%
% {\it 6.} applications of the examined route/trajectory DM problems
% in combinatorial chemistry (e.g., drug design);
%
%%%%%%%%%%%% COMPUTER-AIDED  SUPPORT SYSTEM
%
 {\it 7.} designing a special support computer-aided tools
 for the route/trajectory DM problems
 including the following stages:
 (i )problem analysis and descriptions, generation/formulation;
 (ii) building a  ``design/solving space'',
 (iii) planning the solving processes and problem solving;
 (iv) results analysis;
 and
%%%%%%%%%%%%%% EDUCATION
%
 {\it 8.} usage of the considered route/trajectory DM problems
 in education (CS, applied mathematics, engineering, management).

%%%%%%%%%%%%%%%%%%%%%%%%%%%%%%%%%%%%%%%%%%%%%%%%%%%%%%%%%%%%%%%
%\section{Acknowledgments}

% This research (without sections 4)
% was partially supported by Russian Science Foundation:
% Grant 14-50-00150 ``Digital technologies and their applications''
% (project of Inst. for Information Transmission Problems).
%
%
% The research material presented in  sections 4
% was partially supported by Russian Foundation for
% Fundamental Research: grant 15-07-01241
% ``Reconfiguration of Solutions in Combinatorial Optimization''
% (principal investigator: Mark Sh. Levin).

%%%%%%%%%%%%%%%%%%%%%%%%%%%%%%%%%%%%%%%%%%%%%%%%%%%%%%%%
%%%%%%%%%%%%%%%%%%%%%%%%%%%%%%%%%%%%%%%%%%%%%%%%%%%%%%%%

\end{document}